%% file: acl_latex.tex
\pdfoutput=1

\documentclass[11pt]{article}

\usepackage[final]{acl}

\usepackage{times}
\usepackage{latexsym}
\usepackage{graphicx}
\usepackage{subcaption}
\usepackage{enumitem}
\setlist{nolistsep,leftmargin=*}

\usepackage{enumitem}

\usepackage{mdframed}
\usepackage{hyperref}
\usepackage{booktabs}
\usepackage[T1]{fontenc}

\usepackage[utf8]{inputenc}

\usepackage{microtype}

\usepackage{inconsolata}

\usepackage{graphicx}

\usepackage{cuted}
\usepackage{hyperref}

\input{commands}

\title{Research Borderlands: Analysing Writing Across Research Cultures }

\author{
 \textbf{Shaily Bhatt\textsuperscript{1}\thanks{This work started when all authors were at the Allen Institute for Artificial Intelligence (Ai2).}} \\  \texttt{shaily@cmu.edu} \And
 \textbf{Tal August\textsuperscript{2}}\\  \texttt{tagust@illinois.edu}\And
 \textbf{Maria Antoniak\textsuperscript{3}} \\  \texttt{maria.antoniak@colorado.edu} \AND
 \vspace{-25pt}\\
 \normalsize{
 \textsuperscript{1}Language Technologies Institute, Carnegie Mellon University} \\
 \normalsize{\textsuperscript{2}Siebel School of Computing and Data Science, University of Illinois Urbana-Champaign} \\
 \normalsize{\textsuperscript{3}Pioneer Centre for AI, University of Copenhagen}
}

\begin{document}
\maketitle

\newlist{inlinelist}{enumerate*}{1}
\setlist*[inlinelist,1]{%
  label=(\alph*),
}

\input{v1/main}

\bibliography{custom,long-citations,culture_citations}

\appendix

\input{v1/7-appendix}

\end{document}

%% file: commands.tex
\definecolor{maria-color}{HTML}{7881F2}

\definecolor{tal-color}{HTML}{a45aed}

\definecolor{shaily-color}{HTML}{83ad3b}

\usepackage{xcolor}
\definecolor{general-color}{HTML}{d4fae6}
\definecolor{grounded-color}{HTML}{cff5f5}
\definecolor{tool-color}{HTML}{f9d4fa}

\usepackage{quoting}
\quotingsetup{vskip=2pt}

\usepackage{caption}
\usepackage{booktabs}
\usepackage{array}
\newcolumntype{P}[1]{>{\centering\arraybackslash}p{#1}}
\newcolumntype{L}[1]{>{\raggedright\let\newline\\\arraybackslash\hspace{0pt}}p{#1}}
\newcolumntype{R}[1]{>{\raggedleft\arraybackslash}p{#1}}

\usepackage{pgf}
\usepackage{colortbl}

\definecolor{color1}{HTML}{93003a}
\definecolor{color2}{HTML}{cf3759}
\definecolor{color3}{HTML}{f4777f}
\definecolor{color4}{HTML}{ffbcaf}
\definecolor{color5}{HTML}{ffffe0}
\definecolor{color6}{HTML}{a5d5d8}
\definecolor{color7}{HTML}{73a2c6}
\definecolor{color8}{HTML}{4771b2}
\definecolor{color9}{HTML}{00429d}

\usepackage{array}

\newcommand*{\opacity}{30}
\newcommand*{\minval}{-1}
\newcommand*{\maxval}{1}

\newcommand{\gradient}[1]{
    \ifdimcomp{#1pt}{>}{\maxval pt}{#1}{
        \ifdimcomp{#1pt}{<}{\minval pt}{#1}{
            \pgfmathparse{int(round(8*(#1/(\maxval-\minval))-(\minval*(8/(\maxval-\minval)))))}
            \xdef\tempa{\pgfmathresult}
            \ifcase\tempa
                \cellcolor{color1!\opacity} #1\or
                \cellcolor{color2!\opacity} #1\or
                \cellcolor{color3!\opacity} #1\or
                \cellcolor{color4!\opacity} #1\or
                \cellcolor{color5!\opacity} #1\or
                \cellcolor{color6!\opacity} #1\or
                \cellcolor{color7!\opacity} #1\or
                \cellcolor{color8!\opacity} #1\or
                \cellcolor{color9!\opacity} #1
Ω            \fi
    }}
}

\newcommand{\gradientspecificity}[1]{
    \ifnum#1>0
        \cellcolor{green!10} #1
    \else
        \cellcolor{red!10} #1
    \fi
}

%% file: v1/main.tex
\definecolor{forestgreen}{RGB}{34,139,34}
\definecolor{salmon}{RGB}{250,128,114}

\input{v1/0-abstract}
\input{v1/1-intro}

\input{v1/2-related-work}

\input{v1/3-qualitative-study}
\input{v1/4-quantitative-measures}
\input{v1/6-discussion}
\input{v1/8-limitations}
\input{v1/9-ethical-considerations}

\section*{Acknowledgments}

This work was funded by the Allen Institute for Artificial Intelligence (Ai2) and started when all authors were at the Semantic Scholar team at Ai2. First, we heartily thank our interview participants and survey respondents for their thoughtful engagement with us; this work would not have been possible without their time and efforts.
We thank Lauren Klein for feedback and pointers in conceptualizing the work. We are grateful to Joseph Chang for feedback on our study design, Luca Soldaini and Amanpreet Singh for technical support with Semantic Scholar and related tools, Fernando Diaz for feedback on metric operationalisations, and Deepak Nathani for visualizations.  We thank Lucy Li, Saujas Vaduguru, Sireesh Gururaja, and Jeremiah Milbauer for feedback on early drafts of the manuscript.
We thank our reviewers for their constructive criticisms. We are grateful for our friends and colleagues at Semantic Scholar, Ai2, and LTI for rich discussions throughout the work.

%% file: v1/0-abstract.tex
\begin{abstract}
    Improving cultural competence of language technologies is important. However most recent works rarely engage with the communities they study, and instead rely on synthetic setups and imperfect proxies of culture.   
    In this work, we take a human-centered approach to discover and measure language-based cultural norms, and cultural competence of LLMs.
    We focus on a single kind of culture, \textit{research cultures}, and a single task, \textit{adapting writing across research cultures}.
    Through a set of interviews with interdisciplinary researchers, who are experts at moving between cultures, we create a framework of structural, stylistic, rhetorical, and citational norms that vary across research cultures.
    We operationalise these features with a suite of computational metrics and use them for (a) surfacing latent cultural norms in human-written research papers at scale; and (b) highlighting the lack of cultural competence of LLMs, and their tendency to homogenise writing.
    Overall, our work illustrates the efficacy of a human-centered approach to measuring cultural norms in human-written and LLM-generated texts.
\end{abstract}

%% file: v1/1-intro.tex
\section{Introduction}

\textit{What makes an NLP paper an NLP paper?}
Is it when a paper focuses on language technologies?
Is it that the authors are NLP experts or that they use NLP methods?
What if it discusses user perceptions of an NLP technology --- does that make it an HCI paper?
Maybe it is the descriptive ``Figure 1'' or is it the two-column ACL format?

\input{v1/figures/study-overview}

\input{v1/tables/quotes}

These factors are examples of the norms, expectations, and values that characterize communities  \citep{culture_kk_book} and manifest in a community's communication \cite{deardorff_sage_2009}.
Evaluating and aligning large language models (LLMs) to such cultural norms, which are often undocumented, is a difficult but urgent task  \cite{hovy-yang-2021-importance,sorensen2024roadmap}.
However, in most recent works, the concept of ``culture'' remains vaguely defined, if at all \cite{adilazuarda-etal-2024-towards}. 
Many evaluations rely on synthetic setups that lack grounding in specific tasks or cultural contexts, and  operationalise culture through easy-to-use and imperfect proxies like language or nationality \citep{zhou2025culturetriviasocioculturaltheory,qadri2025casethickevaluationscultural}.

We, instead, take a human-centered approach. 
We tackle the definition of the simultaneously vast and highly contextual question of defining and evaluating culture by zooming in on one specific instantiation of culture, \textit{research cultures}, focusing on a specific task, \textit{adapting writing in research papers}, and centering the \textit{community members}, interdisciplinary researchers who are experts in writing for specific research communities.
Using mixed-methods, we develop and operationalise a framework of cultural norms that holistically characterize the writing from different scientific communities, and we measure LLMs' competence in adhering to these research cultural norms. 
Figure  \hyperref[fig:study-overview]{1} shows a complete overview of our study. \footnote{In the rest of the paper, we use ``scientific communities'' and ``research communities'' interchangeably. We use ``research cultures'' to refer to the culture (encompassing norms, values, and expectations) of research communities. We rely on the definition of cultural competence provided by \citet{deardorff_sage_2009}; which, in context of our task, loosely translates to the ability of an LLM to adhere to a set of cultural norms.}

We survey (N=78) and interview (N=10) interdisciplinary researchers to understand the differences in cultural norms in writing across research communities.
We choose interdisciplinary researchers as they are experts at navigating between multiple communities.
The survey (\S\ref{sec:survey}) confirms the ecological validity of our writing adaptation task and guides our choice of proxy to operationalise culture. 
Our interviews (\S\ref{sec:interviews})  elicit researchers' perceived differences in writing norms across communities.
Through qualitative analysis of the interviews, we develop a framework of language-based writing norms that vary across research cultures (\S\ref{section:feature-taxonomy}).

We operationalise this framework using computational metrics and models into an evaluation suite (\S\ref{section:metrics})
which we use in two large-scale quantitative analyses of human-written and LLM-generated scientific writing. 
First, we analyse a corpora of research papers from different scientific communities, surfacing their variations (\S\ref{sec:paper-ananlysis}). 
We find that our metrics recover differences in writing across research cultures at scale, including anecdotal observations of our interviewees. 
Second, reflecting on the growing use of LLMs in scientific writing \citep{liang2024mappingincreasingusellms}, we evaluate the research cultural competence of LLMs (\S\ref{sec:llms}).
 We find that current LLMs struggle to adhere to cultural norms and tend to homogenise writing across communities. 

We contribute to work in evaluating cultural competence by illustrating an alternative, human-centered method of eliciting and measuring cultural norms in human-written and LLM-generated text. We also open-source\footnote{\url{github.com/shaily99/research_borderlands}} our evaluation suite for future research in science-of-science and evaluation of cultural competence of scientific writing tools.

%% file: v1/figures/study-overview.tex
\begin{figure*}[t]
    \centering
    \footnotesize
    \includegraphics[width=0.891\linewidth]{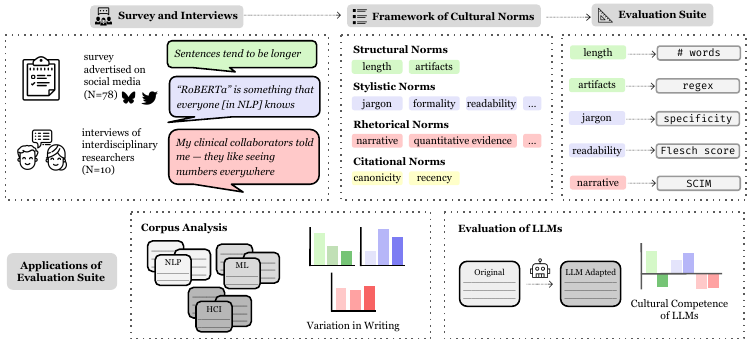}
    \caption{We survey and interview interdisciplinary researchers (\S\ref{sec:qual}) to develop a framework of writing norms that vary across research cultures (\S\ref{section:feature-taxonomy}) and operationalise them using computational metrics (\S\ref{section:metrics}). We then use this evaluation suite for two large-scale quantitative analyses: (a) surfacing variations in writing across 11 communities (\S\ref{sec:paper-ananlysis}); (b) evaluating the cultural competence of LLMs when adapting writing from one community to another (\S\ref{sec:llms}).}
    \label{fig:study-overview}
\end{figure*}

%% file: v1/tables/quotes.tex
\begin{table}[ht]
    \centering
    \small
    \begin{tabular}{p{7cm}}
        \toprule
        P5 on adapting writing: \textit{There's a way to write ... that makes it way more likely a paper with the same results gets accepted or not.}\\
        \vspace{0.025cm}
        
        P2 on tacit norms: \textit{All these norms because they're not stated, you can only speculate based on observation.} \\ 
        \vspace{0.025cm}
        
        P6 on framing: \textit{You can have findings that absolutely blow people away that will get lost if your introduction is not framed in the right way.}
        \\
        \bottomrule
    \end{tabular}
    \caption{Quotes from our interviewees (senior interdisciplinary scholars) on writing across research cultures.
    }
    \label{tab:quotes}
\end{table}        

%% file: v1/2-related-work.tex
\section{Related Work}

\subsection{Understanding Research Communities}

Most prior works in understanding research communities have either focused on a single differentiating feature across many communities, or on understanding a specific aspect in one community.

For example, \citet{lucy-etal-2023-words} analyse the lexical choices and specialized jargon used in different research communities, computationally and at a large scale to measure the \textit{specificity} of a community's written work.  Many prior works have also focused on deep explorations of citational practices, within and across fields \citep{chen2025noisypathsourcecitation,jurgens-etal-2018-measuring,leydesdorff2019interdisciplinarity}.

On the more qualitative side, \citet{birhane2022values} and \citet{jiang2025automaticdetectionresearchvalues} study the values encoded in machine learning research, \citet{michael-etal-2023-nlp} survey the values and beliefs of the NLP research community, \citet{gururaja-etal-2023-build} study the factors that have shaped NLP as a field, and \citet{linxen2021weird} study the demographic biases in studies at the CHI.
Some works have also focused on the framing of specific terms within a research community, such as ``democratization'' \citep{subramonian-etal-2024-understanding} and ``bias'' \citep{blodgett-etal-2020-language}, or  ``intersectionality'' \citep{10.1145/3600211.3604705}.

We add to this body of work by eliciting and analysing a wide variety of writing norms that vary across research communities using mixed-methods.

\subsection{LLM-tools for Scientific Research}

Recently, there has been a rapid rise in building tools to assist researchers in ideation \cite{si2024llmsgeneratenovelresearch}, sense-making \cite{fok2023scim}, data analysis \cite{majumder2024datadrivendiscoverylargegenerative}, agentic assistance \cite{schmidgall2025agentlaboratoryusingllm,nathani2025mlgymnewframeworkbenchmark} and writing \cite{Robinson_2024}. 
These developments either focus on the needs of a single community \cite{Robinson_2024,si2024llmsgeneratenovelresearch,nathani2025mlgymnewframeworkbenchmark} or create general-purpose tools \cite{majumder2024datadrivendiscoverylargegenerative} that do not account for the differing norms across research communities.

Here, we evaluate the potential of general-purpose LLMs as scientific writing assistants with a focus on understanding if they can support the needs of different research communities.

\subsection{Cultural Competence of LLMs}
Recent advancements in the capabilities of LLMs have facilitated their use in a wide range of tasks by diverse users. 
To be useful for people across the world, these systems should adapt depending on the cultural context \cite{hovy-yang-2021-importance}. 
This has led to a surge in interest in evaluating and improving cultural competence\footnote{Also called cultural alignment or cultural awareness.} of LLMs \cite{sorensen2024roadmap,pawar2024surveyculturalawarenesslanguage}. 
However, most contemporary works rarely define ``culture'' or the expected variation in communication norms across cultures. Instead, culture is operationalised through broad proxies like nationality, language, and so on \cite{adilazuarda-etal-2024-towards}. Often, without incorporating perspectives from community members \cite{qadri2025casethickevaluationscultural} or considering the ecological validity of the evaluation setup \cite{bhatt-diaz-2024-extrinsic,zhou2025culturetriviasocioculturaltheory}. These limitations are widely acknowledged \cite{zhou2025culturetriviasocioculturaltheory}.
Contrary to this, we center \textit{community members} and take a holistic lens to discovering and measuring cultural norms, specifically in \textit{research cultures}.

\citet{rao2024normad} is methodologically closest to our approach. They create their evaluation dataset of cultural norms using expert-curated documentation about differences for geographical cultures. However, the documentation they used was not created with an explicit task in mind and the evaluation setting was artificially constructed. In contrast, we focus on a real use case: \textit{adapting research writing for a specific community}. Structuring our interviews around this specific task allows us to gather deep insights into how this task is performed in practice by our interviewees, who as interdisciplinary researchers are experts at this task. 
This grounds our evaluation setup in the needs and expertise of community members.

%% file: v1/3-qualitative-study.tex
\section{Discovering Research Cultural Norms}
\label{sec:qual}
To uncover the tacit knowledge of the cultural norms of different research communities, we conducted a formative survey (N=78) and interview (N=10) study with interdisciplinary researchers.\footnote{We use ``interdisciplinary'' to refer to researchers who both work across communities (cross-disciplinary) and/or who draw on multiple communities (interdisciplinary).} We focused on interdisciplinary researchers as they have experience moving between communities and thus, insight into differing cultural norms. 

\subsection{Preliminary Survey}
\label{sec:survey}

\begin{figure}[t]
    \centering
    \includegraphics[width=0.7\linewidth]{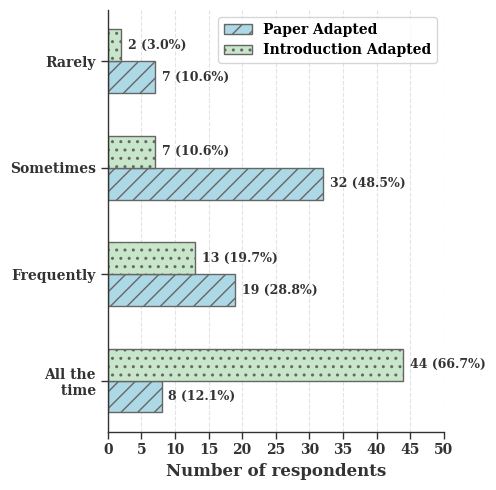}
    \caption{Frequency of adapting papers (blue) and adapting introduction sections when adapting papers ({\color{forestgreen}green}).}
        \label{fig:survey_frequency}
\end{figure}

We conduct a preliminary survey to understand how often researchers adapt written work on the same or similar research for new target communities (e.g., resubmitting a manuscript to a different community after rejection from one) and how they define their communities. We advertised it on our social media accounts and through organizational channels, reaching our professional networks.

We received 78 responses.
After filtering for participants who answered ``Yes'' to 
having adapted papers to different communities, we had 66 responses.

Participants came from a wide variety of communities, listed in Appendix~\ref{app:survey-responses}. Many were CS-focused, with a particular emphasis on NLP (N=32).
The community membership question was free-response (i.e., we did not provide a predefined list of communities).
Only 11 respondents used names of specific publication venues (e.g., ``ICLR'' or ``EMNLP''), while all others used names of fields (e.g., ``NLP'', ``HCI''). Based on this, we structured the rest of our analyses around such communities,
rather than around publication venues.

Figure \ref{fig:survey_frequency} shows that our respondents report varying frequency of adapting papers; however, most indicated that they adapt papers at least sometimes, and when they do adapt papers, they nearly always adapt the introduction section. This validated our initial intuition that interdisciplinary researchers often adapt  writing to specific research communities, confirming it is a real use case. 
Moreover, since the frequency of adaptation of introduction was high, and because it is a section that appears across communities, we chose to focus on introduction sections of  papers for our analyses.

Our respondents were mostly experienced, with 68.8\% having over 5 years of research experience.

The survey confirmed that our chosen task was a real use case, and helped us make design choices including: (a) focusing on introduction sections, and (b) using ``communities'' as the proxy of culture. 
Appendix~\ref{app:survey} contains more details.

\subsection{Expert Interviews}
\label{sec:interviews}

Next, our interview study aimed to identify common features
that researchers adapt, in practice, when writing for different communities.

\paragraph{Participant Selection}
We interviewed 10 of the 39 survey respondents who indicated interest. We prioritized seniority and diversity of communities when selecting participants. Appendix Table \ref{tab:participant_info} lists their self-described communities and expertise.

\paragraph{Protocol}
We conducted semi-structured interviews lasting 60 minutes. 
Before the interview, the participants were asked to select two or more versions of an introduction of one of their papers which they had written for different communities (e.g., a rejected ACL paper rewritten for FAccT).

During interviews, we asked researchers about their perceptions of perceived norms and differences across the communities they worked in and how they adapted papers for them.
We grounded many of these questions to introductions shared by the participants before the interviews.
We asked participants to share their screens and walk us through the differences between the provided samples and their rationales for making those changes. We also discussed whether they used or envisioned any AI tools that could help with this process.
The complete list of questions is in Appendix \ref{appendix:interview-Qs}.

\paragraph{Analysing Interview Transcripts} With the permission of the participants, we recorded and transcribed the interviews. To discover the features that vary across cultures, first, two authors independently coded the first two interview transcripts, labelling any features participants mentioned changing in their papers. Then, through iterative discussion over three weeks, during which additional interviews were coded, all authors agreed upon the framework of cultural norms. Finally, one author coded all interviews with this framework.

\section{Framework of Cultural Norms}

We now discuss the key features that emerged in the interviews as important norms that vary when adapting writing across communities.  These span four categories: structural norms, stylistic norms, rhetorical norms, and citational norms.
\label{section:feature-taxonomy}

\subsection{Structural Norms}
\paragraph{Length}
Most participants (7/10) pointed out that length was one of the major aspects that changed when moving between communities. This is represented in the varying page limits of publication venues across communities.
For example, papers in many NLP conference venues are 8-9 pages, while papers in  FAccT 
are 14 pages.

\paragraph{Artifacts like Tables or Figures}
Four participants mentioned that artifacts like tables or figures were one norm that varied by community. For example, P4 mentioned that ``\textit{[having a] figure having audio spectrograms was really interpretable [in the] community [of] audio researchers}'' for ``\textit{visual storytelling}''.
Similarly, P1 found it useful to know whether a community ``\textit{prefer[s] looking at figures [or] they prefer looking at tables?}'' in context of summarizing their findings at a healthcare venue.

\subsection{Stylistic Norms}

\paragraph{Jargon and Specialized Language} Four participants described that one of the major changes in writing across communities is adjusting the technical jargon and specialized language to match the shared vocabulary in the target community.  For example, P10 mentions not having to define some terms because ``\textit{RoBERTa is something that everyone [in a *CL conference] knows}''. Two other participants reflected, in hindsight, that they should have adapted the jargon in their writing more.
\noindent

Specialized language goes beyond technical jargon. P2 mentioned having to avoid ``\textit{red flag}'' words to prevent being seen as an outsider.

\begin{quoting}
\noindent\textit{P2:  I said the word ``minorities'' and I think [the reviewers] got really upset about that word ... people have very polarized views about what you should be using and so if you use the wrong [word] or if you're not up with the trends ... then you really situate yourself as an outsider}
\end{quoting}

\noindent These findings 
are in line with past work on specialized language in scientific communities \cite{lucy-etal-2023-words, west-portenoy-2016-delineating}.

\paragraph{Readability}
Readability, varying due to both syntax and vocabulary, is another factor researchers adapt. P2 contrasted NLP and education:

\begin{quoting}
\noindent\textit{P2: 
    So in NLP, you might say something like ``much work has talked about largely English models'', but then in an education journal, you'll see the word ``preponderance of work" 
    }
\end{quoting}

\paragraph{Formality}
Two participants pointed out that the ``quality of prose'' (P3, P6) varied across communities.
P3 reflected that informal prose in humanities context allowed for wider variety of argument presentations. In contrast, P4 interpreted formality as ``\textit{stating things mathematically, that maybe could be stated in natural language}''.

\noindent

\paragraph{Verbosity}
Communities had different expectations around verbosity in writing. For example, 
P3 contrasted scientific papers, which value concise language, to humanities:

\begin{quoting}
\noindent\textit{P3: with the humanities context the audience may be a little bit more diverse. You have more space. There's not as much pressure to be concise and you have more time to [show the audience], “why should you care? And how is this related to things that you understand?”}
\end{quoting}

\noindent Interestingly, P5 attributes this difference to the layout of papers because 
``\textit{in this [two-column layout] if you have a longer paragraph ... it'll often take up a whole column which would look sort of unusual.}''

\subsection{Rhetorical Norms}

\paragraph{Quantitative Evidence}
Five participants described 
some communities as having a strong bias in favour of quantitative evidence. Quantitative evidence includes the use of numerical evidence to support a claim, reporting participant statistics,  description of the scale of data and experiments, or the metrics of an algorithm. For example, P1 mentioned that ``\textit{my clinical collaborators told me --- they like seeing numbers everywhere}''.  P2 noted that ``\textit{In NLP we care about numbers. I don't think in education they care about these quantitative things, [or] the scale of things}''. P4 echoed a similar sentiment for CV and ML communities valuing ``\textit{technical contributions and numerical evidence.}''

\noindent

\paragraph{Figurative Language}
Two participants (P3, P6), described frequent use of figurative language and qualitative evidence, including examples and anecdotes in some communities. P6 observed that ``\textit{[in humanities] the article is sometimes trying to capture attention through its lyrical style}'' and ``\textit{there is an expectation for certain publications, an emphasis on introduction as a piece of storytelling}''.

\paragraph{Framing} 
Participants overwhelmingly (9/10) agreed that adapting papers across research communities involved highlighting different aspects, or  ``reframing'' the contributions of the paper.
\begin{quoting}
\noindent\textit{P8: what should be viewed as the icing on the cake versus what's the ... the value of the paper [varies]... the priorities of the community [play a] big part.}
\end{quoting}

\noindent
P6 reflected on the importance of re-framing because ``\textit{you can have findings that absolutely blow people away that will get lost if your introduction is not framed in the right way.}''

\paragraph{Narrative Organization}
Expectations around the narrative organization (or argument structure) of the writing varied. P1 reflected on learning best practices around ``\textit{where do they expect you to start talking about the key contributions? Where do they expect you to fit with existing research more?}''

The relative importance of different types of contributions impacted the narrative organization. P6 observed that if ``\textit{the innovation in the method might be the most important thing about the paper [then] you're going to talk about the method first. }''

Moreover, narrative organization may be more or less formulaic. P3 described that computational communities often have a ``\textit{formulaic structure with background, data, methods, results so on}'' in a similar order. P5 called these ``\textit{recipes}'' but noted that they might be followed to varying degrees.

\subsection{Citational Norms}

\paragraph{Canonicity} Canonical citations varied across communities. P8 reflected 
``\textit{a very similar concept exists in each community and there's a very different canonical citation for it}''. As an example, P4 reflected on analogies between ``\textit{mental models}'' in Cognitive Science and ``\textit{folk theories}'' in HCI.  
Using the right citation was considered important.

\noindent

\begin{quoting}
    \noindent\textit{P3: some of citation is showing your audience, look ``I have read that classic piece that you would want to make sure that I'm aware of''}
\end{quoting}

\noindent
This also highlighted the expectations to cite and engage with the foundational works of a community.
However, other participants described taking a more organic approach to citations, citing ``\textit{whatever seems most appropriate to the project.}'' (P5).

\paragraph{Engagement Style} Two participants highlighted the differences in the forms of engagement with cited works.
``\textit{Using direct quotation very early in a piece is really common [in humanities]}'' but is a ``\textit{little bit less common on the computational side.}''

%% file: v1/4-quantitative-measures.tex
\section{Evaluation Suite}
\label{section:metrics}
We now operationalise a tractable subset of the norms identified in \S\ref{section:feature-taxonomy} using computational metrics.\footnote{We exclude verbosity and figurative language because of lack of reliable metrics, and citations as in-text citations could not be reliably mapped to the respective papers for analysis.}
We use these metrics to surface differences across research communities (\S\ref{sec:paper-ananlysis}) and evaluate LLMs' adherence to these cultural norms (\S\ref{sec:llms}).

\input{v1/figures/paper-metric-figures}

\subsection{Structural Norms}

\paragraph{Length} For each introduction, we record the number of words and sentences. We pre-process the text by lower-casing and stripping URLs and special characters and then use tokenizer from NLTK \citep{bird2009natural}.\footnote{Specifically, word\_tokenize, sent\_tokenize  from \href{https://www.nltk.org}{NLTK}} 

\paragraph{Tables and Figures}
We use regular expressions to find the terms ``table'', ``figure'', and their shorter variants and record a binary label for whether an introduction contains a table, and a binary label for if it contains a figure. 
Details are in Appendix \ref{app:table-figure}. 

\subsection{Stylistic Norms}
\paragraph{Jargon}
We use specificity scores \citep{zhang2017community} to measure jargon, which have been used for this purpose in prior work
\cite{lucy-etal-2023-words}. We first calculate the normalized pointwise mutual information (NPMI) between words and communities.\footnote{We ignore the words with a frequency of <~3 in the corpora, and appear in <~2 communities in the NPMI calculation.} 
The specificity score of an introduction is calculated as the average NPMI of its words to the target community and indicates the uniqueness of introduction's vocabulary to the target community.

\paragraph{Formality} We compute the formality score for an introduction as the average of the formality scores for all its sentences. For sentence-level formality scoring, we use DeBERTa-large fine-tuned on the GYAFC formality classification dataset \cite{dementieva-etal-2023-detecting, rao-tetreault-2018-dear}.\footnote{We use \href{https://huggingface.co/s-nlp/deberta-large-formality-ranker}{DeBerTa-large finetuned on GYAFC dataset}}

\paragraph{Readability} We measure readability as the average sentence-level Flesch reading-ease score \cite{flesch1948new}, calculated using \href{https://github.com/textstat/textstat}{textstat}. A higher score implies the text is easier to read.\footnote{See \href{https://en.wikipedia.org/wiki/Flesch–Kincaid_readability_tests\#Flesch_reading_ease}{Flesch reading ease} for interpretation of this score.}

\subsection{Rhetorical Norms} 

\paragraph{Quantitative Evidence}
We use an LLM-as-a-judge setup \cite{zheng2023judging} to ascertain if a sentence contains quantitative evidence. We then compute the percentage of sentences that contain quantitative evidence in an introduction.
For each sentence in an introduction, we prompt Llama 3.1 70B Instruct with detailed instructions and examples to obtain a binary (``yes'', ``no'') label. We obtain an average agreement of 93\% between LLM ratings and human annotations on a sample of 250 data points. More details are in Appendix \ref{app:quant-evidence}.

\paragraph{Narrative Organization} 
\citet{fok2023scim} categorize the narrative function of sentences in a research paper as describing its \textit{background}, \textit{objectives}, \textit{methods}, or \textit{results}.\footnote{We ignore the \textit{other} category in our analysis.}
Using their multinomial classifier, we obtain a category prediction for each sentence.
We then compute the distribution of length-normalized indices where each category occurs in the introductions and use its skew to capture the relative position of each category.

\paragraph{Framing}
We operationalise framing by identifying \textit{research values} expressed in the sentences \cite{birhane2022values}. We use the 10 values identified by \citet{jiang2025automaticdetectionresearchvalues} and use their human-annotated data to create a multi-label multi-class lexicon classifier. We used the training set (435 samples) to build our initial lexicon, and the validation set (299 samples) to iteratively improve the lexicon. Our final lexicon classifier has an average precision of 72.95\% on the test set.
For an introduction, we record the percentage of sentences in which the each of the 10 value is encoded. We use this to represent an introduction as a 10 dimensional vector. To compare two introductions, we use cosine similarity between their vectors.

\section{Variation of Norms in Research Papers}
\label{sec:paper-ananlysis}

Research papers are a large and tangible collection of text, written by and for a community, implicitly encoding the community's cultural norms. We analyse introductions from 11 communities with our metrics to surface these latent variations, at scale.

\subsection{Data}
We collect a dataset of 81,178 research papers from 11 CS communities (e.g., \textit{NLP}), spanning 38 unique venues. 
We manually map venues to communities\footnote{Our venues to community map is in Appendix \ref{app:paper_data}.}. We use the communities, rather than venues, in our analyses, as motivated by our survey results in \S\ref{sec:survey}. 
We use Semantic Scholar\footnote{\href{https://pys2.readthedocs.io/en/latest/}{PyS2 library}} 
to collect the raw data.
We extract the introduction sections from these texts using regular expression matching of the section titles.
Appendix Figure \ref{fig:field_counts} lists the communities
and their introduction counts.

\input{v1/tables/results_combined_tables}
\subsection{Results}

Figure \ref{fig:all-paper-metrics} and Appendix Figures \ref{fig:all-paper-metrics-appendix} and \ref{fig:scim_distribution} show the metric values for different communities calculated using the evaluation suite in \S\ref{section:metrics}. We also show 95\% confidence intervals of the estimated metric value computed by generating 1000 bootstrap samples.

\paragraph{Syntactic Norms}
We confirm that lengths vary across communities with \textit{Economics \& Computation}
having the longest introductions, both by word and sentence count (Figure \ref{fig:all-paper-metrics-appendix}.a).
\textit{Computer Vision} has the highest frequency of figures, which makes sense given the community's focus on vision, while \textit{NLP} has the most frequency of tables (Figure \ref{fig:all-paper-metrics}.a). 

\paragraph{Stylistic Norms}
Figure \ref{fig:all-paper-metrics}.c shows positive values of specificity scores for all communities,  replicating prior work and confirming interview evidence on the use  of jargon. 
The specificity scores vary more than all the other metrics, and \textit{Education} is the most distinctive community in our corpus. Formality is relatively constant across communities (Figure \ref{fig:all-paper-metrics-appendix}.b).
\textit{Economics \& Computation} has the highest readability, while \textit{Education} has the highest variance in readability (Figure \ref{fig:all-paper-metrics-appendix}.c).  

\paragraph{Rhetorical Norms}
Figure \ref{fig:all-paper-metrics}.b shows that, somewhat surprisingly, \textit{Education} and \textit{Economics \& Computation} have a high percentage of quantitative evidence, albeit a high degree of variance. The variance is smallest for \textit{ML}, \textit{NLP}, and \textit{AI} suggesting that the quantity of quantitative evidence is a strong cultural norm in these communities.\footnote{We also used standard deviation as a measure of strength of the cultural norm and include results in table \ref{table:quant-evidence-std-dev}} 
This matches our participants' observations around \textit{ML} and \textit{NLP} communities valuing a quantitative and numerical evidence (P2, P4).

Figure \ref{fig:scim_distribution} shows the positional density of sentences describing \textit{background}, \textit{objective}, \textit{methods}, and \textit{results} throughout the length of the introduction. Predictably, the density of \textit{background} sentences is highest at the beginning and decreases thereafter. We find that the \textit{objective} sentences are positioned earlier in the introduction
for \textit{ML}, \textit{NLP}, \textit{AI}, \textit{Economics \& Computation}, where other fields have a relatively monotonic increase. Similarly, we observe that \textit{results} are often described earlier in some communities, such as \textit{AI}. 
This could be related to our participants' observations (P2, P4, P6) that some communities, like \textit{AI}, \textit{ML}, and \textit{NLP} value quantitive success of proposed methods. 

Overall, most of our metrics are successful in surfacing the structural, stylistic, and rhetorical norms differences across the communities. Importantly, at scale, we recover many of the same insights that our participants expressed in the interviews. This provides additional evidence for our framework's utility in capturing norms across research cultures.

\section{Cultural Competence of LLMs}
\label{sec:llms}

Four interviewees mentioned having experimented with LLMs with varying degrees of success during writing.
Six participants found utility in the idea of having an ``LLM beta-reviewer'' that could give feedback from the perspective of a specific community before submission, echoing prior work \citep{liao2024llmsresearchtoolslarge}. To support (interdisciplinary) research in such ways, LLMs would need to understand and replicate the nuances of writing in different research cultures. 
To explore this possibility, we evaluate the research cultural competence of LLMs. Using our metrics from \S\ref{section:metrics}, we evaluate whether LLMs adhere to the structural, stylistic, and rhetorical norms of the target community when adapting an introduction from a source community.

\subsection{Experimental Design}
\paragraph{Task} 
We evaluate LLMs' ability to adapt writing from a source to a target community, a task similar to that performed by interdisciplinary researchers. We input the introduction section and prompt an LLM to output an adapted version for a different research community. We then compare the change in metric values of the LLM generations to that of human written data from the target community.

\paragraph{Source Data} 

We use two methods to sample source introductions from the human data corpus from  \S\ref{sec:paper-ananlysis}.
(a) We \textbf{randomly} sample 100 source introductions from each of the 10 remaining communities. (b) For every source-target community pair, we obtain the top 100 most \textbf{specific} introductions, as measured by the specificity metric (\S\ref{section:metrics}), from the source community to the target community. This selects papers that are closer (in vocabulary) to the target community, serving as more realistic examples for adaptation. 
Each sampling method yields 11,000 introductions across all source-target pairs.

\paragraph{Models} 
We use two closed-source and three open~weight models: GPT 3.5 Turbo, GPT 4o Mini, Llama 3.1 8B Instruct, Llama 3.3 70B Instruct, and Mistral Ministral 8B Instruct. 
We sampled five responses per prompt, resulting in 550,000 generations across all LLMs, community pairs, and sampling methods.
Details about the prompt, cost, and hyperparameters are in Appendix \ref{app:llm-adaptations}.

\subsection{Results}

Table \ref{table:ml-nlp} shows the \textbf{change} ($\Delta$) in metrics  after adaptation by the three different LLMs for two target communities.\footnote{The desirable direction of change for a metric is not universal and depends on the target community. For example, for length, the average length of introductions from ML is higher than the weighted average from all other communities. This implies, that the desirable direction of change in length is $\uparrow$. However, NLP introductions are shorter on average than that of other communities, so the opposite is true.}
Results for all other communities and remaining models are in Appendix \ref{sec:all-tables}.

\paragraph{Successful Vocabulary Adaptation}
We see that LLMs' adaptations almost always increase specificity scores, indicating successful adaptation of the vocabulary.
This implies that models do have knowledge of some vocabulary differences across these research communities and that they make lexical changes that remove source-community jargon and/or introduce target-community jargon. 

\paragraph{Homogeneity in Other Metrics}
We observe that across the board, models move all other metrics in a single direction after adaptation. 
For example, the $\Delta$,
of the length in the introduction is negative for all communities, i.e., model outputs are always shorter than inputs. Thus, the model only ``succeeds'' in adapting appropriately when that direction happens to match the community (as for \textit{NLP}, where the introductions are shorter than in other communities).
Prior work in mapping LLM use in scientific writing has also found that papers written with LLMs are shorter \cite{liang2024mappingincreasingusellms}.

Similarly, LLMs always reduce the mention of tables and figures, lower readability, and slightly increase the percentage of sentences with quantitative evidence. 
Narrative organization follows a similar trend, with background and method skew increasing, and objective skew decreasing. 

Overall, LLMs introduce desirable word-level changes, but homogenise all other aspects of writing when adapting writing across communities.

%% file: v1/figures/paper-metric-figures.tex
\begin{figure*}[ht]
    \centering

    \includegraphics[width=0.9\linewidth]{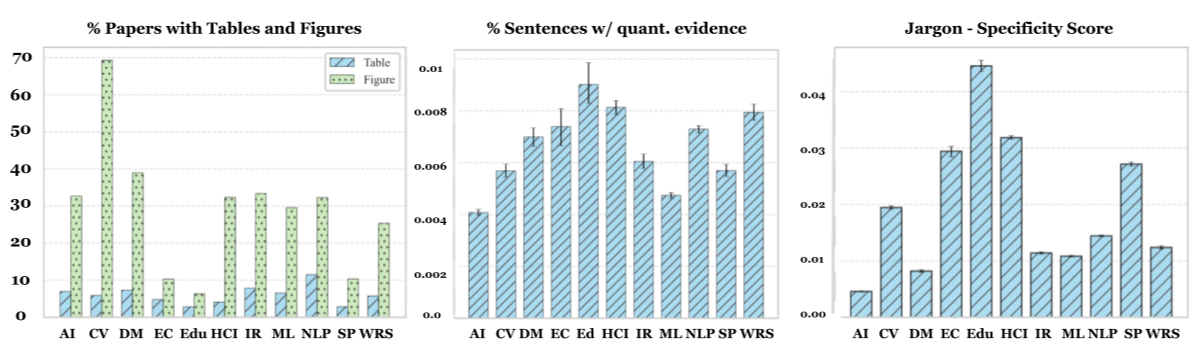}

    \caption{Metric values for four metrics across communities. We observe strong differences for some metrics (e.g., specificity) and less variation for others (e.g., formality) See figures \ref{fig:all-paper-metrics-appendix} and \ref{fig:scim_distribution} in appendix for other metrics.}
    \label{fig:all-paper-metrics}
\end{figure*}

%% file: v1/tables/results_combined_tables.tex
\begin{table*}[th!]
    \centering
    \resizebox{1\linewidth}{!}{
    \footnotesize
    \addtolength{\tabcolsep}{-0.4em}
    \begin{tabular}{@{}lccccccccccccccccc}
    \toprule
    \multicolumn{1}{c}{\textbf{ML}}& \multicolumn{2}{c}{Baselines} & \multicolumn{2}{c}{Adapted by GPT} & \multicolumn{2}{c}{Adapted by Llama} & \multicolumn{2}{c}{Adapted by Mistral} & \multicolumn{1}{c}{\textbf{NLP}}& \multicolumn{2}{c}{Baselines} & \multicolumn{2}{c}{Adapted by GPT} & \multicolumn{2}{c}{Adapted by Llama} & \multicolumn{2}{c}{Adapted by Mistral}
    \\ \cmidrule(lr){2-3}\cmidrule(lr){4-5}\cmidrule(lr){6-7}\cmidrule(lr){8-9}\cmidrule(lr){11-12}\cmidrule(lr){13-14}\cmidrule(lr){15-16}\cmidrule(lr){17-18}
    metric & target & others & random & specific & random & specific & random & specific & metric & target & others & random & specific & random & specific & random & specific
    \\
    \midrule 
    \\ [-10pt]
    \multicolumn{18}{c}{\textit{Structural Norms}}
    \\\midrule 

    Avg. \# words   $\uparrow$    & $+695.38$ & $+601.8$  & \cellcolor{salmon!25} $-363.7$ & \cellcolor{salmon!25} $-282.73$ & \cellcolor{salmon!25} $-185.82$ & \cellcolor{salmon!25} $-123.69$ & \cellcolor{salmon!25} $-106.49$ & \cellcolor{salmon!25} $-37.42$ &   $\downarrow$  & $+648.46$ & $+530.82$  & \cellcolor{forestgreen!40} $-320.43$ & \cellcolor{forestgreen!40} $-302.05$ & \cellcolor{forestgreen!40} $-115.37$ & \cellcolor{forestgreen!40} -117.29 & \cellcolor{forestgreen!40} $-92.45$ & \cellcolor{forestgreen!40} $-84.45$ \\
    Avg. \# sentences $\uparrow$    & $+33.00$ & $+28.70$  & \cellcolor{salmon!25} $-18.03$ & \cellcolor{salmon!25} $-14.94$ & \cellcolor{salmon!25} $-9.54$ & \cellcolor{salmon!25} $-7.01$ & \cellcolor{salmon!25} $-4.86$ & \cellcolor{salmon!25} $-2.08$ &  $\downarrow$  & $+31.36$ & $+23.67$ & \cellcolor{forestgreen!40} $-16.66$ & \cellcolor{forestgreen!40} $-15.60$ & \cellcolor{forestgreen!40} $-6.66$ & \cellcolor{forestgreen!40} $-6.31$ & \cellcolor{forestgreen!40} $-4.18$ & \cellcolor{forestgreen!40} $-3.23$ \\
    \% papers w. table   $\downarrow$    & $+6.57$ & $+601.8$  & \cellcolor{forestgreen!40} -4.87 & \cellcolor{forestgreen!40} -3.1 & \cellcolor{forestgreen!40} -2.35 & \cellcolor{forestgreen!40} -1.92 & \cellcolor{forestgreen!40} -1.91 & \cellcolor{forestgreen!40} -0.97 &   $\uparrow$  & $+11.51$ & $+6.06$  & \cellcolor{salmon!25} -5.31 & \cellcolor{salmon!25} -9.09 & \cellcolor{salmon!25} -2.11 & \cellcolor{salmon!25} -3.23 & \cellcolor{salmon!25} -2.15 & \cellcolor{salmon!25} -3.25 \\
    \% papers w. figure   $\downarrow$    & $29.58$ & +31.8  & \cellcolor{forestgreen!40} -25.85 & \cellcolor{forestgreen!40} -14.46 & \cellcolor{forestgreen!40} -10.95 & \cellcolor{forestgreen!40} -6.56 & \cellcolor{forestgreen!40} -9.73 & \cellcolor{forestgreen!40} -4.41 &  $\uparrow$  & $+32.31$ & $+31.04$ & \cellcolor{salmon!25} -24.04 & \cellcolor{salmon!25} -24.99 & \cellcolor{salmon!25} -6.92 & \cellcolor{salmon!25} -6.05 & \cellcolor{salmon!25} -8.66 & \cellcolor{salmon!25} -7.05 \\
    \midrule 
    \\ [-10pt]
    \multicolumn{18}{c}{\textit{Stylistic Norms}}
    \\\midrule 
    Specificity ($10^{-2}$) $\uparrow$    & $+1.08$ & $-1.66$ & \cellcolor{forestgreen!40} $+0.15$ & \cellcolor{salmon!25} $-0.22$ & \cellcolor{forestgreen!40} $+0.3$ & \cellcolor{forestgreen!40} $+0.04$ & \cellcolor{forestgreen!40} $+0.45$ & \cellcolor{forestgreen!40} $+0.06$ & $\uparrow$    & $+1.44$ & $-1.58$ & \cellcolor{forestgreen!40} $+1.01$ & \cellcolor{forestgreen!40} $+0.31$ & \cellcolor{forestgreen!40} $+1.09$ & \cellcolor{forestgreen!40} $+0.39$ & \cellcolor{forestgreen!40} $+1.08$ & \cellcolor{forestgreen!40} $+0.32$ \\
    Formality ($10^{-2}$)   $\uparrow$    & $+5.62$ & $+5.53$  & \cellcolor{salmon!25} $-0.48$ & \cellcolor{salmon!25} $-0.52$ & \cellcolor{salmon!25} $-0.22$ & \cellcolor{salmon!25} $-0.26$ & \cellcolor{salmon!25} $-0.11$ & \cellcolor{salmon!25} $-0.12$ &  $\uparrow$    & $+5.57$ & $+5.54$ & \cellcolor{salmon!25} $-0.41$ & \cellcolor{salmon!25} $-0.5$ & \cellcolor{salmon!25} $-0.15$ & \cellcolor{salmon!25} $-0.19$ &\cellcolor{salmon!25} $-0.08$ & \cellcolor{salmon!25} $-0.12$ \\
    Readability $\downarrow$  & $+28.74$ & $+29.43$ & \cellcolor{forestgreen!40} $-18.24$ & \cellcolor{forestgreen!40} $-15.21$ & \cellcolor{forestgreen!40} $-5.92$ & \cellcolor{forestgreen!40} $-3.93$ & \cellcolor{forestgreen!40} $-4.08$ & \cellcolor{forestgreen!40} $-3.55$ &       $\uparrow$    & $+32.87$ & $+28.23$ & \cellcolor{salmon!25} $-18.33$ & \cellcolor{salmon!25} $-18.38$ & \cellcolor{salmon!25} $-4.09$ & \cellcolor{salmon!25} $-4.20$ & \cellcolor{salmon!25} $-3.92$ & \cellcolor{salmon!25} $-3.58$ \\
    \midrule 
    \\ [-10pt]
    \multicolumn{18}{c}{\textit{Rhetorical Norms}}
    \\\midrule
    \% Quant. Evidence  $\downarrow$  & $+0.47$ & $+0.63$ & \cellcolor{salmon!25} $+1.03 $& \cellcolor{salmon!25} $+1.31$ & \cellcolor{salmon!25} $+3.48$ & \cellcolor{salmon!25} $+3.07$ & \cellcolor{salmon!25}$ +3.04$ & \cellcolor{salmon!25} $+2.96 $&    $\uparrow$    & +0.73 & +0.56 & \cellcolor{forestgreen!40} +1.23 & \cellcolor{forestgreen!40} +1.51 & \cellcolor{forestgreen!40} +3.15 & \cellcolor{forestgreen!40} +3.45 & \cellcolor{forestgreen!40} +2.82 & \cellcolor{forestgreen!40} +3.38 \\
    Sim. in Framing $\uparrow$   & base & 0.88 & \cellcolor{forestgreen!40} +0.01 & 0.0 & \cellcolor{forestgreen!40} +0.02 & 0.0 & 0.0 & 0.0 &  $\uparrow$ & base & 0.97 & \cellcolor{forestgreen!40} + 0.01 & \cellcolor{forestgreen!40} +0.01 & \cellcolor{forestgreen!40} +0.01 & \cellcolor{forestgreen!40} +0.01 & \cellcolor{forestgreen!40}  +0.03 & 0.0 \\
    Background Skew $\uparrow$   & +0.56 & +0.5 & \cellcolor{forestgreen!40} +0.31 & \cellcolor{forestgreen!40} +0.29 & \cellcolor{forestgreen!40} +0.05 & \cellcolor{salmon!25} -0.02 & \cellcolor{forestgreen!40} +0.09 & \cellcolor{forestgreen!40} +0.01 &  $\uparrow$    & +0.58 & +0.51 & \cellcolor{forestgreen!40} +0.23 & \cellcolor{forestgreen!40} +0.19 & \cellcolor{forestgreen!40} +0.04 & \cellcolor{forestgreen!40} +0.01 & \cellcolor{forestgreen!40} +0.1 & \cellcolor{forestgreen!40} +0.07 \\
    Objective Skew   $\uparrow$  & +0.02 & 0.0 & \cellcolor{salmon!25} -0.5 & \cellcolor{salmon!25} -0.4 & \cellcolor{salmon!25} -0.17 & \cellcolor{salmon!25} -0.07 & \cellcolor{salmon!25} -0.12 & \cellcolor{salmon!25} -0.04  &   $\uparrow$    & +0.16 & -0.05 & \cellcolor{salmon!25} -0.47 & \cellcolor{salmon!25} -0.46 & \cellcolor{salmon!25} -0.09 & \cellcolor{salmon!25} -0.05 & \cellcolor{salmon!25} -0.07 & \cellcolor{salmon!25} -0.04 \\
    Method Skew   $\uparrow$     & -0.31 & -0.42 & \cellcolor{forestgreen!40} +0.18 & \cellcolor{forestgreen!40} +0.18 & \cellcolor{forestgreen!40} +0.01 & \cellcolor{forestgreen!40} +0.01 & \cellcolor{forestgreen!40} +0.07 & \cellcolor{forestgreen!40} +0.06  &    $\uparrow$    & -0.37 & -0.4 & \cellcolor{forestgreen!40} +0.11 & \cellcolor{forestgreen!40} +0.03 & \cellcolor{forestgreen!40} +0.02 & \cellcolor{salmon!25} -0.03 & \cellcolor{forestgreen!40} +0.09 & \cellcolor{forestgreen!40} +0.02 \\
    Result Skew   $\uparrow$     & -0.18 & -0.27 & \cellcolor{salmon!25} -0.48 & \cellcolor{salmon!25} -0.53 & \cellcolor{salmon!25} -0.01 & \cellcolor{salmon!25} -0.01 & \cellcolor{salmon!25} -0.06 & \cellcolor{forestgreen!40} +0.01   &  $\downarrow$  & -0.35 & -0.23 & \cellcolor{forestgreen!40} -0.26 & \cellcolor{forestgreen!40} -0.38 & \cellcolor{forestgreen!40} -0.06 & \cellcolor{salmon!25} +0.03 & \cellcolor{forestgreen!40} -0.1 & \cellcolor{forestgreen!40} -0.04 \\

    \bottomrule \\
    \end{tabular}
    }
    \caption{\small
        Cultural competence of LLMs towards \textbf{ML} and \textbf{NLP}. The \textit{target} column shows the metric value for human-written papers from the community, and the \textit{others} column shows the weighted average of all other communities. Based on these, $\uparrow$ indicates that the metric should increase after adaptation and $\downarrow$ indicates the vice versa. The model columns show the \textbf{change} ($\Delta$) in metric value after adaptation for the respective samples.  Cells where $\Delta$ follows the expected trend {\color{forestgreen}green} and others are {\color{salmon}red}.
        }
    \label{table:ml-nlp}
\end{table*}

%% file: v1/6-discussion.tex
\section{Discussion}

\paragraph{Engaging with Community Members}
Methodologically, our approach to understanding and evaluating cultural norms is different from contemporary work. 
Most recent works take a `top-down' approach by operationalising culture with a specific proxy, like nationality. This takes a narrow view of culture \cite{zhou2025culturetriviasocioculturaltheory} and does not consider the relevance of the proxy for the task and societal context \cite{qadri2025casethickevaluationscultural}. Prior work has cautioned against such naive adoption of identity axes from the western world to other cultural contexts \cite{sambasivan2021reimagining,Bhatt2022RecontextualizingFI}.

Our approach, in contrast, is bottom-up. We derived our choice of proxy through surveying community members. 
To determine the salient cultural norms important to writing across these communities, we interviewed community members who regularly perform this task.
This allowed us to build and operationalise a holistic framework of cultural norms.
The success of this approach is demonstrated by its utility for our quantitative analyses of real and synthetic scientific text.

\paragraph{Looking Ahead to LLM Tools} 

Our evaluation suggests that LLMs, at least in a zero-shot manner, do not perform well on the task of adapting writing to research communities. 
Concerningly, we observe that LLMs tend to move the metrics on almost all of our features in a single direction irrespective of the target community. 
We posit that this is a symptom of the larger issues of homogeneity in writing that are starting to be discovered in LLMs, including reduction in both linguistic \cite{guo-etal-2024-curious} and rhetorical diversity \cite{xu2024echoesaiquantifyinglack} in writing. 
Work on tracking the usage of ChatGPT in scientific writings has raised similar concerns about papers written with LLMs being more similar to each other \cite{liang2024mappingincreasingusellms}. 
These risks homogeneity of writing and ideas in the scientific community could be detrimental in the long run \citep{liao2024llmsresearchtoolslarge}. 
Whether personalized and community-specific systems could reduce these risks remains to be seen.
LLMs also regurgitate varying lengths of structural and lexical sequences from their training data \cite{shaib-etal-2024-detection,lu2025aihumanityssalieriquantifying}, which could inadvertently amount to plagiarism. 
These risks need to be weighed against potential benefits, e.g., for non-fluent English speakers such adaptations might help in overcoming structural barriers \citep{lepp2025you}.

Finally, we emphasize the importance of the human process of interdisciplinary writing adaptation, and scientific writing more broadly (P5: ``\textit{I tend to write to think so, I don't think I would want anything that helps that tries to write for me.}'').
Through scientific writing, researchers engage deeply with their communities, not only reflecting on their contributions, but also shaping the community's values through integration or critique \citep{birhane2022values}. As such, the future of scientific writing with LLMs likely lies in interactive writing assistance that exposes community barriers without fully automating the writing process.

%% file: v1/8-limitations.tex
\section{Conclusion}
In this work, we illustrate a human-centered approach to discovering and measuring cultural norms in writing. 
We use qualitative methods to engage with interdisciplinary researchers, develop a framework of language-based norms,
operationalise this framework using computational metrics, and demonstrate its efficacy in analysing human-written and LLM-generated scientific text. 
We hope our work serves as a motivational case-study in adapting participatory approaches to evaluating cultural considerations for LLMs.

\section{Limitations}

The introduction texts used in this study are restricted to computer science disciplines, and we do not consider disciplines like sociology, art history, biology, etc. except as they intersect with computer science.
Similarly, our interview participants included researchers in ML, NLP, cultural analytics, and computational social science, reflecting biases in our social media recruiting methods.

We chose to uncover cultural aspects by asking interdisciplinary researchers to explain how they write across disciplines, but we could have used another interview method (e.g., asking single-discipline experts to describe their discipline) to answer the same research questions, which might have revealed a different feature set.

We found that our formality metric might not be sensitive to capturing differences if any exist.
This could be because all of our human-written research papers are from CS communities, and so the differences in formality and framing across these communities might not be as pronounced. Or it could be because the research paper data is somewhat out of distribution for the classifier being used. More broadly, our choice metrics represent one version out of many possible operationalisation of the cultural norms; we leave further exploration of metric improvement and choices to future work. 

%% file: v1/9-ethical-considerations.tex
\section{Ethical Considerations}

Our study was approved by the IRB at the Allen Institute for AI, and participants were paid \$40 for their participation.
Participants signed a consent form agreeing to recording of the interviews.
We do not release the raw interview data, such as recordings or transcripts, and only provide summary statistics and brief quotations in this paper.

We also cannot release the full set of research papers used for this work due to licensing.

Our work takes a human-centered, bottom-up approach to studying culture.
Prior work in fairness research has called for similar participatory frameworks for better outcomes in understanding harms and impacts of technology on minoritized users, while also cautioning against doing this in a superficial way \citep{delgado2023participatory}. 
Coming from a similar philosophy, we hope our approach serves both as motivation and as illustration in conducting thoughtful, nuanced, and human-centered studies of cultural competence in LLMs.

%% file: v1/7-appendix.tex
\clearpage
\newpage
\section{Survey Details}
\label{app:survey}
\subsection{Survey Questions}
\label{app:survey-qs}

We asked the following questions in the survey:
\begin{enumerate}
    \item Have you ever needed to adapt a paper written for one research community to another research community? [
    \begin{inlinelist}
        \item~Yes
        \item~No
    \end{inlinelist}
    ]
    \item What were the situations in which you needed to adapt a paper written for one research community to another research community? Select all that apply. [
    \begin{inlinelist}
        \item A paper got rejected from one community, and I re-wrote it for a different community
        \item The paper was better suited to a different community, than what I initially conceived
        \item I re-wrote a draft written by a co-author who was not familiar with the particular community
        \item I was unfamiliar with the community, and a co-author helped me re-write the content
        \item Other
    \end{inlinelist}
    ]
    \item How often do you encounter the situations in the previous question? [
    \begin{inlinelist}
        \item All the time
        \item Frequently
        \item Sometimes
        \item Rarely
        \item Never
    \end{inlinelist}
    ]
    \item When adapting a paper for a research community, do you adapt the introduction section? [
    \begin{inlinelist}
        \item All the time
        \item Frequently
        \item Sometimes
        \item Rarely
        \item Never
    \end{inlinelist}
    ]
    \item How long have you been doing research? [
    \begin{inlinelist}
        \item I have no research experience
        \item 0-1 years
        \item 2-4 years
        \item 5-7 years
        \item 8+ years
    \end{inlinelist}
    ]
    \item What are the different research communities that you have needed to adapt between? List all those communities here.
    \item Please share your email if you would be willing to give a longer interview about your experience with writing across research communities.

\end{enumerate}

\subsection{Survey Response Answers}
\label{app:survey-responses}
\begin{figure}
    \includegraphics[width=\linewidth]{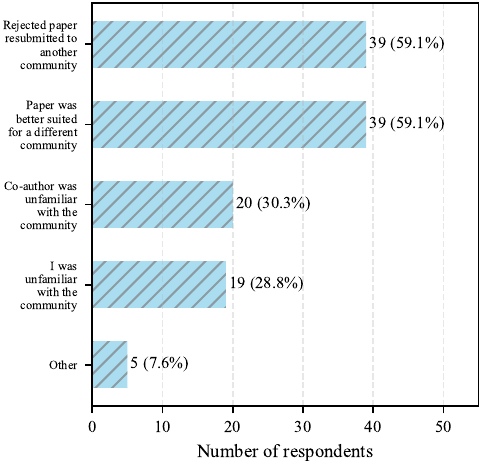}
    \caption{Situations of adaptations}
    \label{fig:survey_situation}
    \caption{Situations in which participant described having performed writing adaptations.}
\end{figure}

Figure \ref{fig:survey_situation} depicts the situations under which participants performed writing adaptations. Common situations in which such adaptation need to be made include revising a paper for another community or having co-authors more or less familiar with the community.

Table \ref{tab:field-counts-1} and  \ref{tab:field-counts-2} represents the self-reported communities of survey respondents. We normalized the original free-form responses to combine different names used for the same thing (for example, Natural Language Processing and NLP). Since most of the participants self-reported community or field-level information, we mapped those who reported venue level information to the respective fields.

\begin{table}[h]
\centering
\begin{tabular}{lc}

\toprule
\textbf{Community} & \textbf{Count} \\
\midrule
NLP & 32 \\
ML & 13 \\
HCI & 11 \\
Digital Humanities & 11 \\
Computational Social Science & 8 \\
Political Science & 7 \\
Linguistics & 7 \\
CV & 6 \\
Communication & 5 \\
Information Science & 5 \\
FAccT & 4 \\
Literary Studies & 4 \\
Psychology & 4 \\
Social Science & 4 \\
Speech & 4 \\
Healthcare & 4 \\
History & 3 \\
Law & 3 \\
Cognitive Science & 2 \\
Cultural Analytics & 2 \\
Data Science & 2 \\
Design & 2 \\
English & 2 \\
Environmental Humanities & 2 \\
Security & 2 \\
Social Sciences & 2 \\
Sociology & 2 \\
Web & 2 \\
Library Science & 2 \\
Culture & 2 \\
\bottomrule
\end{tabular}
\caption{Communities of survey respondents (Part 1)}
\label{tab:field-counts-1}
\end{table}

\begin{table}[h]
\centering
\begin{tabular}{lc}

\toprule
\textbf{Community} & \textbf{Count} \\
\midrule
Astronomy & 1 \\
Civil Society & 1 \\
Computer Graphics & 1 \\
Conservation Biology & 1 \\
Construction & 1 \\
Control Systems & 1 \\
Cryptography & 1 \\
Data Mining & 1 \\
Database Management & 1 \\
Ecology & 1 \\
Economics & 1 \\
Education & 1 \\
Environmental Management & 1 \\
Geoinformatics & 1 \\
Human Rights & 1 \\
Humanities & 1 \\
Law and Society & 1 \\
Molecular Biology & 1 \\
Narratology & 1 \\
Optimization & 1 \\
Philosophy & 1 \\
Policy & 1 \\
Programming Languages & 1 \\
Robotics & 1 \\
Science of Science & 1 \\
Semiotics & 1 \\
Theory & 1 \\
\bottomrule
\end{tabular}
\caption{Communities of survey respondents (Part 2)}
\label{tab:field-counts-2}
\end{table}

\section{Interview Details}
\subsection{Interview Questions}
\label{appendix:interview-Qs}
We divided the questions into three types: (a) general -- questions related to the fields and the participant's research broadly. (b) grounded -- questions based on the introduction sampled provided by the participant. (c) tool use -- questions related to the types of tools that participants use when adapting writing. Following is the complete list of questions.

\begin{enumerate}
    \item \colorbox{general-color}{\texttt{general}} Can you give a brief description of your research? 
    \item \colorbox{general-color}{\texttt{general}} Can you describe the communities you have written for?
    \item \colorbox{general-color}{\texttt{general}} What is your general approach towards writing introductions? Does your approach change when you write introductions for the different communities?
    \item \colorbox{general-color}{\texttt{general}} Are there specific things that these communities expect out of introductions that are different? 
    \item \colorbox{grounded-color}{\texttt{grounded}} Can you give me a brief overview of the project itself?
    \item \colorbox{grounded-color}{\texttt{grounded}} Can you give me a brief overview of the different versions of the introductions you have here, in terms of which communities they were written for?
    \item \colorbox{grounded-color}{\texttt{grounded}} What were the parts of the introduction that were adapted for the particular paper? How did these vary over the different versions?
    \item \colorbox{grounded-color}{\texttt{grounded}} Why did you make these changes?
    \item \colorbox{grounded-color}{\texttt{grounded}} Do you think the final introduction reflects what a typical introduction looks like for this community?
    \item \colorbox{grounded-color}{\texttt{grounded}} Do you think the reviews you got back from the community reflected these expectations that you tried to match when you were making those adaptations?
    \item \colorbox{grounded-color}{\texttt{grounded}} What are some of the aspects you keep in mind (that may or may not have come up so far) when translating between these communities?
    \item \colorbox{tool-color}{\texttt{tool use}} \colorbox{general-color}{\texttt{general}} What are some of the challenges that you face in making these adaptations? How do you overcome these challenges right now?
    \item \colorbox{tool-color}{\texttt{tool use}} Can you think of what kind of technology might be helpful for making your life easier when doing this task?
    \item \colorbox{tool-color}{\texttt{tool use}} Would you rather have a retrieval system, or a beta-reviewer, or a rewriter?
\end{enumerate}

\subsection{Interview Participant Expertise}
Expertise of interview participants is reported in \ref{tab:participant_info}
\begin{table*}[h!]
\centering
\small
\begin{tabular}{clcl}
\toprule
\textbf{Participant} & \textbf{Position} &  \textbf{Years of  Experience} & \textbf{Communities} \\
\midrule
P1 & Industry Researcher & 6-8 & NLP, Healthcare \\
P2 & PhD Candidate & 6-8 & NLP, Education, Cultural Analytics \\
P3 & Professor & 8+ & NLP/CL, CSS, Digital Humanities, Literary Studies \\
P4 & Asst. Professor & 8+ & ML, CV, Audio, HCI \\
P5 & Assoc. Professor & 8+ & NLP, CSS, Sociology, Political Science \\
P6 & Asst. Professor & 8+ & Data Science, Cultural Analytics, History \\& & & Digital Humanities, Literary Studies\\
P7 & Postdoctoral Researcher & 6-8 & ML, Human Rights, Civil Society, Policy, Humanities \\
P8 & Asst. Professor & 8+ & NLP, ML, Computer Vision, Robotics \\
P9 & Asst. Professor & 8+ & NLP, Linguistics \\
P10 & Industry Researcher & 6-8 & NLP, ML, Speech, HCI \\
\bottomrule
\end{tabular}
\caption{Expertise of interview participants.}
\label{tab:participant_info}
\end{table*}

\section{Feature operationalisations}
\label{appendix-features}

\subsection{Tables and Figures}
\label{app:table-figure}
We use a custom lexicon to detect whether an introduction contains tables and figures. This is because, since we only parse the text of the papers, we do not get tables and figures in line in the text. By hand-labelling a few samples by looking at their real PDFs, it is clear that if an introduction has a table or a figure, it usually contains phrases in the text that explicitly mention the table or figure. We use this information to construct a simple lexicon that we match across the introduction section. We record a binary label to whether an introduction section contains table and a binary label to whether an introduction section contains a figure. After matching using this lexicon, authors hand-validated a sample of 10 introductions per community for correctness of labels and found a 100\% accuracy. The lexicon we used is in table \ref{tab:tab_fig_lexicon}

\begin{table}[th!]
    \centering
    \footnotesize
    \begin{tabular}{ll}
    \toprule
        \textbf{Artefact} & \textbf{Lexicon} \\
         \midrule
         Table & "table", "tab", "tab.", "tabs", \\& "tabs.", "tables"\\
         Figure & "figure", "fig", "fig.", "figs", \\& "figs.", "figures", "figure."\\
         \bottomrule
    \end{tabular}
    \caption{Lexicon for detecting tables and figures.}
    \label{tab:tab_fig_lexicon}
\end{table}

\subsection{Quantitative Evidence}
\label{app:quant-evidence}

We used the prompt in table \ref{tab:quant_prompt} to prompt Llama 3.1 70B Instruct \cite{grattafiori2024llama3herdmodels} in an auto-rating setup. We sample one output per prompt at a temperature of 0.0 and set max tokens to 5. The model was loaded using vLLM\footnote{\url{vllm.ai}} with 4-bit bitsandbytes quantization \footnote{\url{https://github.com/bitsandbytes-foundation/bitsandbytes}}. For human raters, we use the exact same informational instructions, definition, valid examples, and invalid examples as used in the LLM prompt. We remove the output format instructions at the end the prompt for human raters. We calculate agreement between three human raters and the LLM as the percentage times they agree on a label. We obtain an average agreement of 93.68\%.

\begin{table*}[h]
\centering
\footnotesize
\begin{tabular}{lc}
\toprule
\textbf{Quantitative Evidence Prompt} \\

\midrule
I'm going to show you a sentence from a computer science research paper. I want you to tell me if the \\
sentence contains quantitative evidence or not. Quantitative evidence is any evidence that is expressed \\
in numbers, such as statistics, percentages, or measurements used to support an implicit or explicit claim.
\\\\
\# Definition \\
A sentence contains quantitative evidence if it includes numbers that provide measurable support for a \\
claim. This may be expressed as percentages, statistics, population counts, measurements, metric scores, \\
monetary values,frequencies, ratios, probabilities, time measurements, and specifications of hardware, \\
software or algorithms.

\\\\
\# Examples: \\
- Percentages: "50\% of the students passed the exam." \\
- Statistics: "Only 75\% of the participants agreed that internet was an essential part of their life." \\
- Population and counts: "The city's population is 1 million." or "The website has 100,000 users." \\
- Measurements: "The table is 2 meters long." \\
- Metric scores: "The F1 score improved by 10 points." \\
- Monetary values: "The project cost \$1,000." \\
- Frequencies: "The event occurs 3 times a week." or "An average user posts about three hundred tweets in a year". \\
- Ratios: "The ratio of students to teachers is 20:1." \\
- Probabilities: "There is a 70\% chance of rain tomorrow." \\
- Time measurements: "The meeting will last 2 hours." \\
- Hardware specifications: "The computer has 16GB of RAM." \\
- Software specifications: "The software requires 4GB of disk space." \\
- Algorithm specifications: "The score is computed with an O(1) time complexity." \\
- Quantification of scale: "Domain experts often read through millions of documents to identify relevant information."
\\
Numerical data may appear in different formats (e.g., "50 percent" or \\"fifty percent" or "hundred images") but still qualify as quantitative evidence.

\\\\
\# Invalid Examples: \\
Note, any mention of a number is not quantitative evidence. \\
You should ignore numbers that appear to be part of: \\
- Citations: "The Internet is important [1]" or "The internet is important (Smith et. al. 1982)" \\
- Bullet points: "1. The internet is important" or \\ "Our contributions are: (1) Student teacher ratio matters to learning outcomes." \\
- Information about historical events: "The internet was invented in 1969". \\
- Mathematical expressions: "x = 1" or "2 f + 2 < N" \\
- Names of models, metrics, datasets, or algorithms: "We Llama 2 for the experiments." or \\ "L2 regularization is used to prevent overfitting." \\
- References to the structure of the documents like figures, tables, sections, \\algorithms, theorems, or appendices: "As shown in Figure 1" or \\ "We describe our dataset in Table 3" or "We prove theorem 1 in Appendix 5.6". \\
- Non-quantitative Uses: "one of the approaches we use" or "in the first experiment" or "our two main contributions". \\
- Incoherent text artifacts: Numbers from DOI links, web addresses, arXiv links, email addresses, \\ unicode characters, other metadata text, or unclean and incoherent text should be ignored. \\

\\\
\# Output Format:\\
- Please answer with "yes" if the sentence contains quantitative evidence and "no" if it does not. \\
- Do not answer with anything else. \\
- Do not add any explanation or justification to your answer. \\
- In case the prompt does not contain a sentence or contains only incoherent characters, please answer with "no" \\
\\
Remember to evaluate the sentence carefully and your best judgment to determine \\ if the sentence contains quantitative evidence or not.  \\ Remember, you should focus on the content of the sentence and judge \\ whether it expresses evidence for a claim quantitatively. \\
\bottomrule

    \end{tabular}
    \caption{Prompt for LLM-as-judge setup}
    \label{tab:quant_prompt}
\end{table*}

\subsection{Framing}
\label{appendix:framing}

We capture the framing of research papers by measuring the \textit{values} encoded in each sentence.
Values represent ``desirable attributes'' and are used to frame a study's motivations and justifications \citep{birhane2022values}.
We develop a custom lexicon to capture values, evaluating on a dataset of 1.1k sentences hand-annotated with ten values (e.g., efficiency, generalizability) encoded in these sentences obtained from \cite{jiang2025automaticdetectionresearchvalues}. Table \ref{tab:csvalues_lexicon_precision} shows our validation and test precisions.

The following is the list of values we use, sourced from \citet{jiang2025automaticdetectionresearchvalues}
\paragraph{Performance} refers to the effectiveness (success rate) of a method or model, often (but not always) in comparison to existing approaches with quantitative measures such as accuracy, loss, and error rate. 

\paragraph{Novelty} refers to the pursuit of introducing new things to a field, often by resolving existing gaps in research, extending the boundaries, or opening up new possibilities.

\paragraph{Efficiency} refers to the ability to achieve desired outcomes with minimal resource expenditure, such as time, money, data, memory, storage, and computational power. Scalability is also part of efficiency, because it often seeks to handle large amounts of users, data, and traffic efficiently.

\paragraph{Generalizability} refers to the ability to adapt and perform well across a wide range of tasks, conditions, and scenarios. Also known as generalization, universality, adaptability, robustness, flexibility, extensibility.

\paragraph{Understanding} (Phenomenon Understanding \& Theoretical Grounding) refers to understanding phenomena by (1) by providing empirical evidence and insights, (2) by citing, developing, and applying theories, proofs, and theoretical frameworks.

\paragraph{Simplicity} refers to the pursuit of creating simple and elegant methods, models, and theories that minimize complexity.

\paragraph{Fairness} (Fairness, Bias, Privacy \& Ethics) refers to the commitment to promote equity and social justice, avoid social bias, ensure privacy and security, and address ethical issues in the use of computer technologies.

\paragraph{Society} (Societal Implications) refer to the potential of research to impact social change, promote well-being, and address challenges faced by societies and communities. We focus on societal level impacts, instead of individual-level usability.

\paragraph{Openness} (Openness, Reproducibility, Collaboration, \& Future Work) refers to promoting open science (keeping transparent and sharing information about research procedures, data, methods, and results), reproducibility (ensuring others can repeat the process to obtain the same results), collaboration across different fields, and discussions of future work.

\paragraph{Usability} refers to the commitment to improve user experience and real-world applications by making systems more user-friendly, easy-to-use, interpretable, engaging, popular, inclusive, and accessible.

\begin{table}[]
    \centering
    \footnotesize
    \begin{tabular}{lcc}
        \toprule
        Value & Train-Val & Test \\
        \midrule
        Efficiency & 0.81 & 0.79 \\
        Fairness & 0.84 & 0.96 \\
        Generalizability & 0.74 & 0.87 \\
        Novelty & 0.85 & 0.77 \\
        Openness & 0.78 & 0.67 \\
        Performance & 0.94 & 0.83 \\
        Simplicity & 0.82 & 0.71 \\
        Society & 0.65 & 0.33 \\
        Understanding & 0.71 & 0.67 \\
        Usability & 0.73 & 0.68 \\
        \bottomrule
    \end{tabular}
    \caption{Precision of lexicon-based classifier for framing}
    \label{tab:csvalues_lexicon_precision}
\end{table}

\section{Research Papers Data}
\label{app:paper_data}
\begin{figure}
    \centering
    \includegraphics[width=\linewidth]{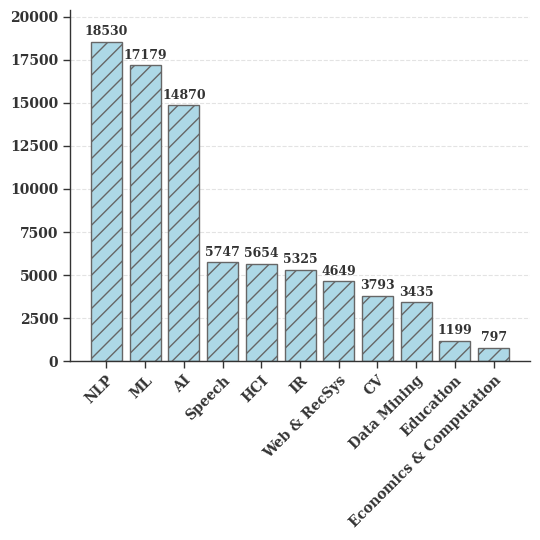}
    \caption{Number of introductions sections from each community in our dataset. We analyse over 81K introductions spanning 11 CS communities.
    }
    \label{fig:field_counts}
\end{figure}

\begin{figure*}[ht]
    \centering
    \begin{subfigure}[b]{0.32\textwidth}
        \includegraphics[width=\textwidth]{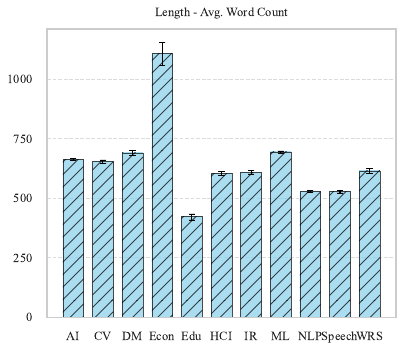}
    \end{subfigure}
    \begin{subfigure}[b]{0.32\textwidth}
        \includegraphics[width=\textwidth]{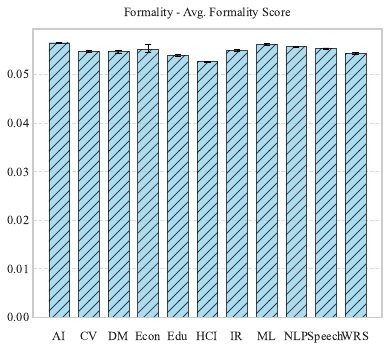}
    \end{subfigure}
    \begin{subfigure}[b]{0.32\textwidth}
        \includegraphics[width=\textwidth]{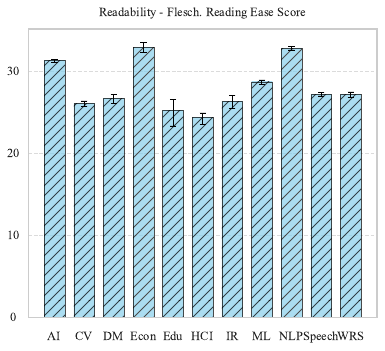}
    \end{subfigure}

    \caption{
    Metric values for four features across fields. We observe strong variation for some features (e.g., specificity) and less variation for others (e.g., formality), perhaps due to our focus on computer science fields.}
    \label{fig:all-paper-metrics-appendix}
\end{figure*}

\begin{figure*}
    \centering
    \includegraphics[width=\linewidth]{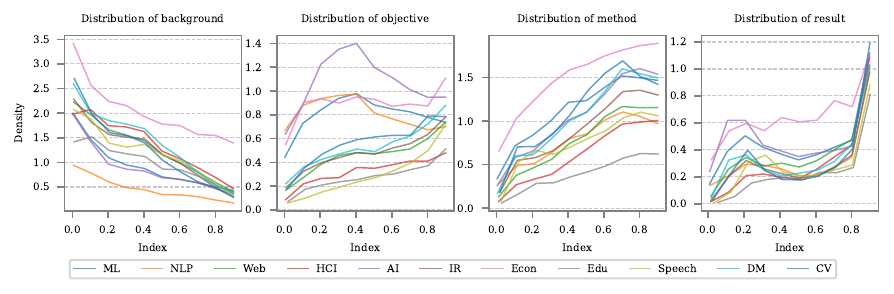}
    \caption{Positional Density of background, objective, method, and result sentences along the length of the introduction sections across communities.}
    \label{fig:scim_distribution}
\end{figure*}

Table \ref{tab:field-venue} describes the list of fields and respective venue for which we have data. Figure \ref{fig:field_counts} shows the number of papers for every field in our dataset. Figures \ref{fig:all-paper-metrics-appendix} and figure \ref{fig:scim_distribution} contain the remaining results from section \S\ref{sec:paper-ananlysis}.

\begin{table*}[h!]
\centering
\footnotesize
\begin{tabular}{ll}
\toprule
\textbf{Field} & \textbf{Venues} \\
\midrule
Machine Learning (ML) & ICLR, ICML, NeurIPS, COLT \\
Natural Language Processing (NLP) & ACL, NAACL, COLING, EMNLP, LREC, WMT \\
Web \& Recommendation Systems  & WWW, RecSys, ICWSM \\
Human Computer Interaction (HCI) & CHI, UbiComp, UIST, CSCW \\
Artificial Intelligence (AI) & AAAI, IJCAI \\
Information Retrieval (IR) & ECIR, CIKM, SIGIR \\
Economics \& Computation (Econ. / EC) & EC, WINE \\
Education (Edu) & EDM, SIGCSE, AIED, L@S \\
Speech & INTERSPEECH, ICASSP \\
Data Mining (DM) & KDD, SIGKDD, ICDM, WSDM, PAKDD \\
Computer Vision (CV) & CVPR, ECCV, ICCV \\
\bottomrule
\end{tabular}
\caption{Field and Venue Mapping}
\label{tab:field-venue}
\end{table*}

\begin{table}[!htbp]
    \centering
    \begin{tabular}{lc} 
        \toprule
        \textbf{Community} & \textbf{Quantitative Evidence Std Dev}  \\
        \midrule
        AI                      & 0.0069 \\
        ML                      & 0.0075 \\
        CV                      & 0.0079 \\
        Speech                  & 0.0091 \\
        IR                      & 0.0100 \\
        Data Mining             & 0.0109 \\
        NLP                     & 0.0101 \\
        Economics \& Computation & 0.0109 \\
        Web \& RecSys           & 0.0109 \\
        HCI                     & 0.0103 \\
        Education               & 0.0136 \\
        \bottomrule
    \end{tabular}
    
    \caption{Standard Deviation of \% sentences with quantitative evidence across fields}
    \label{table:quant-evidence-std-dev}
\end{table}

\section{LLM Adaptations}
\label{app:llm-adaptations}
\subsection{Generation Parameters}
For each of the three models, we set temperature to 0.7, top\_p to 1.0 and max output tokens to 4096. For Mistral Ministral 8B Instruct, we used Rope scaling with a factor of 2 to enable the model to handle longer context sizes. GPT 3.5 Turbo was queried between January 25-February 10, 2025 and GPT 4o Mini was queried between March 25-30, 2025. While both the open-weights models were loaded from Huggingface using vllm. The \$ cost of obtaining GPT 3.5 Turbo adaptations, including our initial experimentation was about \$500. The compute cost of generating responses from open-weights models was roughly equivalent to 5000 A6000 GPU hours.

\subsection{Results for all fields}
\label{sec:all-tables}
\input{v1/tables/all_feature_results}

\input{v1/tables/new_models_results}

Tables \ref{table:ai} to \ref{table:wrs} show the metrics for adaptations to each of the target community for GPT 3.5 Turbo, Llama 3.1 8B Instruct and Mistral Ministral 8B Instruct. Further, tables \ref{table:new-ai} to \ref{table:new-wrs} show results for larger and newer models: GPT 4o Mini annd Llama 3.3 70B Instruct.

%% file: v1/tables/all_feature_results.tex
\begin{table*}[th!]
    \centering
    \resizebox{1\linewidth}{!}{
    \footnotesize
    \begin{tabular}{llccccccccccc}
    \toprule
        \multicolumn{2}{c}{Target = "AI"}& \multicolumn{4}{c}{Baselines} & \multicolumn{2}{c}{Adapted by GPT} & \multicolumn{2}{c}{Adapted by Llama} & \multicolumn{2}{c}{Adapted by Mistral} 
        \\ \cmidrule(lr){1-2}\cmidrule(lr){3-5}\cmidrule(lr){6-7}\cmidrule(lr){8-9}\cmidrule(lr){10-11}
        feature & metric & out-comm. & in-comm. & random & specific & random & specific & random & specific & random  & specific 
		\\
    \midrule 
        \\ [-10pt]
        \multicolumn{2}{c}{Structural Norms}
        \\\cmidrule(lr){1-2}
        Length & Avg. \# words     $\uparrow$    & +612.27 & +663.26 & +656.07 & +527.93 & \cellcolor{salmon!25} -355.5 & \cellcolor{salmon!25} -266.46 & \cellcolor{salmon!25} -146.58 & \cellcolor{salmon!25} -100.15 & \cellcolor{salmon!25} -126.58 & \cellcolor{salmon!25} -72.29 \\
               & Avg. \# sentences $\uparrow$    & +28.41 & +34.93 & +30.3 & +25.48 & \cellcolor{salmon!25} -17.49 & \cellcolor{salmon!25} -14.1 & \cellcolor{salmon!25} -7.56 & \cellcolor{salmon!25} -5.88 & \cellcolor{salmon!25} -5.62 & \cellcolor{salmon!25} -3.78 \\
        Structural Artefacts & \% papers w/ tables $\downarrow$  & +7.38 & +6.96 & +6.31 & +3.9 & \cellcolor{forestgreen!25} -4.91 & \cellcolor{forestgreen!25} -2.9 & \cellcolor{forestgreen!25} -2.17 & \cellcolor{forestgreen!25} -1.6 & \cellcolor{forestgreen!25} -1.93 & \cellcolor{forestgreen!25} -1.38 \\
                            & \% papers w/ figures $\uparrow$    & +31.03 & +32.67 & +27.93 & +24.92 & \cellcolor{salmon!25} -24.73 & \cellcolor{salmon!25} -22.38 & \cellcolor{salmon!25} -8.29 & \cellcolor{salmon!25} -7.1 & \cellcolor{salmon!25} -10.15 & \cellcolor{salmon!25} -8.25 \\
	\midrule
        \\ [-10pt]
       \multicolumn{2}{c}{Stylistics Norms}
        \\\cmidrule(lr){1-2}
       Jargon      & Specificity score ($10^{-2}$) $\uparrow$    & -0.7 & +0.46 & -0.79 & +0.44 & \cellcolor{forestgreen!25} +0.12 & \cellcolor{salmon!25} -0.19 & \cellcolor{forestgreen!25} +0.21 & \cellcolor{salmon!25} -0.02 & \cellcolor{forestgreen!25} +0.18 & \cellcolor{salmon!25} -0.05 \\
       Formality   & Formality score ($10^{-2}$)   $\uparrow$    & +5.52 & +5.65 & +5.44 & +5.67 & \cellcolor{salmon!25} -0.39 & \cellcolor{salmon!25} -0.52 & \cellcolor{salmon!25} -0.14 & \cellcolor{salmon!25} -0.27 & \cellcolor{salmon!25} -0.07 & \cellcolor{salmon!25} -0.16 \\
       Readability & Flesch reading ease       $\uparrow$    & +28.83 & +31.32 & +28.42 & +27.04 & \cellcolor{salmon!25} -19.49 & \cellcolor{salmon!25} -16.89 & \cellcolor{salmon!25} -6.15 & \cellcolor{salmon!25} -5.17 & \cellcolor{salmon!25} -5.37 & \cellcolor{salmon!25} -5.36 \\
	\midrule
        \\ [-10pt]
        \multicolumn{2}{c}{Rhetorical Norms}
        \\\cmidrule(lr){1-2}
       Quant. Evidence & \% Sent. with QE     $\downarrow$  & +0.01 & -0.0 & +0.01 & -0.0 & \cellcolor{salmon!25} +0.01 & \cellcolor{salmon!25} +0.01 & \cellcolor{salmon!25} +0.04 & \cellcolor{salmon!25} +0.02 & \cellcolor{salmon!25} +0.04 & \cellcolor{salmon!25} +0.02 \\
      Framing & Cosine Sim. in values $\uparrow$ & 0.99 & base & 0.95 & 0.98 & \cellcolor{forestgreen!25} 0.0 & \cellcolor{forestgreen!25} +0.01 & \cellcolor{forestgreen!25} +0.01 & \cellcolor{forestgreen!25} +0.01 & \cellcolor{salmon!25} -0.02 & \cellcolor{salmon!25} -0.01 \\
       Narrative Organization & Background Skew $\uparrow$    & +0.51 & +0.54 & +0.44 & +0.48 & \cellcolor{forestgreen!25} +0.28 & \cellcolor{forestgreen!25} +0.23 & \cellcolor{forestgreen!25} +0.03 & \cellcolor{salmon!25} -0.01 & \cellcolor{forestgreen!25} +0.08 & \cellcolor{forestgreen!25} +0.03 \\
                              & Objective Skew   $\uparrow$    & -0.03 & +0.08 & -0.12 & -0.12 & \cellcolor{salmon!25} -0.51 & \cellcolor{salmon!25} -0.48 & \cellcolor{salmon!25} -0.11 & \cellcolor{salmon!25} -0.12 & \cellcolor{salmon!25} -0.1 & \cellcolor{salmon!25} -0.07 \\
                              & Method Skew   $\downarrow$  & -0.38 & -0.45 & -0.38 & -0.36 & \cellcolor{salmon!25} +0.16 & \cellcolor{salmon!25} +0.19 & \cellcolor{salmon!25} +0.04 & \cellcolor{salmon!25} +0.03 & \cellcolor{salmon!25} +0.08 & \cellcolor{salmon!25} +0.05 \\
                              & Result Skew   $\uparrow$    & -0.33 & +0.01 & -0.37 & -0.31 & \cellcolor{salmon!25} -0.35 & \cellcolor{salmon!25} -0.45 & \cellcolor{salmon!25} -0.01 & \cellcolor{forestgreen!25} +0.01 & \cellcolor{salmon!25} -0.07 & \cellcolor{salmon!25} -0.06 \\
            
        \bottomrule \\
    \end{tabular}
    }
    \caption{\small
        Results for the \textbf{artificial intelligence} community. The in-community column shows the metric value of papers from the community, out-community column shows the weighted average of data from all other communities. The random and specificity baselines show metric values before adaptation. The last six model columns show the \textbf{change} in value after adaptation from the random and specificity baselines, respectively. $\uparrow$ indicates that the metric should increase because the in-community value is > out-community value while $\downarrow$ indicates the vice versa. The cells where the $\Delta$ follows the expected trend are coloured {\color{forestgreen}green} while those that don't are coloured {\color{salmon}red}
    }
    \label{table:ai}
\end{table*}

\begin{table*}[th!]
    \centering
    \resizebox{1\linewidth}{!}{
    \footnotesize
    \begin{tabular}{llccccccccccc}
    \toprule
        \multicolumn{2}{c}{Target = "Computer Vision"}& \multicolumn{4}{c}{Baselines} & \multicolumn{2}{c}{Adapted by GPT} & \multicolumn{2}{c}{Adapted by Llama} & \multicolumn{2}{c}{Adapted by Mistral} 
        \\ \cmidrule(lr){1-2}\cmidrule(lr){3-5}\cmidrule(lr){6-7}\cmidrule(lr){8-9}\cmidrule(lr){10-11}
        feature & metric & out-comm. & in-comm. & random & specific & random & specific & random & specific & random  & specific 
	\\
    \midrule 
        \\ [-10pt]
        \multicolumn{2}{c}{Structural Norms}
        \\\cmidrule(lr){1-2}
        Length & Avg. \# words     $\uparrow$    & +619.97 & +655.03 & +641.11 & +596.78 & \cellcolor{salmon!25} -327.56 & \cellcolor{salmon!25} -306.96 & \cellcolor{salmon!25} -112.77 & \cellcolor{salmon!25} -97.56 & \cellcolor{salmon!25} -65.21 & \cellcolor{salmon!25} -56.17 \\
               & Avg. \# sentences $\uparrow$    & +29.47 & +32.41 & +29.7 & +28.08 & \cellcolor{salmon!25} -16.11 & \cellcolor{salmon!25} -15.62 & \cellcolor{salmon!25} -5.92 & \cellcolor{salmon!25} -5.19 & \cellcolor{salmon!25} -2.97 & \cellcolor{salmon!25} -2.52 \\
        Structural Artefacts & \% papers w/ tables $\downarrow$  & +7.37 & +5.88 & +8.02 & +3.7 & \cellcolor{forestgreen!25} -6.06 & \cellcolor{forestgreen!25} -3.3 & \cellcolor{forestgreen!25} -1.98 & \cellcolor{forestgreen!25} -1.24 & \cellcolor{forestgreen!25} -2.04 & \cellcolor{forestgreen!25} -0.84 \\
                             & \% papers w/ figures $\uparrow$    & +29.47 & +69.34 & +26.98 & +47.55 & \cellcolor{salmon!25} -22.88 & \cellcolor{salmon!25} -42.81 & \cellcolor{salmon!25} -5.2 & \cellcolor{salmon!25} -10.39 & \cellcolor{salmon!25} -7.46 & \cellcolor{salmon!25} -10.71 \\
	\midrule
        \\ [-10pt]
       \multicolumn{2}{c}{Stylistics Norms}
        \\\cmidrule(lr){1-2}
       Jargon      & Specificity score ($10^{-2}$) $\uparrow$    & -2.14 & +1.95 & -2.3 & +0.79 & \cellcolor{forestgreen!25} +1.6 & \cellcolor{forestgreen!25} +0.57 & \cellcolor{forestgreen!25} +1.54 & \cellcolor{forestgreen!25} +0.45 & \cellcolor{forestgreen!25} +1.32 & \cellcolor{forestgreen!25} +0.42 \\
       Formality   & Formality score ($10^{-2}$)   $\downarrow$  & +5.55 & +5.48 & +5.49 & +5.53 & \cellcolor{forestgreen!25} -0.43 & \cellcolor{forestgreen!25} -0.48 & \cellcolor{forestgreen!25} -0.15 & \cellcolor{forestgreen!25} -0.23 & \cellcolor{forestgreen!25} -0.08 & \cellcolor{forestgreen!25} -0.1 \\
       Readability & Flesch reading ease       $\downarrow$  & +29.44 & +26.1 & +28.52 & +25.14 & \cellcolor{forestgreen!25} -19.39 & \cellcolor{forestgreen!25} -17.14 & \cellcolor{forestgreen!25} -4.54 & \cellcolor{forestgreen!25} -3.93 & \cellcolor{forestgreen!25} -5.7 & \cellcolor{forestgreen!25} -46.89 \\
	\midrule
        \\ [-10pt]
        \multicolumn{2}{c}{Rhetorical Norms}
        \\\cmidrule(lr){1-2}
       Quant. Evidence & \% Sent. with QE     $\downarrow$  & +0.01 & +0.01 & +0.01 & +0.01 & \cellcolor{salmon!25} +0.01 & \cellcolor{salmon!25} +0.01 & \cellcolor{salmon!25} +0.03 & \cellcolor{salmon!25} +0.03 & \cellcolor{salmon!25} +0.03 & \cellcolor{salmon!25} +0.03 \\
       Narrative Organization & Background Skew $\uparrow$    & +0.51 & +0.63 & +0.47 & +0.55 & \cellcolor{forestgreen!25} +0.25 & \cellcolor{forestgreen!25} +0.26 & \cellcolor{salmon!25} -0.01 & \cellcolor{salmon!25} -0.02 & \cellcolor{forestgreen!25} +0.02 & \cellcolor{forestgreen!25} +0.04 \\
                              & Objective Skew   $\downarrow$  & +0.01 & -0.25 & -0.09 & -0.12 & \cellcolor{forestgreen!25} -0.47 & \cellcolor{forestgreen!25} -0.47 & \cellcolor{forestgreen!25} -0.15 & \cellcolor{forestgreen!25} -0.15 & \cellcolor{forestgreen!25} -0.09 & \cellcolor{forestgreen!25} -0.08 \\
                              & Method Skew   $\downarrow$  & -0.39 & -0.4 & -0.39 & -0.43 & \cellcolor{salmon!25} +0.15 & \cellcolor{salmon!25} +0.2 & \cellcolor{salmon!25} +0.04 & \cellcolor{salmon!25} +0.04 & \cellcolor{salmon!25} +0.09 & \cellcolor{salmon!25} +0.07 \\
                              & Result Skew   $\downarrow$  & -0.24 & -0.47 & -0.28 & -0.41 & \cellcolor{forestgreen!25} -0.35 & \cellcolor{forestgreen!25} -0.32 & \cellcolor{salmon!25} +0.04 & \cellcolor{salmon!25} +0.1 & \cellcolor{forestgreen!25} -0.02 & \cellcolor{forestgreen!25} -0.02 \\

        \bottomrule \\
    \end{tabular}
    }
    \caption{\small
        Results for the \textbf{computer vision} community. The in-community column shows the metric value of papers from the community, out-community column shows the weighted average of data from all other communities. The random and specificity baselines show metric values before adaptation. The last six model columns show the \textbf{change} in value after adaptation from the random and specificity baselines, respectively. $\uparrow$ indicates that the metric should increase because the in-community value is > out-community value while $\downarrow$ indicates the vice versa. The cells where the $\Delta$ follows the expected trend are coloured {\color{forestgreen}green} while those that don't are coloured {\color{salmon}red}
    }
    \label{table:cv}
\end{table*}

\begin{table*}[th!]
    \centering
    \resizebox{1\linewidth}{!}{
    \footnotesize
    \begin{tabular}{llccccccccccc}
    \toprule
        \multicolumn{2}{c}{Target = "Data Mining"}& \multicolumn{4}{c}{Baselines} & \multicolumn{2}{c}{Adapted by GPT} & \multicolumn{2}{c}{Adapted by Llama} & \multicolumn{2}{c}{Adapted by Mistral} 
        \\ \cmidrule(lr){1-2}\cmidrule(lr){3-5}\cmidrule(lr){6-7}\cmidrule(lr){8-9}\cmidrule(lr){10-11}
        feature & metric & out-comm. & in-comm. & random & specific & random & specific & random & specific & random  & specific 
	\\
    \midrule 
        \\ [-10pt]
        \multicolumn{2}{c}{Structural Norms}
        \\\cmidrule(lr){1-2}
        Length & Avg. \# words     $\uparrow$    & +618.55 & +690.88 & +644.89 & +517.61 & \cellcolor{salmon!25} -342.95 & \cellcolor{salmon!25} -252.64 & \cellcolor{salmon!25} -142.88 & \cellcolor{salmon!25} -129.13 & \cellcolor{salmon!25} -100.91 & \cellcolor{salmon!25} -90.01 \\
               & Avg. \# sentences $\uparrow$    & +29.46 & +33.02 & +30.29 & +25.0 & \cellcolor{salmon!25} -17.27 & \cellcolor{salmon!25} -13.46 & \cellcolor{salmon!25} -7.56 & \cellcolor{salmon!25} -7.16 & \cellcolor{salmon!25} -4.02 & \cellcolor{salmon!25} -1.65 \\

        Structural Artefacts & \% papers w/ tables $\uparrow$    & +7.3 & +7.34 & +6.5 & +3.3 & \cellcolor{salmon!25} -5.22 & \cellcolor{salmon!25} -2.86 & \cellcolor{salmon!25} -2.42 & \cellcolor{salmon!25} -0.88 & \cellcolor{salmon!25} -2.38 & \cellcolor{salmon!25} -0.9 \\
                            & \% papers w/ figures $\uparrow$    & +30.99 & +38.95 & +26.7 & +28.13 & \cellcolor{salmon!25} -23.64 & \cellcolor{salmon!25} -24.55 & \cellcolor{salmon!25} -8.56 & \cellcolor{salmon!25} -9.47 & \cellcolor{salmon!25} -8.68 & \cellcolor{salmon!25} -9.35 \\

               \midrule
        \\ [-10pt]
       \multicolumn{2}{c}{Stylistics Norms}
        \\\cmidrule(lr){1-2}
       Jargon      & Specificity score ($10^{-2}$) $\uparrow$    & -1.25 & +0.82 & -0.9 & +0.96 & \cellcolor{forestgreen!25} +0.87 & \cellcolor{forestgreen!25} +0.17 & \cellcolor{forestgreen!25} +0.63 & \cellcolor{forestgreen!25} +0.09 & \cellcolor{forestgreen!25} +0.65 & \cellcolor{forestgreen!25} +0.1 \\
       Formality   & Formality score ($10^{-2}$)   $\downarrow$  & +5.55 & +5.47 & +5.54 & +5.67 & \cellcolor{forestgreen!25} -0.56 & \cellcolor{forestgreen!25} -0.67 & \cellcolor{forestgreen!25} -0.25 & \cellcolor{forestgreen!25} -0.42 & \cellcolor{forestgreen!25} -0.18 & \cellcolor{forestgreen!25} -0.24 \\
       Readability & Flesch reading ease       $\downarrow$  & +29.4 & +26.72 & +28.43 & +27.73 & \cellcolor{forestgreen!25} -18.49 & \cellcolor{forestgreen!25} -17.39 & \cellcolor{forestgreen!25} -5.77 & \cellcolor{forestgreen!25} -8.42 & \cellcolor{forestgreen!25} -4.95 & \cellcolor{forestgreen!25} -6.15 \\
	\midrule
        \\ [-10pt]
        \multicolumn{2}{c}{Rhetorical Norms}
        \\\cmidrule(lr){1-2}
       Quant. Evidence & \% Sent. with QE     $\uparrow$    & +0.01 & +0.01 & +0.01 & +0.01 & \cellcolor{forestgreen!25} +0.01 & \cellcolor{forestgreen!25} +0.02 & \cellcolor{forestgreen!25} +0.03 & \cellcolor{forestgreen!25} +0.04 & \cellcolor{forestgreen!25} +0.03 & \cellcolor{forestgreen!25} +0.03 \\
       Narrative Organization & Background Skew $\downarrow$  & +0.52 & +0.49 & +0.44 & +0.42 & \cellcolor{salmon!25} +0.25 & \cellcolor{salmon!25} +0.27 & \cellcolor{salmon!25} +0.02 & \cellcolor{salmon!25} +0.06 & \cellcolor{salmon!25} +0.1 & \cellcolor{salmon!25} +0.09 \\
                              & Objective Skew   $\downarrow$  & +0.01 & -0.29 & -0.11 & -0.19 & \cellcolor{forestgreen!25} -0.5 & \cellcolor{forestgreen!25} -0.5 & \cellcolor{forestgreen!25} -0.1 & \cellcolor{forestgreen!25} -0.18 & \cellcolor{forestgreen!25} -0.07 & \cellcolor{salmon!25} +0.01 \\
                              & Method Skew   $\downarrow$  & -0.39 & -0.46 & -0.36 & -0.44 & \cellcolor{salmon!25} +0.14 & \cellcolor{salmon!25} +0.18 & \cellcolor{salmon!25} +0.03 & \cellcolor{salmon!25} +0.04 & \cellcolor{salmon!25} +0.06 & \cellcolor{salmon!25} +0.07 \\
                              & Result Skew   $\downarrow$  & -0.25 & -0.37 & -0.28 & -0.4 & \cellcolor{forestgreen!25} -0.44 & \cellcolor{forestgreen!25} -0.34 & \cellcolor{forestgreen!25} -0.02 & \cellcolor{forestgreen!25} -0.07 & \cellcolor{forestgreen!25} -0.09 & \cellcolor{forestgreen!25} -0.12 \\
        \bottomrule \\
    \end{tabular}
    }
    \caption{\small
        Results for the \textbf{data mining} community. The in-community column shows the metric value of papers from the community, out-community column shows the weighted average of data from all other communities. The random and specificity baselines show metric values before adaptation. The last six model columns show the \textbf{change} in value after adaptation from the random and specificity baselines, respectively. $\uparrow$ indicates that the metric should increase because the in-community value is > out-community value while $\downarrow$ indicates the vice versa. The cells where the $\Delta$ follows the expected trend are coloured {\color{forestgreen}green} while those that don't are coloured {\color{salmon}red}
    }
    \label{table:dm}
\end{table*}

\begin{table*}[th!]
    \centering
    \resizebox{1\linewidth}{!}{
    \footnotesize
    \begin{tabular}{llccccccccccc}
    \toprule
        \multicolumn{2}{c}{Target = "Economics and Computation"}& \multicolumn{4}{c}{Baselines} & \multicolumn{2}{c}{Adapted by GPT} & \multicolumn{2}{c}{Adapted by Llama} & \multicolumn{2}{c}{Adapted by Mistral} 
        \\ \cmidrule(lr){1-2}\cmidrule(lr){3-5}\cmidrule(lr){6-7}\cmidrule(lr){8-9}\cmidrule(lr){10-11}
        feature & metric & out-comm. & in-comm. & random & specific & random & specific & random & specific & random  & specific 
        \\
    \midrule 
        \\ [-10pt]
        \multicolumn{2}{c}{Structural Norms}
        \\\cmidrule(lr){1-2}
        Length & Avg. \# words     $\uparrow$    & +616.79 & +1107.37 & +619.48 & +613.61 & \cellcolor{salmon!25} -299.83 & \cellcolor{salmon!25} -341.15 & \cellcolor{salmon!25} -186.6 & \cellcolor{salmon!25} -251.84 & \cellcolor{salmon!25} -104.61 & \cellcolor{salmon!25} -119.69 \\
               & Avg. \# sentences $\uparrow$    & +29.4 & +50.67 & +29.64 & +28.46 & \cellcolor{salmon!25} -15.94 & \cellcolor{salmon!25} -16.85 & \cellcolor{salmon!25} -10.32 & \cellcolor{salmon!25} -12.58 & \cellcolor{salmon!25} -5.21 & \cellcolor{salmon!25} -4.01 \\
        Structural Artefacts & \% papers w/ tables $\downarrow$  & +7.33 & +4.77 & +6.5 & +5.01 & \cellcolor{forestgreen!25} -5.36 & \cellcolor{forestgreen!25} -4.29 & \cellcolor{forestgreen!25} -2.64 & \cellcolor{forestgreen!25} -2.93 & \cellcolor{forestgreen!25} -2.18 & \cellcolor{forestgreen!25} -1.57 \\
                            & \% papers w/ figures $\downarrow$  & +31.54 & +10.29 & +30.3 & +19.04 & \cellcolor{forestgreen!25} -26.14 & \cellcolor{forestgreen!25} -15.06 & \cellcolor{forestgreen!25} -13.84 & \cellcolor{forestgreen!25} -10.88 & \cellcolor{forestgreen!25} -11.0 & \cellcolor{forestgreen!25} -7.13 \\

               \midrule
        \\ [-10pt]
       \multicolumn{2}{c}{Stylistics Norms}
        \\\cmidrule(lr){1-2}
       Jargon      & Specificity score ($10^{-2}$) $\uparrow$    & -3.59 & +2.94 & -3.28 & +0.76 & \cellcolor{salmon!25} -0.02 & \cellcolor{salmon!25} -0.54 & \cellcolor{forestgreen!25} +0.52 & \cellcolor{salmon!25} -0.06 & \cellcolor{forestgreen!25} +0.59 & \cellcolor{forestgreen!25} +0.01 \\
       Formality   & Formality score ($10^{-2}$)   $\downarrow$  & +5.55 & +5.51 & +5.51 & +5.65 & \cellcolor{forestgreen!25} -0.57 & \cellcolor{forestgreen!25} -0.67 & \cellcolor{forestgreen!25} -0.33 & \cellcolor{forestgreen!25} -0.46 & \cellcolor{forestgreen!25} -0.19 & \cellcolor{forestgreen!25} -0.26 \\
       Readability & Flesch reading ease       $\uparrow$    & +29.25 & +32.95 & +28.07 & +31.19 & \cellcolor{salmon!25} -25.3 & \cellcolor{salmon!25} -25.06 & \cellcolor{salmon!25} -13.03 & \cellcolor{salmon!25} -13.06 & \cellcolor{salmon!25} -8.9 & \cellcolor{salmon!25} -7.55 \\
	\midrule
        \\ [-10pt]
        \multicolumn{2}{c}{Rhetorical Norms}
        \\\cmidrule(lr){1-2}
       Quant. Evidence & \% Sent. with QE     $\uparrow$    & +0.01 & +0.01 & +0.01 & +0.01 & \cellcolor{forestgreen!25} +0.01 & \cellcolor{forestgreen!25} +0.02 & \cellcolor{forestgreen!25} +0.03 & \cellcolor{forestgreen!25} +0.05 & \cellcolor{forestgreen!25} +0.03 & \cellcolor{forestgreen!25} +0.04 \\
       Narrative Organization & Background Skew $\downarrow$  & +0.52 & +0.29 & +0.5 & +0.34 & \cellcolor{salmon!25} +0.21 & \cellcolor{salmon!25} +0.28 & \cellcolor{salmon!25} +0.07 & \cellcolor{salmon!25} +0.12 & \cellcolor{salmon!25} +0.09 & \cellcolor{salmon!25} +0.08 \\
                              & Objective Skew   $\downarrow$  & -0.0 & -0.0 & -0.12 & -0.11 & \cellcolor{forestgreen!25} -0.5 & \cellcolor{forestgreen!25} -0.47 & \cellcolor{forestgreen!25} -0.23 & \cellcolor{forestgreen!25} -0.27 & \cellcolor{forestgreen!25} -0.11 & \cellcolor{forestgreen!25} -0.06 \\
                              & Method Skew   $\uparrow$    & -0.4 & -0.22 & -0.42 & -0.36 & \cellcolor{forestgreen!25} +0.18 & \cellcolor{forestgreen!25} +0.11 & \cellcolor{forestgreen!25} +0.01 & \cellcolor{salmon!25} -0.0 & \cellcolor{forestgreen!25} +0.07 & \cellcolor{forestgreen!25} +0.04 \\
                              & Result Skew   $\uparrow$    & -0.25 & -0.17 & -0.32 & -0.3 & \cellcolor{salmon!25} -0.31 & \cellcolor{salmon!25} -0.41 & \cellcolor{salmon!25} -0.0 & \cellcolor{salmon!25} -0.18 & \cellcolor{salmon!25} -0.07 & \cellcolor{salmon!25} -0.14 \\
        
        \bottomrule \\
    \end{tabular}
    }
    \caption{\small
        Results for the \textbf{economics and computation} community. The in-community column shows the metric value of papers from the community, out-community column shows the weighted average of data from all other communities. The random and specificity baselines show metric values before adaptation. The last six model columns show the \textbf{change} in value after adaptation from the random and specificity baselines, respectively. $\uparrow$ indicates that the metric should increase because the in-community value is > out-community value while $\downarrow$ indicates the vice versa. The cells where the $\Delta$ follows the expected trend are coloured {\color{forestgreen}green} while those that don't are coloured {\color{salmon}red}
    }
    \label{table:ec}
\end{table*}

\begin{table*}[th!]
    \centering
    \resizebox{1\linewidth}{!}{
    \footnotesize
    \begin{tabular}{llccccccccccc}
    \toprule
        \multicolumn{2}{c}{Target = "Education"}& \multicolumn{4}{c}{Baselines} & \multicolumn{2}{c}{Adapted by GPT} & \multicolumn{2}{c}{Adapted by Llama} & \multicolumn{2}{c}{Adapted by Mistral} 
        \\ \cmidrule(lr){1-2}\cmidrule(lr){3-5}\cmidrule(lr){6-7}\cmidrule(lr){8-9}\cmidrule(lr){10-11}
        feature & metric & out-comm. & in-comm. & random & specific & random & specific & random & specific & random  & specific 
	\\
    \midrule 
        \\ [-10pt]
        \multicolumn{2}{c}{Structural Norms}
        \\\cmidrule(lr){1-2}
        Length & Avg. \# words     $\downarrow$  & +624.59 & +422.89 & +687.76 & +488.39 & \cellcolor{forestgreen!25} -362.71 & \cellcolor{forestgreen!25} -225.3 & \cellcolor{forestgreen!25} -137.93 & \cellcolor{forestgreen!25} -115.43 & \cellcolor{forestgreen!25} -128.25 & \cellcolor{forestgreen!25} -78.98 \\
               & Avg. \# sentences $\downarrow$  & +29.77 & +18.8 & +33.04 & +22.77 & \cellcolor{forestgreen!25} -18.97 & \cellcolor{forestgreen!25} -11.57 & \cellcolor{forestgreen!25} -8.2 & \cellcolor{forestgreen!25} -6.3 & \cellcolor{forestgreen!25} -6.53 & \cellcolor{forestgreen!25} -0.64 \\
        Structural Artefacts & \% papers w/ tables $\downarrow$  & +7.37 & +2.84 & +6.51 & +4.31 & \cellcolor{forestgreen!25} -5.39 & \cellcolor{forestgreen!25} -3.81 & \cellcolor{forestgreen!25} -2.11 & \cellcolor{forestgreen!25} -1.91 & \cellcolor{forestgreen!25} -2.53 & \cellcolor{forestgreen!25} -1.43 \\
                             & \% papers w/ figures $\downarrow$  & +31.7 & +6.34 & +30.56 & +17.64 & \cellcolor{forestgreen!25} -26.04 & \cellcolor{forestgreen!25} -14.14 & \cellcolor{forestgreen!25} -9.06 & \cellcolor{forestgreen!25} -5.9 & \cellcolor{forestgreen!25} -12.28 & \cellcolor{forestgreen!25} -6.1 \\
    
       \midrule
        \\ [-10pt]
       \multicolumn{2}{c}{Stylistics Norms}
        \\\cmidrule(lr){1-2}
       Jargon      & Specificity score ($10^{-2}$) $\uparrow$    & -4.46 & +4.46 & -3.97 & +0.24 & \cellcolor{forestgreen!25} +3.04 & \cellcolor{forestgreen!25} +1.84 & \cellcolor{forestgreen!25} +2.07 & \cellcolor{forestgreen!25} +1.36 & \cellcolor{forestgreen!25} +1.82 & \cellcolor{forestgreen!25} +1.24 \\
       Formality   & Formality score ($10^{-2}$)   $\downarrow$  & +5.55 & +5.39 & +5.52 & +5.87 & \cellcolor{forestgreen!25} -0.65 & \cellcolor{forestgreen!25} -0.95 & \cellcolor{forestgreen!25} -0.32 & \cellcolor{forestgreen!25} -0.67 & \cellcolor{forestgreen!25} -0.25 & \cellcolor{forestgreen!25} -0.44 \\
       Readability & Flesch reading ease       $\downarrow$  & +29.35 & +25.31 & +28.64 & +27.3 & \cellcolor{forestgreen!25} -26.23 & \cellcolor{forestgreen!25} -23.77 & \cellcolor{forestgreen!25} -8.94 & \cellcolor{forestgreen!25} -12.05 & \cellcolor{forestgreen!25} -9.36 & \cellcolor{forestgreen!25} -10.28 \\
	\midrule
        \\ [-10pt]
        \multicolumn{2}{c}{Rhetorical Norms}
        \\\cmidrule(lr){1-2}
       Quant. Evidence & \% Sent. with QE     $\uparrow$    & +0.01 & +0.01 & +0.01 & +0.01 & \cellcolor{forestgreen!25} +0.01 & \cellcolor{forestgreen!25} +0.01 & \cellcolor{forestgreen!25} +0.03 & \cellcolor{forestgreen!25} +0.04 & \cellcolor{forestgreen!25} +0.03 & \cellcolor{forestgreen!25} +0.03 \\
       Narrative Organization & Background Skew $\downarrow$  & +0.52 & +0.41 & +0.47 & +0.38 & \cellcolor{salmon!25} +0.23 & \cellcolor{salmon!25} +0.26 & \cellcolor{salmon!25} +0.01 & \cellcolor{salmon!25} +0.07 & \cellcolor{salmon!25} +0.09 & \cellcolor{salmon!25} +0.09 \\
                              & Objective Skew   $\downarrow$  & +0.01 & -0.33 & -0.11 & -0.11 & \cellcolor{forestgreen!25} -0.49 & \cellcolor{forestgreen!25} -0.44 & \cellcolor{forestgreen!25} -0.14 & \cellcolor{forestgreen!25} -0.15 & \cellcolor{forestgreen!25} -0.11 & \cellcolor{forestgreen!25} -0.02 \\
                              & Method Skew   $\downarrow$  & -0.39 & -0.45 & -0.37 & -0.43 & \cellcolor{salmon!25} +0.12 & \cellcolor{salmon!25} +0.05 & \cellcolor{forestgreen!25} -0.0 & \cellcolor{forestgreen!25} -0.01 & \cellcolor{salmon!25} +0.05 & \cellcolor{salmon!25} +0.03 \\
                              & Result Skew   $\downarrow$  & -0.25 & -0.54 & -0.28 & -0.39 & \cellcolor{forestgreen!25} -0.3 & \cellcolor{forestgreen!25} -0.45 & \cellcolor{salmon!25} +0.07 & \cellcolor{forestgreen!25} -0.09 & \cellcolor{forestgreen!25} -0.08 & \cellcolor{forestgreen!25} -0.11 \\
                
        \bottomrule \\
    \end{tabular}
    }
    \caption{\small
        Results for the \textbf{education} community. The in-community column shows the metric value of papers from the community, out-community column shows the weighted average of data from all other communities. The random and specificity baselines show metric values before adaptation. The last six model columns show the \textbf{change} in value after adaptation from the random and specificity baselines, respectively. $\uparrow$ indicates that the metric should increase because the in-community value is > out-community value while $\downarrow$ indicates the vice versa. The cells where the $\Delta$ follows the expected trend are coloured {\color{forestgreen}green} while those that don't are coloured {\color{salmon}red}
    }
    \label{table:education}
\end{table*}

\begin{table*}[th!]
    \centering
    \resizebox{1\linewidth}{!}{
    \footnotesize
    \begin{tabular}{llccccccccccc}
    \toprule
    \multicolumn{2}{c}{Target = "Human Computer Interaction"}& \multicolumn{4}{c}{Baselines} & \multicolumn{2}{c}{Adapted by GPT} & \multicolumn{2}{c}{Adapted by Llama} & \multicolumn{2}{c}{Adapted by Mistral} 
    \\ \cmidrule(lr){1-2}\cmidrule(lr){3-5}\cmidrule(lr){6-7}\cmidrule(lr){8-9}\cmidrule(lr){10-11}
    feature & metric & out-comm. & in-comm. & random & specific & random & specific & random & specific & random  & specific 
	\\
    \midrule 
        \\ [-10pt]
        \multicolumn{2}{c}{Structural Norms}
        \\\cmidrule(lr){1-2}
        Length & Avg. \# words     $\downarrow$  & +622.85 & +605.04 & +656.23 & +474.86 & \cellcolor{forestgreen!25} -348.59 & \cellcolor{forestgreen!25} -225.22 & \cellcolor{forestgreen!25} -224.94 & \cellcolor{forestgreen!25} -171.67 & \cellcolor{forestgreen!25} -114.74 & \cellcolor{forestgreen!25} -82.85 \\
               & Avg. \# sentences $\downarrow$  & +29.89 & +25.86 & +31.06 & +22.18 & \cellcolor{forestgreen!25} -17.81 & \cellcolor{forestgreen!25} -11.61 & \cellcolor{forestgreen!25} -11.84 & \cellcolor{forestgreen!25} -8.89 & \cellcolor{forestgreen!25} -5.52 & \cellcolor{forestgreen!25} -1.16 \\
        Structural Artefacts & \% papers w/ tables $\downarrow$  & +7.54 & +4.1 & +6.4 & +3.41 & \cellcolor{forestgreen!25} -4.9 & \cellcolor{forestgreen!25} -2.79 & \cellcolor{forestgreen!25} -2.98 & \cellcolor{forestgreen!25} -1.63 & \cellcolor{forestgreen!25} -2.24 & \cellcolor{forestgreen!25} -1.2 \\
                             & \% papers w/ figures $\uparrow$    & +31.25 & +32.37 & +29.5 & +18.04 & \cellcolor{salmon!25} -24.48 & \cellcolor{salmon!25} -14.6 & \cellcolor{salmon!25} -12.86 & \cellcolor{salmon!25} -8.28 & \cellcolor{salmon!25} -10.36 & \cellcolor{salmon!25} -5.81 \\
       \midrule
        \\ [-10pt]
       \multicolumn{2}{c}{Stylistics Norms}
        \\\cmidrule(lr){1-2}
       Jargon      & Specificity score ($10^{-2}$) $\uparrow$    & -3.49 & +3.18 & -2.54 & +1.74 & \cellcolor{forestgreen!25} +2.56 & \cellcolor{forestgreen!25} +1.5 & \cellcolor{forestgreen!25} +1.14 & \cellcolor{forestgreen!25} +0.73 & \cellcolor{forestgreen!25} +1.45 & \cellcolor{forestgreen!25} +0.79 \\
       Formality   & Formality score ($10^{-2}$)   $\downarrow$  & +5.57 & +5.26 & +5.5 & +5.67 & \cellcolor{forestgreen!25} -0.4 & \cellcolor{forestgreen!25} -0.63 & \cellcolor{forestgreen!25} -0.24 & \cellcolor{forestgreen!25} -0.51 & \cellcolor{forestgreen!25} -0.1 & \cellcolor{forestgreen!25} -0.24 \\
       Readability & Flesch reading ease       $\downarrow$  & +29.65 & +24.4 & +27.61 & +25.51 & \cellcolor{forestgreen!25} -20.37 & \cellcolor{forestgreen!25} -19.96 & \cellcolor{forestgreen!25} -9.53 & \cellcolor{forestgreen!25} -12.66 & \cellcolor{forestgreen!25} -5.92 & \cellcolor{forestgreen!25} -6.55 \\
	\midrule
        \\ [-10pt]
        \multicolumn{2}{c}{Rhetorical Norms}
        \\\cmidrule(lr){1-2}
       Quant. Evidence & \% Sent. with QE     $\uparrow$    & +0.01 & +0.01 & +0.01 & +0.01 & \cellcolor{forestgreen!25} +0.01 & \cellcolor{forestgreen!25} +0.02 & \cellcolor{forestgreen!25} +0.03 & \cellcolor{forestgreen!25} +0.04 & \cellcolor{forestgreen!25} +0.03 & \cellcolor{forestgreen!25} +0.03 \\
       Narrative Organization & Background Skew $\downarrow$  & +0.52 & +0.45 & +0.49 & +0.35 & \cellcolor{salmon!25} +0.25 & \cellcolor{salmon!25} +0.31 & \cellcolor{salmon!25} +0.07 & \cellcolor{salmon!25} +0.14 & \cellcolor{salmon!25} +0.1 & \cellcolor{salmon!25} +0.1 \\
                              & Objective Skew   $\downarrow$  & +0.01 & -0.26 & -0.08 & -0.16 & \cellcolor{forestgreen!25} -0.53 & \cellcolor{forestgreen!25} -0.4 & \cellcolor{forestgreen!25} -0.25 & \cellcolor{forestgreen!25} -0.23 & \cellcolor{forestgreen!25} -0.15 & \cellcolor{forestgreen!25} -0.01 \\
                              & Method Skew   $\downarrow$  & -0.39 & -0.56 & -0.39 & -0.39 & \cellcolor{salmon!25} +0.22 & \cellcolor{salmon!25} +0.09 & \cellcolor{salmon!25} +0.03 & \cellcolor{forestgreen!25} -0.05 & \cellcolor{salmon!25} +0.08 & \cellcolor{salmon!25} +0.02 \\
                              & Result Skew   $\downarrow$  & -0.22 & -0.7 & -0.27 & -0.4 & \cellcolor{forestgreen!25} -0.37 & \cellcolor{forestgreen!25} -0.45 & \cellcolor{forestgreen!25} -0.07 & \cellcolor{forestgreen!25} -0.26 & \cellcolor{forestgreen!25} -0.11 & \cellcolor{forestgreen!25} -0.14 \\
        
        \bottomrule \\
    \end{tabular}
    }
    \caption{\small
        Results for the \textbf{human computer interaction} community. The in-community column shows the metric value of papers from the community, out-community column shows the weighted average of data from all other communities. The random and specificity baselines show metric values before adaptation. The last six model columns show the \textbf{change} in value after adaptation from the random and specificity baselines, respectively. $\uparrow$ indicates that the metric should increase because the in-community value is > out-community value while $\downarrow$ indicates the vice versa. The cells where the $\Delta$ follows the expected trend are coloured {\color{forestgreen}green} while those that don't are coloured {\color{salmon}red}
    }
    \label{table:hci}
\end{table*}

\begin{table*}[th!]
    \centering
    \resizebox{1\linewidth}{!}{
    \footnotesize
    \begin{tabular}{llccccccccccc}
    \toprule
    \multicolumn{2}{c}{Target = "Information Retrieval"}& \multicolumn{4}{c}{Baselines} & \multicolumn{2}{c}{Adapted by GPT} & \multicolumn{2}{c}{Adapted by Llama} & \multicolumn{2}{c}{Adapted by Mistral} 
    \\ \cmidrule(lr){1-2}\cmidrule(lr){3-5}\cmidrule(lr){6-7}\cmidrule(lr){8-9}\cmidrule(lr){10-11}
    feature & metric & out-comm. & in-comm. & random & specific & random & specific & random & specific & random  & specific 
		\\
    \midrule 
        \\ [-10pt]
        \multicolumn{2}{c}{Structural Norms}
        \\\cmidrule(lr){1-2}
        Length & Avg. \# words     $\downarrow$  & +622.39 & +610.41 & +654.45 & +555.38 & \cellcolor{forestgreen!25} -346.41 & \cellcolor{forestgreen!25} -276.17 & \cellcolor{forestgreen!25} -151.67 & \cellcolor{forestgreen!25} -123.56 & \cellcolor{forestgreen!25} -88.32 & \cellcolor{forestgreen!25} -64.03 \\
               & Avg. \# sentences $\downarrow$  & +29.64 & +29.06 & +30.34 & +27.0 & \cellcolor{forestgreen!25} -17.23 & \cellcolor{forestgreen!25} -14.81 & \cellcolor{forestgreen!25} -7.95 & \cellcolor{forestgreen!25} -7.22 & \cellcolor{forestgreen!25} -3.77 & \cellcolor{salmon!25} +1.79 \\
        Structural Artefacts & \% papers w/ tables $\uparrow$    & +7.26 & +7.85 & +5.41 & +5.61 & \cellcolor{salmon!25} -4.39 & \cellcolor{salmon!25} -4.37 & \cellcolor{salmon!25} -1.73 & \cellcolor{salmon!25} -1.17 & \cellcolor{salmon!25} -1.37 & \cellcolor{salmon!25} -1.43 \\
                            & \% papers w/ figures $\uparrow$    & +31.19 & +33.37 & +28.13 & +30.33 & \cellcolor{salmon!25} -24.13 & \cellcolor{salmon!25} -25.03 & \cellcolor{salmon!25} -7.79 & \cellcolor{salmon!25} -7.91 & \cellcolor{salmon!25} -8.41 & \cellcolor{salmon!25} -8.21 \\
	\midrule
        \\ [-10pt]
       \multicolumn{2}{c}{Stylistics Norms}
        \\\cmidrule(lr){1-2}
       Jargon      & Specificity score ($10^{-2}$) $\uparrow$    & -1.26 & +1.14 & -0.91 & +1.27 & \cellcolor{forestgreen!25} +0.91 & \cellcolor{forestgreen!25} +0.3 & \cellcolor{forestgreen!25} +0.95 & \cellcolor{forestgreen!25} +0.2 & \cellcolor{forestgreen!25} +1.09 & \cellcolor{forestgreen!25} +0.36 \\
       Formality   & Formality score ($10^{-2}$)   $\downarrow$  & +5.55 & +5.49 & +5.46 & +5.86 & \cellcolor{forestgreen!25} -0.45 & \cellcolor{forestgreen!25} -0.76 & \cellcolor{forestgreen!25} -0.19 & \cellcolor{forestgreen!25} -0.54 & \cellcolor{forestgreen!25} -0.07 & \cellcolor{forestgreen!25} -0.3 \\
       Readability & Flesch reading ease       $\downarrow$  & +29.49 & +26.39 & +27.54 & +28.75 & \cellcolor{forestgreen!25} -20.48 & \cellcolor{forestgreen!25} -20.13 & \cellcolor{forestgreen!25} -6.33 & \cellcolor{forestgreen!25} -8.32 & \cellcolor{forestgreen!25} -4.9 & \cellcolor{forestgreen!25} -4.79 \\
	\midrule
        \\ [-10pt]
        \multicolumn{2}{c}{Rhetorical Norms}
        \\\cmidrule(lr){1-2}
       Quant. Evidence & \% Sent. with QE     $\uparrow$    & +0.01 & +0.01 & +0.01 & +0.01 & \cellcolor{forestgreen!25} +0.01 & \cellcolor{forestgreen!25} +0.01 & \cellcolor{forestgreen!25} +0.03 & \cellcolor{forestgreen!25} +0.03 & \cellcolor{forestgreen!25} +0.03 & \cellcolor{forestgreen!25} +0.03 \\
       Narrative Organization & Background Skew $\downarrow$  & +0.52 & +0.48 & +0.46 & +0.46 & \cellcolor{salmon!25} +0.24 & \cellcolor{salmon!25} +0.25 & \cellcolor{salmon!25} +0.04 & \cellcolor{salmon!25} +0.07 & \cellcolor{salmon!25} +0.05 & \cellcolor{salmon!25} +0.08 \\
                              & Objective Skew   $\downarrow$  & +0.01 & -0.27 & -0.07 & -0.11 & \cellcolor{forestgreen!25} -0.57 & \cellcolor{forestgreen!25} -0.49 & \cellcolor{forestgreen!25} -0.2 & \cellcolor{forestgreen!25} -0.24 & \cellcolor{forestgreen!25} -0.13 & \cellcolor{forestgreen!25} -0.01 \\
                              & Method Skew   $\downarrow$  & -0.39 & -0.46 & -0.4 & -0.42 & \cellcolor{salmon!25} +0.13 & \cellcolor{salmon!25} +0.11 & \cellcolor{salmon!25} +0.04 & \cellcolor{salmon!25} +0.02 & \cellcolor{salmon!25} +0.11 & \cellcolor{salmon!25} +0.08 \\
                              & Result Skew   $\downarrow$  & -0.24 & -0.39 & -0.3 & -0.36 & \cellcolor{forestgreen!25} -0.37 & \cellcolor{forestgreen!25} -0.35 & \cellcolor{salmon!25} +0.04 & \cellcolor{forestgreen!25} -0.02 & \cellcolor{forestgreen!25} -0.01 & \cellcolor{forestgreen!25} -0.06 \\
        
        \bottomrule \\
    \end{tabular}
    }
    \caption{\small
        Results for the \textbf{information retrieval} community. The in-community column shows the metric value of papers from the community, out-community column shows the weighted average of data from all other communities. The random and specificity baselines show metric values before adaptation. The last six model columns show the \textbf{change} in value after adaptation from the random and specificity baselines, respectively. $\uparrow$ indicates that the metric should increase because the in-community value is > out-community value while $\downarrow$ indicates the vice versa. The cells where the $\Delta$ follows the expected trend are coloured {\color{forestgreen}green} while those that don't are coloured {\color{salmon}red}
    }
    \label{table:ir}
\end{table*}

\begin{table*}[th!]
    \centering
    \resizebox{1\linewidth}{!}{
    \footnotesize
    \begin{tabular}{llccccccccccc}
    \toprule
    \multicolumn{2}{c}{Target = "Machine Learning"}& \multicolumn{4}{c}{Baselines} & \multicolumn{2}{c}{Adapted by GPT} & \multicolumn{2}{c}{Adapted by Llama} & \multicolumn{2}{c}{Adapted by Mistral} 
    \\ \cmidrule(lr){1-2}\cmidrule(lr){3-5}\cmidrule(lr){6-7}\cmidrule(lr){8-9}\cmidrule(lr){10-11}
    feature & metric & out-comm. & in-comm. & random & specific & random & specific & random & specific & random  & specific 
	\\
    \midrule 
        \\ [-10pt]
        \multicolumn{2}{c}{Structural Norms}
        \\\cmidrule(lr){1-2}
        Length & Avg. \# words     $\uparrow$    & +601.8 & +695.38 & +651.35 & +547.91 & \cellcolor{salmon!25} -363.7 & \cellcolor{salmon!25} -282.73 & \cellcolor{salmon!25} -185.82 & \cellcolor{salmon!25} -123.69 & \cellcolor{salmon!25} -106.49 & \cellcolor{salmon!25} -37.42 \\
               & Avg. \# sentences $\uparrow$    & +28.7 & +33.0 & +30.44 & +26.48 & \cellcolor{salmon!25} -18.03 & \cellcolor{salmon!25} -14.94 & \cellcolor{salmon!25} -9.54 & \cellcolor{salmon!25} -7.01 & \cellcolor{salmon!25} -4.86 & \cellcolor{salmon!25} -2.08 \\
        Structural Artefacts & \% papers w/ tables $\downarrow$  & +7.5 & +6.57 & +6.21 & +4.02 & \cellcolor{forestgreen!25} -4.87 & \cellcolor{forestgreen!25} -3.1 & \cellcolor{forestgreen!25} -2.35 & \cellcolor{forestgreen!25} -1.92 & \cellcolor{forestgreen!25} -1.91 & \cellcolor{forestgreen!25} -0.97 \\
                     & \% papers w/ figures $\downarrow$  & +31.8 & +29.58 & +28.73 & +16.08 & \cellcolor{forestgreen!25} -25.85 & \cellcolor{forestgreen!25} -14.46 & \cellcolor{forestgreen!25} -10.95 & \cellcolor{forestgreen!25} -6.56 & \cellcolor{forestgreen!25} -9.73 & \cellcolor{forestgreen!25} -4.41 \\
	\midrule
        \\ [-10pt]
       \multicolumn{2}{c}{Stylistics Norms}
        \\\cmidrule(lr){1-2}
       Jargon      & Specificity score ($10^{-2}$) $\uparrow$    & -1.66 & +1.08 & -1.69 & +0.8 & \cellcolor{forestgreen!25} +0.15 & \cellcolor{salmon!25} -0.22 & \cellcolor{forestgreen!25} +0.3 & \cellcolor{forestgreen!25} +0.04 & \cellcolor{forestgreen!25} +0.45 & \cellcolor{forestgreen!25} +0.06 \\
       Formality   & Formality score ($10^{-2}$)   $\uparrow$    & +5.53 & +5.62 & +5.5 & +5.6 & \cellcolor{salmon!25} -0.48 & \cellcolor{salmon!25} -0.52 & \cellcolor{salmon!25} -0.22 & \cellcolor{salmon!25} -0.26 & \cellcolor{salmon!25} -0.11 & \cellcolor{salmon!25} -0.12 \\
       Readability & Flesch reading ease       $\downarrow$  & +29.43 & +28.74 & +28.16 & +25.94 & \cellcolor{forestgreen!25} -18.24 & \cellcolor{forestgreen!25} -15.21 & \cellcolor{forestgreen!25} -5.92 & \cellcolor{forestgreen!25} -3.93 & \cellcolor{forestgreen!25} -4.08 & \cellcolor{forestgreen!25} -3.55 \\
	\midrule
        \\ [-10pt]
        \multicolumn{2}{c}{Rhetorical Norms}
        \\\cmidrule(lr){1-2}
       Quant. Evidence & \% Sent. with QE     $\downarrow$  & +0.01 & -0.0 & +0.01 & +0.01 & \cellcolor{salmon!25} +0.01 & \cellcolor{salmon!25} +0.01 & \cellcolor{salmon!25} +0.03 & \cellcolor{salmon!25} +0.03 & \cellcolor{salmon!25} +0.03 & \cellcolor{salmon!25} +0.03 \\
       Narrative Organization & Background Skew $\uparrow$    & +0.5 & +0.56 & +0.45 & +0.52 & \cellcolor{forestgreen!25} +0.31 & \cellcolor{forestgreen!25} +0.29 & \cellcolor{forestgreen!25} +0.05 & \cellcolor{salmon!25} -0.02 & \cellcolor{forestgreen!25} +0.09 & \cellcolor{forestgreen!25} +0.01 \\
                              & Objective Skew   $\uparrow$    & -0.0 & +0.02 & -0.09 & -0.17 & \cellcolor{salmon!25} -0.5 & \cellcolor{salmon!25} -0.4 & \cellcolor{salmon!25} -0.17 & \cellcolor{salmon!25} -0.07 & \cellcolor{salmon!25} -0.12 & \cellcolor{salmon!25} -0.04 \\
                              & Method Skew   $\uparrow$    & -0.42 & -0.31 & -0.41 & -0.33 & \cellcolor{forestgreen!25} +0.18 & \cellcolor{forestgreen!25} +0.18 & \cellcolor{forestgreen!25} +0.01 & \cellcolor{forestgreen!25} +0.01 & \cellcolor{forestgreen!25} +0.07 & \cellcolor{forestgreen!25} +0.06 \\
                              & Result Skew   $\uparrow$    & -0.27 & -0.18 & -0.34 & -0.33 & \cellcolor{salmon!25} -0.48 & \cellcolor{salmon!25} -0.53 & \cellcolor{salmon!25} -0.01 & \cellcolor{salmon!25} -0.01 & \cellcolor{salmon!25} -0.06 & \cellcolor{forestgreen!25} +0.01 \\

        \bottomrule \\
    \end{tabular}
    }
    \caption{\small
        Results for the \textbf{machine learning} community. The in-community column shows the metric value of papers from the community, out-community column shows the weighted average of data from all other communities. The random and specificity baselines show metric values before adaptation. The last six model columns show the \textbf{change} in value after adaptation from the random and specificity baselines, respectively. $\uparrow$ indicates that the metric should increase because the in-community value is > out-community value while $\downarrow$ indicates the vice versa. The cells where the $\Delta$ follows the expected trend are coloured {\color{forestgreen}green} while those that don't are coloured {\color{salmon}red} 
    }
    \label{table:ml}
\end{table*}

\begin{table*}[th!]
    \centering
    \resizebox{1\linewidth}{!}{
    \footnotesize
    \begin{tabular}{llccccccccccc}
    \toprule
    \multicolumn{2}{c}{Target = "Natural Language Processing"}& \multicolumn{4}{c}{Baselines} & \multicolumn{2}{c}{Adapted by GPT} & \multicolumn{2}{c}{Adapted by Llama} & \multicolumn{2}{c}{Adapted by Mistral} 
    \\ \cmidrule(lr){1-2}\cmidrule(lr){3-5}\cmidrule(lr){6-7}\cmidrule(lr){8-9}\cmidrule(lr){10-11}
    feature & metric & out-comm. & in-comm. & random & specific & random & specific & random & specific & random  & specific 
	\\
    \midrule 
        \\ [-10pt]
        \multicolumn{2}{c}{Structural Norms}
        \\\cmidrule(lr){1-2}
        Length & Avg. \# words     $\downarrow$  & +648.46 & +530.82 & +650.24 & +596.75 & \cellcolor{forestgreen!25} -320.43 & \cellcolor{forestgreen!25} -302.05 & \cellcolor{forestgreen!25} -115.37 & \cellcolor{forestgreen!25} -117.29 & \cellcolor{forestgreen!25} -92.45 & \cellcolor{forestgreen!25} -84.45 \\
               & Avg. \# sentences $\downarrow$  & +31.36 & +23.67 & +30.81 & +28.35 & \cellcolor{forestgreen!25} -16.66 & \cellcolor{forestgreen!25} -15.6 & \cellcolor{forestgreen!25} -6.66 & \cellcolor{forestgreen!25} -6.31 & \cellcolor{forestgreen!25} -4.18 & \cellcolor{forestgreen!25} -3.23 \\
        Structural Artefacts & \% papers w/ tables $\uparrow$    & +6.06 & +11.51 & +6.31 & +11.31 & \cellcolor{salmon!25} -5.31 & \cellcolor{salmon!25} -9.09 & \cellcolor{salmon!25} -2.11 & \cellcolor{salmon!25} -3.23 & \cellcolor{salmon!25} -2.15 & \cellcolor{salmon!25} -3.25 \\
                     & \% papers w/ figures $\uparrow$    & +31.04 & +32.31 & +29.16 & +29.43 & \cellcolor{salmon!25} -24.04 & \cellcolor{salmon!25} -24.99 & \cellcolor{salmon!25} -6.92 & \cellcolor{salmon!25} -6.05 & \cellcolor{salmon!25} -8.66 & \cellcolor{salmon!25} -7.05 \\

       \midrule
        \\ [-10pt]
       \multicolumn{2}{c}{Stylistics Norms}
        \\\cmidrule(lr){1-2}
       Jargon      & Specificity score ($10^{-2}$) $\uparrow$    & -1.58 & +1.44 & -1.7 & +0.93 & \cellcolor{forestgreen!25} +1.01 & \cellcolor{forestgreen!25} +0.31 & \cellcolor{forestgreen!25} +1.09 & \cellcolor{forestgreen!25} +0.39 & \cellcolor{forestgreen!25} +1.08 & \cellcolor{forestgreen!25} +0.32 \\
       Formality   & Formality score ($10^{-2}$)   $\uparrow$    & +5.54 & +5.57 & +5.47 & +5.63 & \cellcolor{salmon!25} -0.41 & \cellcolor{salmon!25} -0.5 & \cellcolor{salmon!25} -0.15 & \cellcolor{salmon!25} -0.19 & \cellcolor{salmon!25} -0.08 & \cellcolor{salmon!25} -0.12 \\
       Readability & Flesch reading ease       $\uparrow$    & +28.23 & +32.87 & +28.02 & +29.39 & \cellcolor{salmon!25} -18.33 & \cellcolor{salmon!25} -18.38 & \cellcolor{salmon!25} -4.09 & \cellcolor{salmon!25} -4.2 & \cellcolor{salmon!25} -3.92 & \cellcolor{salmon!25} -3.58 \\
	\midrule
        \\ [-10pt]
        \multicolumn{2}{c}{Rhetorical Norms}
        \\\cmidrule(lr){1-2}
       Quant. Evidence & \% Sent. with QE     $\uparrow$    & +0.01 & +0.01 & +0.01 & +0.01 & \cellcolor{forestgreen!25} +0.01 & \cellcolor{forestgreen!25} +0.02 & \cellcolor{forestgreen!25} +0.03 & \cellcolor{forestgreen!25} +0.03 & \cellcolor{forestgreen!25} +0.03 & \cellcolor{forestgreen!25} +0.03 \\
       Narrative Organization & Background Skew $\uparrow$    & +0.51 & +0.58 & +0.47 & +0.53 & \cellcolor{forestgreen!25} +0.23 & \cellcolor{forestgreen!25} +0.19 & \cellcolor{forestgreen!25} +0.04 & \cellcolor{forestgreen!25} +0.01 & \cellcolor{forestgreen!25} +0.1 & \cellcolor{forestgreen!25} +0.07 \\
                              & Objective Skew   $\uparrow$    & -0.05 & +0.16 & -0.12 & -0.05 & \cellcolor{salmon!25} -0.47 & \cellcolor{salmon!25} -0.46 & \cellcolor{salmon!25} -0.09 & \cellcolor{salmon!25} -0.05 & \cellcolor{salmon!25} -0.07 & \cellcolor{salmon!25} -0.04 \\
                              & Method Skew   $\uparrow$    & -0.4 & -0.37 & -0.41 & -0.38 & \cellcolor{forestgreen!25} +0.11 & \cellcolor{forestgreen!25} +0.03 & \cellcolor{forestgreen!25} +0.02 & \cellcolor{salmon!25} -0.03 & \cellcolor{forestgreen!25} +0.09 & \cellcolor{forestgreen!25} +0.02 \\
                              & Result Skew   $\downarrow$  & -0.23 & -0.35 & -0.37 & -0.46 & \cellcolor{forestgreen!25} -0.26 & \cellcolor{forestgreen!25} -0.38 & \cellcolor{forestgreen!25} -0.06 & \cellcolor{salmon!25} +0.03 & \cellcolor{forestgreen!25} -0.1 & \cellcolor{forestgreen!25} -0.04 \\
                
        \bottomrule \\
    \end{tabular}
    }
    \caption{\small
        Results for the \textbf{natural language processing} community. The in-community column shows the metric value of papers from the community, out-community column shows the weighted average of data from all other communities. The random and specificity baselines show metric values before adaptation. The last six model columns show the \textbf{change} in value after adaptation from the random and specificity baselines, respectively. $\uparrow$ indicates that the metric should increase because the in-community value is > out-community value while $\downarrow$ indicates the vice versa. The cells where the $\Delta$ follows the expected trend are coloured {\color{forestgreen}green} while those that don't are coloured {\color{salmon}red}
    }
    \label{table:nlp}
\end{table*}

\begin{table*}[th!]
    \centering
    \resizebox{1\linewidth}{!}{
    \footnotesize
    \begin{tabular}{llccccccccccc}
    \toprule
    \multicolumn{2}{c}{Target = "Speech"}& \multicolumn{4}{c}{Baselines} & \multicolumn{2}{c}{Adapted by GPT} & \multicolumn{2}{c}{Adapted by Llama} & \multicolumn{2}{c}{Adapted by Mistral} 
    \\ \cmidrule(lr){1-2}\cmidrule(lr){3-5}\cmidrule(lr){6-7}\cmidrule(lr){8-9}\cmidrule(lr){10-11}
    feature & metric & out-comm. & in-comm. & random & specific & random & specific & random & specific & random  & specific 
	\\
    \midrule 
        \\ [-10pt]
        \multicolumn{2}{c}{Structural Norms}
        \\\cmidrule(lr){1-2}
        Length & Avg. \# words     $\downarrow$  & +628.73 & +528.17 & +671.9 & +589.02 & \cellcolor{forestgreen!25} -344.25 & \cellcolor{forestgreen!25} -294.3 & \cellcolor{forestgreen!25} -75.48 & \cellcolor{forestgreen!25} -61.86 & \cellcolor{forestgreen!25} -56.44 & \cellcolor{forestgreen!25} -44.85 \\
               & Avg. \# sentences $\downarrow$  & +29.94 & +25.18 & +31.43 & +27.48 & \cellcolor{forestgreen!25} -17.12 & \cellcolor{forestgreen!25} -14.51 & \cellcolor{forestgreen!25} -4.63 & \cellcolor{forestgreen!25} -3.41 & \cellcolor{forestgreen!25} -2.92 & \cellcolor{forestgreen!25} -1.88 \\
        Structural Artefacts & \% papers w/ tables $\downarrow$  & +7.64 & +2.87 & +7.1 & +4.11 & \cellcolor{forestgreen!25} -5.74 & \cellcolor{forestgreen!25} -3.35 & \cellcolor{forestgreen!25} -1.88 & \cellcolor{forestgreen!25} -0.81 & \cellcolor{forestgreen!25} -1.31 & \cellcolor{forestgreen!25} -0.91 \\
                            & \% papers w/ figures $\downarrow$  & +32.93 & +10.35 & +30.5 & +25.58 & \cellcolor{forestgreen!25} -24.6 & \cellcolor{forestgreen!25} -21.12 & \cellcolor{forestgreen!25} -3.12 & \cellcolor{forestgreen!25} -1.76 & \cellcolor{forestgreen!25} -5.98 & \cellcolor{forestgreen!25} -4.82 \\
	\midrule
        \\ [-10pt]
       \multicolumn{2}{c}{Stylistics Norms}
        \\\cmidrule(lr){1-2}
       Jargon      & Specificity score ($10^{-2}$) $\uparrow$    & -2.19 & +2.72 & -2.22 & +0.46 & \cellcolor{forestgreen!25} +2.38 & \cellcolor{forestgreen!25} +1.5 & \cellcolor{forestgreen!25} +2.2 & \cellcolor{forestgreen!25} +1.25 & \cellcolor{forestgreen!25} +1.75 & \cellcolor{forestgreen!25} +0.95 \\
       Formality   & Formality score ($10^{-2}$)   $\downarrow$  & +5.55 & +5.54 & +5.49 & +5.71 & \cellcolor{forestgreen!25} -0.49 & \cellcolor{forestgreen!25} -0.65 & \cellcolor{forestgreen!25} -0.15 & \cellcolor{forestgreen!25} -0.31 & \cellcolor{forestgreen!25} -0.11 & \cellcolor{forestgreen!25} -0.23 \\
       Readability & Flesch reading ease       $\downarrow$  & +29.44 & +27.26 & +27.02 & +27.08 & \cellcolor{forestgreen!25} -15.91 & \cellcolor{forestgreen!25} -16.51 & \cellcolor{forestgreen!25} -0.83 & \cellcolor{forestgreen!25} -2.79 & \cellcolor{forestgreen!25} -0.58 & \cellcolor{forestgreen!25} -3.14 \\
	\midrule
        \\ [-10pt]
        \multicolumn{2}{c}{Rhetorical Norms}
        \\\cmidrule(lr){1-2}
       Quant. Evidence & \% Sent. with QE     $\downarrow$  & +0.01 & +0.01 & +0.01 & +0.01 & \cellcolor{salmon!25} +0.01 & \cellcolor{salmon!25} +0.02 & \cellcolor{salmon!25} +0.03 & \cellcolor{salmon!25} +0.04 & \cellcolor{salmon!25} +0.03 & \cellcolor{salmon!25} +0.04 \\
       Narrative Organization & Background Skew $\downarrow$  & +0.52 & +0.42 & +0.47 & +0.51 & \cellcolor{salmon!25} +0.24 & \cellcolor{salmon!25} +0.2 & \cellcolor{salmon!25} +0.01 & \cellcolor{forestgreen!25} -0.0 & \cellcolor{salmon!25} +0.03 & \cellcolor{salmon!25} +0.04 \\
                              & Objective Skew   $\downarrow$  & +0.02 & -0.65 & -0.03 & -0.17 & \cellcolor{forestgreen!25} -0.51 & \cellcolor{forestgreen!25} -0.53 & \cellcolor{forestgreen!25} -0.14 & \cellcolor{forestgreen!25} -0.12 & \cellcolor{forestgreen!25} -0.11 & \cellcolor{forestgreen!25} -0.06 \\
                              & Method Skew   $\uparrow$    & -0.4 & -0.31 & -0.4 & -0.4 & \cellcolor{forestgreen!25} +0.1 & \cellcolor{forestgreen!25} +0.13 & \cellcolor{salmon!25} -0.0 & \cellcolor{salmon!25} -0.01 & \cellcolor{forestgreen!25} +0.08 & \cellcolor{forestgreen!25} +0.05 \\
                              & Result Skew   $\uparrow$    & -0.25 & -0.23 & -0.29 & -0.38 & \cellcolor{salmon!25} -0.32 & \cellcolor{salmon!25} -0.24 & \cellcolor{forestgreen!25} +0.05 & \cellcolor{forestgreen!25} +0.11 & \cellcolor{salmon!25} -0.02 & \cellcolor{forestgreen!25} +0.01 \\
    \bottomrule \\
    \end{tabular}
    }
    \caption{\small
        Results for the \textbf{speech} community. The in-community column shows the metric value of papers from the community, out-community column shows the weighted average of data from all other communities. The random and specificity baselines show metric values before adaptation. The last six model columns show the \textbf{change} in value after adaptation from the random and specificity baselines, respectively. $\uparrow$ indicates that the metric should increase because the in-community value is > out-community value while $\downarrow$ indicates the vice versa. The cells where the $\Delta$ follows the expected trend are coloured {\color{forestgreen}green} while those that don't are coloured {\color{salmon}red}
    }
    \label{table:speech}
\end{table*}

\begin{table*}[th!]
    \centering
    \resizebox{1\linewidth}{!}{
    \footnotesize
    \begin{tabular}{llccccccccccc}
    \toprule
    \multicolumn{2}{c}{Target = "Web and RecSys"}& \multicolumn{4}{c}{Baselines} & \multicolumn{2}{c}{Adapted by GPT} & \multicolumn{2}{c}{Adapted by Llama} & \multicolumn{2}{c}{Adapted by Mistral} 
    \\ \cmidrule(lr){1-2}\cmidrule(lr){3-5}\cmidrule(lr){6-7}\cmidrule(lr){8-9}\cmidrule(lr){10-11}
    feature & metric & out-comm. & in-comm. & random & specific & random & specific & random & specific & random  & specific 
		\\
    \midrule 
        \\ [-10pt]
        \multicolumn{2}{c}{Structural Norms}
        \\\cmidrule(lr){1-2}
        Length & Avg. \# words     $\downarrow$  & +621.97 & +615.61 & +644.49 & +545.42 & \cellcolor{forestgreen!25} -332.39 & \cellcolor{forestgreen!25} -269.61 & \cellcolor{forestgreen!25} -105.31 & \cellcolor{forestgreen!25} -126.31 & \cellcolor{forestgreen!25} -89.31 & \cellcolor{forestgreen!25} -89.59 \\
               & Avg. \# sentences $\downarrow$  & +29.66 & +28.74 & +30.38 & +25.55 & \cellcolor{forestgreen!25} -16.93 & \cellcolor{forestgreen!25} -13.79 & \cellcolor{forestgreen!25} -5.87 & \cellcolor{forestgreen!25} -6.73 & \cellcolor{forestgreen!25} -3.01 & \cellcolor{salmon!25} +2.45 \\
        Structural Artefacts & \% papers w/ tables $\downarrow$  & +7.39 & +5.79 & +5.42 & +4.01 & \cellcolor{forestgreen!25} -4.18 & \cellcolor{forestgreen!25} -3.47 & \cellcolor{forestgreen!25} -1.86 & \cellcolor{forestgreen!25} -1.43 & \cellcolor{forestgreen!25} -2.02 & \cellcolor{forestgreen!25} -1.21 \\
                     & \% papers w/ figures $\downarrow$  & +31.69 & +25.36 & +30.09 & +23.65 & \cellcolor{forestgreen!25} -25.73 & \cellcolor{forestgreen!25} -19.51 & \cellcolor{forestgreen!25} -6.35 & \cellcolor{forestgreen!25} -7.07 & \cellcolor{forestgreen!25} -10.27 & \cellcolor{forestgreen!25} -8.23 \\

	\midrule
        \\ [-10pt]
       \multicolumn{2}{c}{Stylistics Norms}
        \\\cmidrule(lr){1-2}
       Jargon      & Specificity score ($10^{-2}$) $\uparrow$    & -1.4 & +1.24 & -0.88 & +1.59 & \cellcolor{forestgreen!25} +1.31 & \cellcolor{forestgreen!25} +0.85 & \cellcolor{forestgreen!25} +1.3 & \cellcolor{forestgreen!25} +0.66 & \cellcolor{forestgreen!25} +1.04 & \cellcolor{forestgreen!25} +0.47 \\
       Formality   & Formality score ($10^{-2}$)   $\downarrow$  & +5.55 & +5.43 & +5.5 & +5.72 & \cellcolor{forestgreen!25} -0.5 & \cellcolor{forestgreen!25} -0.69 & \cellcolor{forestgreen!25} -0.2 & \cellcolor{forestgreen!25} -0.47 & \cellcolor{forestgreen!25} -0.13 & \cellcolor{forestgreen!25} -0.29 \\
       Readability & Flesch reading ease       $\downarrow$  & +29.41 & +27.19 & +28.28 & +27.19 & \cellcolor{forestgreen!25} -20.73 & \cellcolor{forestgreen!25} -18.02 & \cellcolor{forestgreen!25} -6.76 & \cellcolor{forestgreen!25} -8.15 & \cellcolor{forestgreen!25} -6.39 & \cellcolor{forestgreen!25} -6.21 \\
	\midrule
        \\ [-10pt]
        \multicolumn{2}{c}{Rhetorical Norms}
        \\\cmidrule(lr){1-2}
       Quant. Evidence & \% Sent. with QE     $\uparrow$    & +0.01 & +0.01 & +0.01 & +0.01 & \cellcolor{forestgreen!25} +0.01 & \cellcolor{forestgreen!25} +0.02 & \cellcolor{forestgreen!25} +0.03 & \cellcolor{forestgreen!25} +0.05 & \cellcolor{forestgreen!25} +0.03 & \cellcolor{forestgreen!25} +0.04 \\
       Narrative Organization & Background Skew $\downarrow$  & +0.52 & +0.49 & +0.47 & +0.39 & \cellcolor{salmon!25} +0.26 & \cellcolor{salmon!25} +0.23 & \cellcolor{salmon!25} +0.02 & \cellcolor{salmon!25} +0.07 & \cellcolor{salmon!25} +0.08 & \cellcolor{salmon!25} +0.08 \\
                              & Objective Skew   $\downarrow$  & +0.01 & -0.24 & -0.08 & -0.25 & \cellcolor{forestgreen!25} -0.52 & \cellcolor{forestgreen!25} -0.41 & \cellcolor{forestgreen!25} -0.15 & \cellcolor{forestgreen!25} -0.16 & \cellcolor{forestgreen!25} -0.09 & \cellcolor{salmon!25} +0.12 \\
                              & Method Skew   $\downarrow$  & -0.39 & -0.47 & -0.41 & -0.47 & \cellcolor{salmon!25} +0.18 & \cellcolor{salmon!25} +0.04 & \cellcolor{salmon!25} +0.05 & \cellcolor{salmon!25} +0.02 & \cellcolor{salmon!25} +0.09 & \cellcolor{salmon!25} +0.05 \\
                              & Result Skew   $\downarrow$  & -0.24 & -0.39 & -0.31 & -0.3 & \cellcolor{forestgreen!25} -0.44 & \cellcolor{forestgreen!25} -0.4 & \cellcolor{salmon!25} +0.05 & \cellcolor{forestgreen!25} -0.09 & \cellcolor{forestgreen!25} -0.06 & \cellcolor{forestgreen!25} -0.13 \\

        \bottomrule \\
    \end{tabular}
    }
    \caption{\small
        Results for the \textbf{web and recommendation systems} community. The in-community column shows the metric value of papers from the community, out-community column shows the weighted average of data from all other communities. The random and specificity baselines show metric values before adaptation. The last six model columns show the \textbf{change} in value after adaptation from the random and specificity baselines, respectively. $\uparrow$ indicates that the metric should increase because the in-community value is > out-community value while $\downarrow$ indicates the vice versa. The cells where the $\Delta$ follows the expected trend are coloured {\color{forestgreen}green} while those that don't are coloured {\color{salmon}red}
    }
    \label{table:wrs}
\end{table*}

%% file: v1/tables/new_models_results.tex
\begin{table*}[th!]
    \centering
    \resizebox{1\linewidth}{!}{
    \footnotesize
    \begin{tabular}{llcccccccc}
    \toprule
    \multicolumn{2}{c}{Target = "AI"}& \multicolumn{4}{c}{Baselines} & \multicolumn{2}{c}{Adapted by GPT 4o Mini} & \multicolumn{2}{c}{Adapted by Llama 3.1 70B} \\
    \cmidrule(lr){1-2}\cmidrule(lr){3-6}\cmidrule(lr){7-8}\cmidrule(lr){9-10}
    feature & metric & out-comm. & in-comm. & random & specific & random & specific & random & specific \\
     \midrule 
        \\ [-10pt]
        \multicolumn{2}{c}{Structural Norms}
        \\\cmidrule(lr){1-2}
        Length & Avg. \# words     $\uparrow$    & +612.27 & +663.26 & +656.07 & +527.93 & \cellcolor{salmon!25} -90.16 & \cellcolor{salmon!25} -49.99 & \cellcolor{forestgreen!25} +1119.53 & \cellcolor{forestgreen!25} +1036.88 \\
               & Avg. \# sentences $\uparrow$    & +28.41 & +34.93 & +30.3 & +25.48 & \cellcolor{salmon!25} -5.7 & \cellcolor{salmon!25} -4.32 & \cellcolor{forestgreen!25} +48.19 & \cellcolor{forestgreen!25} +45.9 \\
        Structural Artefacts & \% papers w/ tables $\downarrow$  & +7.38 & +6.96 & +6.31 & +3.9 & \cellcolor{forestgreen!25} -1.21 & \cellcolor{forestgreen!25} -0.78 & \cellcolor{forestgreen!25} -6.31 & \cellcolor{forestgreen!25} -3.9 \\
                     & \% papers w/ figures $\uparrow$    & +31.03 & +32.67 & +27.93 & +24.92 & \cellcolor{salmon!25} -7.05 & \cellcolor{salmon!25} -6.44 & \cellcolor{salmon!25} -27.93 & \cellcolor{salmon!25} -24.92 \\

	\midrule
        \\ [-10pt]
       \multicolumn{2}{c}{Stylistic Norms}
        \\\cmidrule(lr){1-2}
       Jargon      & Specificity score ($10^{-2}$) $\uparrow$    & -0.7 & +0.46 & -0.79 & +0.44 & \cellcolor{forestgreen!25} +0.05 & \cellcolor{salmon!25} -0.14 & \cellcolor{forestgreen!25} +0.08 & \cellcolor{salmon!25} -1.21 \\
       Formality   & Formality score ($10^{-2}$)   $\uparrow$    & +5.52 & +5.65 & +5.44 & +5.67 & \cellcolor{salmon!25} -0.33 & \cellcolor{salmon!25} -0.4 & \cellcolor{salmon!25} -0.13 & \cellcolor{salmon!25} -0.21 \\
       Readability & Flesch reading ease       $\uparrow$    & +28.83 & +31.32 & +28.42 & +27.04 & \cellcolor{salmon!25} -17.77 & \cellcolor{salmon!25} -16.89 & \cellcolor{salmon!25} -28.42 & \cellcolor{salmon!25} -27.04 \\
	\midrule
        \\ [-10pt]
        \multicolumn{2}{c}{Rhetorical Norms}
        \\\cmidrule(lr){1-2}
       Quant. Evidence & \% Sent. with QE     $\downarrow$  & +0.01 & 0.0 & +0.01 & 0.0 & \cellcolor{salmon!25} +0.04 & \cellcolor{salmon!25} +0.02 & \cellcolor{forestgreen!25} -0.01 & \cellcolor{forestgreen!25} -0.0 \\
    \bottomrule
    \end{tabular}
    }
        \caption{\small
        Results for the \textbf{artificial intelligence} community. The in-community column shows the metric value of papers from the community, out-community column shows the weighted average of data from all other communities. The random and specificity baselines show metric values before adaptation. The last six model columns show the \textbf{change} in value after adaptation from the random and specificity baselines, respectively. $\uparrow$ indicates that the metric should increase because the in-community value is > out-community value while $\downarrow$ indicates the vice versa. The cells where the $\Delta$ follows the expected trend are coloured {\color{forestgreen}green} while those that don't are coloured {\color{salmon}red}
    }
    \label{table:new-ai}
\end{table*}

\begin{table*}[th!]
    \centering
    \resizebox{1\linewidth}{!}{
    \footnotesize
    \begin{tabular}{llcccccccc}
    \toprule
    \multicolumn{2}{c}{Target = "CV"}& \multicolumn{4}{c}{Baselines} & \multicolumn{2}{c}{Adapted by GPT 4o Mini} & \multicolumn{2}{c}{Adapted by Llama 3.1 70B} \\
    \cmidrule(lr){1-2}\cmidrule(lr){3-6}\cmidrule(lr){7-8}\cmidrule(lr){9-10}
    feature & metric & out-comm. & in-comm. & random & specific & random & specific & random & specific \\
        \midrule 
        \\ [-10pt]
        \multicolumn{2}{c}{Structural Norms}
        \\\cmidrule(lr){1-2}
        Length & Avg. \# words     $\uparrow$    & +619.97 & +655.03 & +641.11 & +596.78 & \cellcolor{salmon!25} -97.49 & \cellcolor{salmon!25} -76.26 & \cellcolor{forestgreen!25} +1189.17 & \cellcolor{forestgreen!25} +1210.2 \\
               & Avg. \# sentences $\uparrow$    & +29.47 & +32.41 & +29.7 & +28.08 & \cellcolor{salmon!25} -5.41 & \cellcolor{salmon!25} -4.84 & \cellcolor{forestgreen!25} +52.18 & \cellcolor{forestgreen!25} +53.72 \\
        Structural Artefacts & \% papers w/ tables $\downarrow$  & +7.37 & +5.88 & +8.02 & +3.7 & \cellcolor{forestgreen!25} -2.9 & \cellcolor{forestgreen!25} -1.2 & \cellcolor{forestgreen!25} -8.02 & \cellcolor{forestgreen!25} -3.7 \\
                     & \% papers w/ figures $\uparrow$    & +29.47 & +69.34 & +26.98 & +47.55 & \cellcolor{salmon!25} -9.76 & \cellcolor{salmon!25} -11.59 & \cellcolor{salmon!25} -26.98 & \cellcolor{salmon!25} -47.55 \\

	\midrule
        \\ [-10pt]
       \multicolumn{2}{c}{Stylistic Norms}
        \\\cmidrule(lr){1-2}
       Jargon      & Specificity score ($10^{-2}$) $\uparrow$    & -2.14 & +1.95 & -2.3 & +0.79 & \cellcolor{forestgreen!25} +1.43 & \cellcolor{forestgreen!25} +0.34 & \cellcolor{forestgreen!25} +0.45 & \cellcolor{salmon!25} -2.85 \\
       Formality   & Formality score ($10^{-2}$)   $\downarrow$  & +5.55 & +5.48 & +5.49 & +5.53 & \cellcolor{forestgreen!25} -0.33 & \cellcolor{forestgreen!25} -0.37 & \cellcolor{forestgreen!25} -0.09 & \cellcolor{forestgreen!25} -0.18 \\
       Readability & Flesch reading ease       $\downarrow$  & +29.44 & +26.1 & +28.52 & +25.14 & \cellcolor{forestgreen!25} -16.33 & \cellcolor{forestgreen!25} -17.14 & \cellcolor{forestgreen!25} -28.52 & \cellcolor{forestgreen!25} -25.14 \\
	\midrule
        \\ [-10pt]
        \multicolumn{2}{c}{Rhetorical Norms}
        \\\cmidrule(lr){1-2}
       Quant. Evidence & \% Sent. with QE     $\downarrow$  & +0.01 & +0.01 & +0.01 & +0.01 & \cellcolor{salmon!25} +0.03 & \cellcolor{salmon!25} +0.03 & \cellcolor{forestgreen!25} -0.01 & \cellcolor{forestgreen!25} -0.01 \\
    \bottomrule
    \end{tabular}
    }
        \caption{\small
        Results for the \textbf{computer vision} community. The in-community column shows the metric value of papers from the community, out-community column shows the weighted average of data from all other communities. The random and specificity baselines show metric values before adaptation. The last six model columns show the \textbf{change} in value after adaptation from the random and specificity baselines, respectively. $\uparrow$ indicates that the metric should increase because the in-community value is > out-community value while $\downarrow$ indicates the vice versa. The cells where the $\Delta$ follows the expected trend are coloured {\color{forestgreen}green} while those that don't are coloured {\color{salmon}red}
    }
    \label{table:new-cv}
\end{table*}

\begin{table*}[th!]
    \centering
    \resizebox{1\linewidth}{!}{
    \footnotesize
    \begin{tabular}{llcccccccc}
    \toprule
    \multicolumn{2}{c}{Target = "Data Mining"}& \multicolumn{4}{c}{Baselines} & \multicolumn{2}{c}{Adapted by GPT 4o Mini} & \multicolumn{2}{c}{Adapted by Llama 3.1 70B} \\
    \cmidrule(lr){1-2}\cmidrule(lr){3-6}\cmidrule(lr){7-8}\cmidrule(lr){9-10}
    feature & metric & out-comm. & in-comm. & random & specific & random & specific & random & specific \\
        \midrule 
        \\ [-10pt]
        \multicolumn{2}{c}{Structural Norms}
        \\\cmidrule(lr){1-2}
        Length & Avg. \# words     $\uparrow$    & +618.55 & +690.88 & +644.89 & +517.61 & \cellcolor{salmon!25} -97.73 & \cellcolor{salmon!25} -79.32 & \cellcolor{forestgreen!25} +1236.92 & \cellcolor{forestgreen!25} +1191.66 \\
               & Avg. \# sentences $\uparrow$    & +29.46 & +33.02 & +30.29 & +25.0 & \cellcolor{salmon!25} -6.11 & \cellcolor{salmon!25} -5.4 & \cellcolor{forestgreen!25} +54.09 & \cellcolor{forestgreen!25} +52.53 \\
	Structural Artefacts & \% papers w/ tables $\uparrow$    & +7.3 & +7.34 & +6.5 & +3.3 & \cellcolor{salmon!25} -1.52 & \cellcolor{salmon!25} -0.76 & \cellcolor{salmon!25} -6.5 & \cellcolor{salmon!25} -3.3 \\
                     & \% papers w/ figures $\uparrow$    & +30.99 & +38.95 & +26.7 & +28.13 & \cellcolor{salmon!25} -8.42 & \cellcolor{salmon!25} -8.45 & \cellcolor{salmon!25} -26.7 & \cellcolor{salmon!25} -28.13 \\

    \midrule
        \\ [-10pt]
       \multicolumn{2}{c}{Stylistic Norms}
        \\\cmidrule(lr){1-2}
       Jargon      & Specificity score ($10^{-2}$) $\uparrow$    & -1.25 & +0.82 & -0.9 & +0.96 & \cellcolor{forestgreen!25} +0.31 & \cellcolor{salmon!25} -0.07 & \cellcolor{forestgreen!25} +0.26 & \cellcolor{salmon!25} -1.41 \\
       Formality   & Formality score ($10^{-2}$)   $\downarrow$  & +5.55 & +5.47 & +5.54 & +5.67 & \cellcolor{forestgreen!25} -0.38 & \cellcolor{forestgreen!25} -0.44 & \cellcolor{forestgreen!25} -0.18 & \cellcolor{forestgreen!25} -0.32 \\
       Readability & Flesch reading ease       $\downarrow$  & +29.4 & +26.72 & +28.43 & +27.73 & \cellcolor{forestgreen!25} -17.13 & \cellcolor{forestgreen!25} -17.39 & \cellcolor{forestgreen!25} -28.43 & \cellcolor{forestgreen!25} -27.73 \\
	\midrule
        \\ [-10pt]
        \multicolumn{2}{c}{Rhetorical Norms}
        \\\cmidrule(lr){1-2}
       Quant. Evidence & \% Sent. with QE     $\uparrow$    & +0.01 & +0.01 & +0.01 & +0.01 & \cellcolor{forestgreen!25} +0.04 & \cellcolor{forestgreen!25} +0.04 & \cellcolor{salmon!25} -0.01 & \cellcolor{salmon!25} -0.01 \\
    
    \bottomrule
    \end{tabular}
    }
    \caption{\small
        Results for the \textbf{data mining} community. The in-community column shows the metric value of papers from the community, out-community column shows the weighted average of data from all other communities. The random and specificity baselines show metric values before adaptation. The last six model columns show the \textbf{change} in value after adaptation from the random and specificity baselines, respectively. $\uparrow$ indicates that the metric should increase because the in-community value is > out-community value while $\downarrow$ indicates the vice versa. The cells where the $\Delta$ follows the expected trend are coloured {\color{forestgreen}green} while those that don't are coloured {\color{salmon}red}
    }
    \label{table:new-dm}
    
\end{table*}

\begin{table*}[th!]
    \centering
    \resizebox{1\linewidth}{!}{
    \footnotesize
    \begin{tabular}{llcccccccc}
    \toprule
    \multicolumn{2}{c}{Target = "Economics \& Computation"}& \multicolumn{4}{c}{Baselines} & \multicolumn{2}{c}{Adapted by GPT 4o Mini} & \multicolumn{2}{c}{Adapted by Llama 3.1 70B} \\
    \cmidrule(lr){1-2}\cmidrule(lr){3-6}\cmidrule(lr){7-8}\cmidrule(lr){9-10}
    feature & metric & out-comm. & in-comm. & random & specific & random & specific & random & specific \\
    \midrule 
        \\ [-10pt]
        \multicolumn{2}{c}{Structural Norms}
        \\\cmidrule(lr){1-2}
        Length & Avg. \# words     $\uparrow$    & +616.79 & +1107.37 & +619.48 & +613.61 & \cellcolor{salmon!25} -85.7 & \cellcolor{salmon!25} -105.02 & \cellcolor{forestgreen!25} +1075.65 & \cellcolor{forestgreen!25} +993.13 \\
               & Avg. \# sentences $\uparrow$    & +29.4 & +50.67 & +29.64 & +28.46 & \cellcolor{salmon!25} -6.22 & \cellcolor{salmon!25} -6.45 & \cellcolor{forestgreen!25} +44.79 & \cellcolor{forestgreen!25} +40.38 \\
	Structural Artefacts & \% papers w/ tables $\downarrow$  & +7.33 & +4.77 & +6.5 & +5.01 & \cellcolor{forestgreen!25} -2.44 & \cellcolor{forestgreen!25} -1.63 & \cellcolor{forestgreen!25} -6.5 & \cellcolor{forestgreen!25} -5.01 \\
                     & \% papers w/ figures $\downarrow$  & +31.54 & +10.29 & +30.3 & +19.04 & \cellcolor{forestgreen!25} -12.64 & \cellcolor{forestgreen!25} -6.94 & \cellcolor{forestgreen!25} -30.3 & \cellcolor{forestgreen!25} -19.04 \\

    \midrule
        \\ [-10pt]
       \multicolumn{2}{c}{Stylistic Norms}
        \\\cmidrule(lr){1-2}
       Jargon      & Specificity score ($10^{-2}$) $\uparrow$    & -3.59 & +2.94 & -3.28 & +0.76 & \cellcolor{forestgreen!25} +0.21 & \cellcolor{salmon!25} -0.3 & \cellcolor{forestgreen!25} +0.15 & \cellcolor{salmon!25} -3.62 \\
       Formality   & Formality score ($10^{-2}$)   $\downarrow$  & +5.55 & +5.51 & +5.51 & +5.65 & \cellcolor{forestgreen!25} -0.43 & \cellcolor{forestgreen!25} -0.51 & \cellcolor{forestgreen!25} -0.23 & \cellcolor{forestgreen!25} -0.37 \\
       Readability & Flesch reading ease       $\uparrow$    & +29.25 & +32.95 & +28.07 & +31.19 & \cellcolor{salmon!25} -22.32 & \cellcolor{salmon!25} -25.06 & \cellcolor{salmon!25} -28.07 & \cellcolor{salmon!25} -31.19 \\
	\midrule
        \\ [-10pt]
        \multicolumn{2}{c}{Rhetorical Norms}
        \\\cmidrule(lr){1-2}
       Quant. Evidence & \% Sent. with QE     $\uparrow$    & +0.01 & +0.01 & +0.01 & +0.01 & \cellcolor{forestgreen!25} +0.03 & \cellcolor{forestgreen!25} +0.05 & \cellcolor{salmon!25} -0.01 & \cellcolor{salmon!25} -0.01 \\
    
    \bottomrule
    \end{tabular}
    }
    \caption{\small
        Results for the \textbf{economics and computation} community. The in-community column shows the metric value of papers from the community, out-community column shows the weighted average of data from all other communities. The random and specificity baselines show metric values before adaptation. The last six model columns show the \textbf{change} in value after adaptation from the random and specificity baselines, respectively. $\uparrow$ indicates that the metric should increase because the in-community value is > out-community value while $\downarrow$ indicates the vice versa. The cells where the $\Delta$ follows the expected trend are coloured {\color{forestgreen}green} while those that don't are coloured {\color{salmon}red}
    }
    \label{table:new-ec}    

\end{table*}

\begin{table*}[th!]
    \centering
    \resizebox{1\linewidth}{!}{
    \footnotesize
    \begin{tabular}{llcccccccc}
    \toprule
    \multicolumn{2}{c}{Target = "Education"}& \multicolumn{4}{c}{Baselines} & \multicolumn{2}{c}{Adapted by GPT 4o Mini} & \multicolumn{2}{c}{Adapted by Llama 3.1 70B} \\
    \cmidrule(lr){1-2}\cmidrule(lr){3-6}\cmidrule(lr){7-8}\cmidrule(lr){9-10}
    feature & metric & out-comm. & in-comm. & random & specific & random & specific & random & specific \\
    \midrule 
        \\ [-10pt]
        \multicolumn{2}{c}{Structural Norms}
        \\\cmidrule(lr){1-2}
        Length & Avg. \# words     $\downarrow$  & +624.59 & +422.89 & +687.76 & +488.39 & \cellcolor{forestgreen!25} -125.39 & \cellcolor{forestgreen!25} -86.67 & \cellcolor{salmon!25} +1178.45 & \cellcolor{salmon!25} +1048.82 \\
               & Avg. \# sentences $\downarrow$  & +29.77 & +18.8 & +33.04 & +22.77 & \cellcolor{forestgreen!25} -7.71 & \cellcolor{forestgreen!25} -4.55 & \cellcolor{salmon!25} +50.71 & \cellcolor{salmon!25} +44.74 \\
	Structural Artefacts & \% papers w/ tables $\downarrow$  & +7.37 & +2.84 & +6.51 & +4.31 & \cellcolor{forestgreen!25} -3.11 & \cellcolor{forestgreen!25} -2.09 & \cellcolor{forestgreen!25} -6.51 & \cellcolor{forestgreen!25} -4.31 \\
                     & \% papers w/ figures $\downarrow$  & +31.7 & +6.34 & +30.56 & +17.64 & \cellcolor{forestgreen!25} -19.06 & \cellcolor{forestgreen!25} -9.22 & \cellcolor{forestgreen!25} -30.56 & \cellcolor{forestgreen!25} -17.64 \\
    \midrule
        \\ [-10pt]
       \multicolumn{2}{c}{Stylistic Norms}
        \\\cmidrule(lr){1-2}
       Jargon      & Specificity score ($10^{-2}$) $\uparrow$    & -4.46 & +4.46 & -3.97 & +0.24 & \cellcolor{forestgreen!25} +2.68 & \cellcolor{forestgreen!25} +1.53 & \cellcolor{forestgreen!25} +0.68 & \cellcolor{salmon!25} -3.13 \\
       Formality   & Formality score ($10^{-2}$)   $\downarrow$  & +5.55 & +5.39 & +5.52 & +5.87 & \cellcolor{forestgreen!25} -0.49 & \cellcolor{forestgreen!25} -0.55 & \cellcolor{forestgreen!25} -0.26 & \cellcolor{forestgreen!25} -0.62 \\
       Readability & Flesch reading ease       $\downarrow$  & +29.35 & +25.31 & +28.64 & +27.3 & \cellcolor{forestgreen!25} -20.86 & \cellcolor{forestgreen!25} -23.77 & \cellcolor{forestgreen!25} -28.64 & \cellcolor{forestgreen!25} -27.3 \\
	\midrule
        \\ [-10pt]
        \multicolumn{2}{c}{Rhetorical Norms}
        \\\cmidrule(lr){1-2}
       Quant. Evidence & \% Sent. with QE     $\uparrow$    & +0.01 & +0.01 & +0.01 & +0.01 & \cellcolor{forestgreen!25} +0.02 & \cellcolor{forestgreen!25} +0.03 & \cellcolor{salmon!25} -0.01 & \cellcolor{salmon!25} -0.01 \\
    
    \bottomrule
    \end{tabular}
    }
    \caption{\small
        Results for the \textbf{education} community. The in-community column shows the metric value of papers from the community, out-community column shows the weighted average of data from all other communities. The random and specificity baselines show metric values before adaptation. The last six model columns show the \textbf{change} in value after adaptation from the random and specificity baselines, respectively. $\uparrow$ indicates that the metric should increase because the in-community value is > out-community value while $\downarrow$ indicates the vice versa. The cells where the $\Delta$ follows the expected trend are coloured {\color{forestgreen}green} while those that don't are coloured {\color{salmon}red}
    }
    \label{table:new-education}
\end{table*}

\begin{table*}[th!]
    \centering
    \resizebox{1\linewidth}{!}{
    \footnotesize
    \begin{tabular}{llcccccccc}
    \toprule
    \multicolumn{2}{c}{Target = "HCI"}& \multicolumn{4}{c}{Baselines} & \multicolumn{2}{c}{Adapted by GPT 4o Mini} & \multicolumn{2}{c}{Adapted by Llama 3.1 70B} \\
    \cmidrule(lr){1-2}\cmidrule(lr){3-6}\cmidrule(lr){7-8}\cmidrule(lr){9-10}
    feature & metric & out-comm. & in-comm. & random & specific & random & specific & random & specific \\
    \midrule 
        \\ [-10pt]
        \multicolumn{2}{c}{Structural Norms}
        \\\cmidrule(lr){1-2}
        Length & Avg. \# words     $\downarrow$  & +622.85 & +605.04 & +656.23 & +474.86 & \cellcolor{forestgreen!25} -121.62 & \cellcolor{forestgreen!25} -89.98 & \cellcolor{salmon!25} +1079.0 & \cellcolor{salmon!25} +973.07 \\
               & Avg. \# sentences $\downarrow$  & +29.89 & +25.86 & +31.06 & +22.18 & \cellcolor{forestgreen!25} -7.79 & \cellcolor{forestgreen!25} -5.24 & \cellcolor{salmon!25} +44.8 & \cellcolor{salmon!25} +38.78 \\
	Structural Artefacts & \% papers w/ tables $\downarrow$  & +7.54 & +4.1 & +6.4 & +3.41 & \cellcolor{forestgreen!25} -1.44 & \cellcolor{forestgreen!25} -0.91 & \cellcolor{forestgreen!25} -6.4 & \cellcolor{forestgreen!25} -3.41 \\
                     & \% papers w/ figures $\uparrow$    & +31.25 & +32.37 & +29.5 & +18.04 & \cellcolor{salmon!25} -10.5 & \cellcolor{salmon!25} -5.18 & \cellcolor{salmon!25} -29.5 & \cellcolor{salmon!25} -18.04 \\
    \midrule
        \\ [-10pt]
       \multicolumn{2}{c}{Stylistic Norms}
        \\\cmidrule(lr){1-2}
       Jargon      & Specificity score ($10^{-2}$) $\uparrow$    & -3.49 & +3.18 & -2.54 & +1.74 & \cellcolor{forestgreen!25} +1.23 & \cellcolor{forestgreen!25} +0.62 & \cellcolor{forestgreen!25} +0.71 & \cellcolor{salmon!25} -3.1 \\
       Formality   & Formality score ($10^{-2}$)   $\downarrow$  & +5.57 & +5.26 & +5.5 & +5.67 & \cellcolor{forestgreen!25} -0.38 & \cellcolor{forestgreen!25} -0.48 & \cellcolor{forestgreen!25} -0.2 & \cellcolor{forestgreen!25} -0.48 \\
       Readability & Flesch reading ease       $\downarrow$  & +29.65 & +24.4 & +27.61 & +25.51 & \cellcolor{forestgreen!25} -17.58 & \cellcolor{forestgreen!25} -19.96 & \cellcolor{forestgreen!25} -27.61 & \cellcolor{forestgreen!25} -25.51 \\
	\midrule
        \\ [-10pt]
        \multicolumn{2}{c}{Rhetorical Norms}
        \\\cmidrule(lr){1-2}
       Quant. Evidence & \% Sent. with QE     $\uparrow$    & +0.01 & +0.01 & +0.01 & +0.01 & \cellcolor{forestgreen!25} +0.03 & \cellcolor{forestgreen!25} +0.04 & \cellcolor{salmon!25} -0.01 & \cellcolor{salmon!25} -0.01 \\
    
    \bottomrule
    \end{tabular}
    }
    \caption{\small
        Results for the \textbf{human computer interaction} community. The in-community column shows the metric value of papers from the community, out-community column shows the weighted average of data from all other communities. The random and specificity baselines show metric values before adaptation. The last six model columns show the \textbf{change} in value after adaptation from the random and specificity baselines, respectively. $\uparrow$ indicates that the metric should increase because the in-community value is > out-community value while $\downarrow$ indicates the vice versa. The cells where the $\Delta$ follows the expected trend are coloured {\color{forestgreen}green} while those that don't are coloured {\color{salmon}red}
    }
    \label{table:new-hci}    
\end{table*}

\begin{table*}[th!]
    \centering
    \resizebox{1\linewidth}{!}{
    \footnotesize
    \begin{tabular}{llcccccccc}
    \toprule
    \multicolumn{2}{c}{Target = "IR"}& \multicolumn{4}{c}{Baselines} & \multicolumn{2}{c}{Adapted by GPT 4o Mini} & \multicolumn{2}{c}{Adapted by Llama 3.1 70B} \\
    \cmidrule(lr){1-2}\cmidrule(lr){3-6}\cmidrule(lr){7-8}\cmidrule(lr){9-10}
    feature & metric & out-comm. & in-comm. & random & specific & random & specific & random & specific \\
    \midrule 
        \\ [-10pt]
        \multicolumn{2}{c}{Structural Norms}
        \\\cmidrule(lr){1-2}
        Length & Avg. \# words     $\downarrow$  & +622.39 & +610.41 & +654.45 & +555.38 & \cellcolor{forestgreen!25} -115.88 & \cellcolor{forestgreen!25} -89.55 & \cellcolor{salmon!25} +1171.59 & \cellcolor{salmon!25} +1151.84 \\
               & Avg. \# sentences $\downarrow$  & +29.64 & +29.06 & +30.34 & +27.0 & \cellcolor{forestgreen!25} -6.62 & \cellcolor{forestgreen!25} -5.75 & \cellcolor{salmon!25} +50.15 & \cellcolor{salmon!25} +49.7 \\
	Structural Artefacts & \% papers w/ tables $\uparrow$    & +7.26 & +7.85 & +5.41 & +5.61 & \cellcolor{salmon!25} -1.51 & \cellcolor{salmon!25} -1.35 & \cellcolor{salmon!25} -5.41 & \cellcolor{salmon!25} -5.61 \\
                     & \% papers w/ figures $\uparrow$    & +31.19 & +33.37 & +28.13 & +30.33 & \cellcolor{salmon!25} -10.15 & \cellcolor{salmon!25} -8.67 & \cellcolor{salmon!25} -28.13 & \cellcolor{salmon!25} -30.33 \\
    \midrule
        \\ [-10pt]
       \multicolumn{2}{c}{Stylistic Norms}
        \\\cmidrule(lr){1-2}
       Jargon      & Specificity score ($10^{-2}$) $\uparrow$    & -1.26 & +1.14 & -0.91 & +1.27 & \cellcolor{forestgreen!25} +1.02 & \cellcolor{forestgreen!25} +0.11 & \cellcolor{forestgreen!25} +0.37 & \cellcolor{salmon!25} -1.64 \\
       Formality   & Formality score ($10^{-2}$)   $\downarrow$  & +5.55 & +5.49 & +5.46 & +5.86 & \cellcolor{forestgreen!25} -0.38 & \cellcolor{forestgreen!25} -0.5 & \cellcolor{forestgreen!25} -0.17 & \cellcolor{forestgreen!25} -0.54 \\
       Readability & Flesch reading ease       $\downarrow$  & +29.49 & +26.39 & +27.54 & +28.75 & \cellcolor{forestgreen!25} -17.81 & \cellcolor{forestgreen!25} -20.13 & \cellcolor{forestgreen!25} -27.54 & \cellcolor{forestgreen!25} -28.75 \\
	\midrule
        \\ [-10pt]
        \multicolumn{2}{c}{Rhetorical Norms}
        \\\cmidrule(lr){1-2}
       Quant. Evidence & \% Sent. with QE     $\uparrow$    & +0.01 & +0.01 & +0.01 & +0.01 & \cellcolor{forestgreen!25} +0.03 & \cellcolor{forestgreen!25} +0.03 & \cellcolor{salmon!25} -0.01 & \cellcolor{salmon!25} -0.01 \\
    
    \bottomrule
    \end{tabular}
    }
    \caption{\small
        Results for the \textbf{information retrieval} community. The in-community column shows the metric value of papers from the community, out-community column shows the weighted average of data from all other communities. The random and specificity baselines show metric values before adaptation. The last six model columns show the \textbf{change} in value after adaptation from the random and specificity baselines, respectively. $\uparrow$ indicates that the metric should increase because the in-community value is > out-community value while $\downarrow$ indicates the vice versa. The cells where the $\Delta$ follows the expected trend are coloured {\color{forestgreen}green} while those that don't are coloured {\color{salmon}red} 
    }
\end{table*}

\begin{table*}[th!]
    \centering
    \resizebox{1\linewidth}{!}{
    \footnotesize
    \begin{tabular}{llcccccccc}
    \toprule
    \multicolumn{2}{c}{Target = "ML"}& \multicolumn{4}{c}{Baselines} & \multicolumn{2}{c}{Adapted by GPT 4o Mini} & \multicolumn{2}{c}{Adapted by Llama 3.1 70B} \\
    \cmidrule(lr){1-2}\cmidrule(lr){3-6}\cmidrule(lr){7-8}\cmidrule(lr){9-10}
    feature & metric & out-comm. & in-comm. & random & specific & random & specific & random & specific \\
    \midrule 
        \\ [-10pt]
        \multicolumn{2}{c}{Structural Norms}
        \\\cmidrule(lr){1-2}
        Length & Avg. \# words     $\uparrow$    & +601.8 & +695.38 & +651.35 & +547.91 & \cellcolor{salmon!25} -100.03 & \cellcolor{salmon!25} -53.63 & \cellcolor{forestgreen!25} +1038.7 & \cellcolor{forestgreen!25} +996.15 \\
               & Avg. \# sentences $\uparrow$    & +28.7 & +33.0 & +30.44 & +26.48 & \cellcolor{salmon!25} -6.3 & \cellcolor{salmon!25} -4.41 & \cellcolor{forestgreen!25} +45.19 & \cellcolor{forestgreen!25} +44.65 \\
	Structural Artefacts & \% papers w/ tables $\downarrow$  & +7.5 & +6.57 & +6.21 & +4.02 & \cellcolor{forestgreen!25} -1.35 & \cellcolor{forestgreen!25} -0.7 & \cellcolor{forestgreen!25} -6.21 & \cellcolor{forestgreen!25} -4.02 \\
                     & \% papers w/ figures $\downarrow$  & +31.8 & +29.58 & +28.73 & +16.08 & \cellcolor{forestgreen!25} -7.55 & \cellcolor{forestgreen!25} -4.3 & \cellcolor{forestgreen!25} -28.73 & \cellcolor{forestgreen!25} -16.08 \\
    \midrule
        \\ [-10pt]
       \multicolumn{2}{c}{Stylistic Norms}
        \\\cmidrule(lr){1-2}
       Jargon      & Specificity score ($10^{-2}$) $\uparrow$    & -1.66 & +1.08 & -1.69 & +0.8 & \cellcolor{forestgreen!25} +0.15 & \cellcolor{salmon!25} -0.08 & \cellcolor{forestgreen!25} +0.2 & \cellcolor{salmon!25} -2.57 \\
       Formality   & Formality score ($10^{-2}$)   $\uparrow$    & +5.53 & +5.62 & +5.5 & +5.6 & \cellcolor{salmon!25} -0.38 & \cellcolor{salmon!25} -0.38 & \cellcolor{salmon!25} -0.17 & \cellcolor{salmon!25} -0.16 \\
       Readability & Flesch reading ease       $\downarrow$  & +29.43 & +28.74 & +28.16 & +25.94 & \cellcolor{forestgreen!25} -16.48 & \cellcolor{forestgreen!25} -15.21 & \cellcolor{forestgreen!25} -28.16 & \cellcolor{forestgreen!25} -25.94 \\
	\midrule
        \\ [-10pt]
        \multicolumn{2}{c}{Rhetorical Norms}
        \\\cmidrule(lr){1-2}
       Quant. Evidence & \% Sent. with QE     $\downarrow$  & +0.01 & 0.0 & +0.01 & +0.01 & \cellcolor{salmon!25} +0.04 & \cellcolor{salmon!25} +0.03 & \cellcolor{forestgreen!25} -0.01 & \cellcolor{forestgreen!25} -0.01 \\
    
    \bottomrule
    \end{tabular}
    }
    \caption{\small
        Results for the \textbf{machine learning} community. The in-community column shows the metric value of papers from the community, out-community column shows the weighted average of data from all other communities. The random and specificity baselines show metric values before adaptation. The last six model columns show the \textbf{change} in value after adaptation from the random and specificity baselines, respectively. $\uparrow$ indicates that the metric should increase because the in-community value is > out-community value while $\downarrow$ indicates the vice versa. The cells where the $\Delta$ follows the expected trend are coloured {\color{forestgreen}green} while those that don't are coloured {\color{salmon}red} 
    }
    \label{table:new-ml}    
\end{table*}

\begin{table*}[th!]
    \centering
    \resizebox{1\linewidth}{!}{
    \footnotesize
    \begin{tabular}{llcccccccc}
    \toprule
    \multicolumn{2}{c}{Target = "NLP"}& \multicolumn{4}{c}{Baselines} & \multicolumn{2}{c}{Adapted by GPT 4o Mini} & \multicolumn{2}{c}{Adapted by Llama 3.1 70B} \\
    \cmidrule(lr){1-2}\cmidrule(lr){3-6}\cmidrule(lr){7-8}\cmidrule(lr){9-10}
    feature & metric & out-comm. & in-comm. & random & specific & random & specific & random & specific \\
    \midrule 
        \\ [-10pt]
        \multicolumn{2}{c}{Structural Norms}
        \\\cmidrule(lr){1-2}
        Length & Avg. \# words     $\downarrow$  & +648.46 & +530.82 & +650.24 & +596.75 & \cellcolor{forestgreen!25} -96.31 & \cellcolor{forestgreen!25} -85.18 & \cellcolor{salmon!25} +1069.44 & \cellcolor{salmon!25} +991.08 \\
               & Avg. \# sentences $\downarrow$  & +31.36 & +23.67 & +30.81 & +28.35 & \cellcolor{forestgreen!25} -6.01 & \cellcolor{forestgreen!25} -5.59 & \cellcolor{salmon!25} +46.98 & \cellcolor{salmon!25} +44.74 \\
	Structural Artefacts & \% papers w/ tables $\uparrow$    & +6.06 & +11.51 & +6.31 & +11.31 & \cellcolor{salmon!25} -2.07 & \cellcolor{salmon!25} -2.79 & \cellcolor{salmon!25} -6.31 & \cellcolor{salmon!25} -11.31 \\
                     & \% papers w/ figures $\uparrow$    & +31.04 & +32.31 & +29.16 & +29.43 & \cellcolor{salmon!25} -10.8 & \cellcolor{salmon!25} -8.49 & \cellcolor{salmon!25} -29.16 & \cellcolor{salmon!25} -29.43 \\
    \midrule
        \\ [-10pt]
       \multicolumn{2}{c}{Stylistic Norms}
        \\\cmidrule(lr){1-2}
       Jargon      & Specificity score ($10^{-2}$) $\uparrow$    & -1.58 & +1.44 & -1.7 & +0.93 & \cellcolor{forestgreen!25} +0.92 & \cellcolor{forestgreen!25} +0.22 & \cellcolor{forestgreen!25} +0.45 & \cellcolor{salmon!25} -2.29 \\
       Formality   & Formality score ($10^{-2}$)   $\uparrow$    & +5.54 & +5.57 & +5.47 & +5.63 & \cellcolor{salmon!25} -0.34 & \cellcolor{salmon!25} -0.36 & \cellcolor{salmon!25} -0.11 & \cellcolor{salmon!25} -0.2 \\
       Readability & Flesch reading ease       $\uparrow$    & +28.23 & +32.87 & +28.02 & +29.39 & \cellcolor{salmon!25} -15.59 & \cellcolor{salmon!25} -18.38 & \cellcolor{salmon!25} -28.02 & \cellcolor{salmon!25} -29.39 \\
	\midrule
        \\ [-10pt]
        \multicolumn{2}{c}{Rhetorical Norms}
        \\\cmidrule(lr){1-2}
       Quant. Evidence & \% Sent. with QE     $\uparrow$    & +0.01 & +0.01 & +0.01 & +0.01 & \cellcolor{forestgreen!25} +0.03 & \cellcolor{forestgreen!25} +0.03 & \cellcolor{salmon!25} -0.01 & \cellcolor{salmon!25} -0.01 \\
    
    \bottomrule
    \end{tabular}
    }
    \caption{\small
        Results for the \textbf{natural language processing} community. The in-community column shows the metric value of papers from the community, out-community column shows the weighted average of data from all other communities. The random and specificity baselines show metric values before adaptation. The last six model columns show the \textbf{change} in value after adaptation from the random and specificity baselines, respectively. $\uparrow$ indicates that the metric should increase because the in-community value is > out-community value while $\downarrow$ indicates the vice versa. The cells where the $\Delta$ follows the expected trend are coloured {\color{forestgreen}green} while those that don't are coloured {\color{salmon}red}
    }
\end{table*}

\begin{table*}[th!]
    \centering
    \resizebox{1\linewidth}{!}{
    \footnotesize
    \begin{tabular}{llcccccccc}
    \toprule
    \multicolumn{2}{c}{Target = "Speech"}& \multicolumn{4}{c}{Baselines} & \multicolumn{2}{c}{Adapted by GPT 4o Mini} & \multicolumn{2}{c}{Adapted by Llama 3.1 70B} \\
    \cmidrule(lr){1-2}\cmidrule(lr){3-6}\cmidrule(lr){7-8}\cmidrule(lr){9-10}
    feature & metric & out-comm. & in-comm. & random & specific & random & specific & random & specific \\
    \midrule 
        \\ [-10pt]
        \multicolumn{2}{c}{Structural Norms}
        \\\cmidrule(lr){1-2}
        Length & Avg. \# words     $\downarrow$  & +628.73 & +528.17 & +671.9 & +589.02 & \cellcolor{forestgreen!25} -105.21 & \cellcolor{forestgreen!25} -82.43 & \cellcolor{salmon!25} +1152.94 & \cellcolor{salmon!25} +1159.77 \\
               & Avg. \# sentences $\downarrow$  & +29.94 & +25.18 & +31.43 & +27.48 & \cellcolor{forestgreen!25} -6.15 & \cellcolor{forestgreen!25} -4.61 & \cellcolor{salmon!25} +50.38 & \cellcolor{salmon!25} +51.35 \\
	Structural Artefacts & \% papers w/ tables $\downarrow$  & +7.64 & +2.87 & +7.1 & +4.11 & \cellcolor{forestgreen!25} -2.96 & \cellcolor{forestgreen!25} -1.57 & \cellcolor{forestgreen!25} -7.1 & \cellcolor{forestgreen!25} -4.11 \\
                     & \% papers w/ figures $\downarrow$  & +32.93 & +10.35 & +30.5 & +25.58 & \cellcolor{forestgreen!25} -10.86 & \cellcolor{forestgreen!25} -8.42 & \cellcolor{forestgreen!25} -30.5 & \cellcolor{forestgreen!25} -25.58 \\
    \midrule
        \\ [-10pt]
       \multicolumn{2}{c}{Stylistic Norms}
        \\\cmidrule(lr){1-2}
       Jargon      & Specificity score ($10^{-2}$) $\uparrow$    & -2.19 & +2.72 & -2.22 & +0.46 & \cellcolor{forestgreen!25} +2.08 & \cellcolor{forestgreen!25} +0.97 & \cellcolor{forestgreen!25} +0.69 & \cellcolor{salmon!25} -2.13 \\
       Formality   & Formality score ($10^{-2}$)   $\downarrow$  & +5.55 & +5.54 & +5.49 & +5.71 & \cellcolor{forestgreen!25} -0.37 & \cellcolor{forestgreen!25} -0.47 & \cellcolor{forestgreen!25} -0.11 & \cellcolor{forestgreen!25} -0.31 \\
       Readability & Flesch reading ease       $\downarrow$  & +29.44 & +27.26 & +27.02 & +27.08 & \cellcolor{forestgreen!25} -12.92 & \cellcolor{forestgreen!25} -16.51 & \cellcolor{forestgreen!25} -27.02 & \cellcolor{forestgreen!25} -27.08 \\
	\midrule
        \\ [-10pt]
        \multicolumn{2}{c}{Rhetorical Norms}
        \\\cmidrule(lr){1-2}
       Quant. Evidence & \% Sent. with QE     $\downarrow$  & +0.01 & +0.01 & +0.01 & +0.01 & \cellcolor{salmon!25} +0.03 & \cellcolor{salmon!25} +0.04 & \cellcolor{forestgreen!25} -0.01 & \cellcolor{forestgreen!25} -0.01 \\
    
    \bottomrule
    \end{tabular}
    }
    \caption{\small
        Results for the \textbf{speech} community. The in-community column shows the metric value of papers from the community, out-community column shows the weighted average of data from all other communities. The random and specificity baselines show metric values before adaptation. The last six model columns show the \textbf{change} in value after adaptation from the random and specificity baselines, respectively. $\uparrow$ indicates that the metric should increase because the in-community value is > out-community value while $\downarrow$ indicates the vice versa. The cells where the $\Delta$ follows the expected trend are coloured {\color{forestgreen}green} while those that don't are coloured {\color{salmon}red}
    }
    \label{table:new-speech}
\end{table*}

\begin{table*}[th!]
    \centering
    \resizebox{1\linewidth}{!}{
    \footnotesize
    \begin{tabular}{llcccccccc}
    \toprule
    \multicolumn{2}{c}{Target = "Web \& RecSys"}& \multicolumn{4}{c}{Baselines} & \multicolumn{2}{c}{Adapted by GPT 4o Mini} & \multicolumn{2}{c}{Adapted by Llama 3.1 70B} \\
    \cmidrule(lr){1-2}\cmidrule(lr){3-6}\cmidrule(lr){7-8}\cmidrule(lr){9-10}
    feature & metric & out-comm. & in-comm. & random & specific & random & specific & random & specific \\
    \midrule 
        \\ [-10pt]
        \multicolumn{2}{c}{Structural Norms}
        \\\cmidrule(lr){1-2}
        Length & Avg. \# words     $\downarrow$  & +621.97 & +615.61 & +644.49 & +545.42 & \cellcolor{forestgreen!25} -112.85 & \cellcolor{forestgreen!25} -98.02 & \cellcolor{salmon!25} +1223.56 & \cellcolor{salmon!25} +1098.99 \\
               & Avg. \# sentences $\downarrow$  & +29.66 & +28.74 & +30.38 & +25.55 & \cellcolor{forestgreen!25} -6.24 & \cellcolor{forestgreen!25} -5.35 & \cellcolor{salmon!25} +53.51 & \cellcolor{salmon!25} +47.44 \\
        Structural Artefacts & \% papers w/ tables $\downarrow$  & +7.39 & +5.79 & +5.42 & +4.01 & \cellcolor{forestgreen!25} -2.06 & \cellcolor{forestgreen!25} -1.21 & \cellcolor{forestgreen!25} -5.42 & \cellcolor{forestgreen!25} -4.01 \\
                             & \% papers w/ figures $\downarrow$  & +31.69 & +25.36 & +30.09 & +23.65 & \cellcolor{forestgreen!25} -10.45 & \cellcolor{forestgreen!25} -6.43 & \cellcolor{forestgreen!25} -30.09 & \cellcolor{forestgreen!25} -23.65 \\
    \midrule
        \\ [-10pt]
       \multicolumn{2}{c}{Stylistic Norms}
        \\\cmidrule(lr){1-2}
       Jargon      & Specificity score ($10^{-2}$) $\uparrow$    & -1.4 & +1.24 & -0.88 & +1.59 & \cellcolor{forestgreen!25} +1.37 & \cellcolor{forestgreen!25} +0.52 & \cellcolor{forestgreen!25} +0.31 & \cellcolor{salmon!25} -1.82 \\
       Formality   & Formality score ($10^{-2}$)   $\downarrow$  & +5.55 & +5.43 & +5.5 & +5.72 & \cellcolor{forestgreen!25} -0.38 & \cellcolor{forestgreen!25} -0.44 & \cellcolor{forestgreen!25} -0.14 & \cellcolor{forestgreen!25} -0.41 \\
       Readability & Flesch reading ease       $\downarrow$  & +29.41 & +27.19 & +28.28 & +27.19 & \cellcolor{forestgreen!25} -18.97 & \cellcolor{forestgreen!25} -18.02 & \cellcolor{forestgreen!25} -28.28 & \cellcolor{forestgreen!25} -27.19 \\
	\midrule
        \\ [-10pt]
        \multicolumn{2}{c}{Rhetorical Norms}
        \\\cmidrule(lr){1-2}
       Quant. Evidence & \% Sent. with QE     $\uparrow$    & +0.01 & +0.01 & +0.01 & +0.01 & \cellcolor{forestgreen!25} +0.03 & \cellcolor{forestgreen!25} +0.04 & \cellcolor{salmon!25} -0.01 & \cellcolor{salmon!25} -0.01 \\
    
    \bottomrule
    \end{tabular}
    }
    \caption{\small
        Results for the \textbf{web and recommendation systems} community. The in-community column shows the metric value of papers from the community, out-community column shows the weighted average of data from all other communities. The random and specificity baselines show metric values before adaptation. The last six model columns show the \textbf{change} in value after adaptation from the random and specificity baselines, respectively. $\uparrow$ indicates that the metric should increase because the in-community value is > out-community value while $\downarrow$ indicates the vice versa. The cells where the $\Delta$ follows the expected trend are coloured {\color{forestgreen}green} while those that don't are coloured {\color{salmon}red}
    }
    \label{table:new-wrs}    
\end{table*}

%% file: acl_latex.bbl
\begin{thebibliography}{46}
\providecommand{\natexlab}[1]{#1}

\bibitem[{Adilazuarda et~al.(2024)Adilazuarda, Mukherjee, Lavania, Singh, Aji, O{'}Neill, Modi, and Choudhury}]{adilazuarda-etal-2024-towards}
Muhammad~Farid Adilazuarda, Sagnik Mukherjee, Pradhyumna Lavania, Siddhant~Shivdutt Singh, Alham~Fikri Aji, Jacki O{'}Neill, Ashutosh Modi, and Monojit Choudhury. 2024.
\newblock \href {https://doi.org/10.18653/v1/2024.emnlp-main.882} {Towards measuring and modeling ``culture'' in {LLM}s: A survey}.
\newblock In \emph{Proceedings of the 2024 Conference on Empirical Methods in Natural Language Processing}, pages 15763--15784, Miami, Florida, USA. Association for Computational Linguistics.

\bibitem[{Bhatt et~al.(2022)Bhatt, Dev, Talukdar, Dave, and Prabhakaran}]{Bhatt2022RecontextualizingFI}
Shaily Bhatt, Sunipa Dev, Partha Talukdar, Shachi Dave, and Vinodkumar Prabhakaran. 2022.
\newblock \href {https://aclanthology.org/2022.aacl-main.55} {Re-contextualizing fairness in {NLP}: The case of {I}ndia}.
\newblock In \emph{Proceedings of the 2nd Conference of the Asia-Pacific Chapter of the Association for Computational Linguistics and the 12th International Joint Conference on Natural Language Processing (Volume 1: Long Papers)}, pages 727--740, Online only. Association for Computational Linguistics.

\bibitem[{Bhatt and Diaz(2024)}]{bhatt-diaz-2024-extrinsic}
Shaily Bhatt and Fernando Diaz. 2024.
\newblock \href {https://doi.org/10.18653/v1/2024.findings-emnlp.942} {Extrinsic evaluation of cultural competence in large language models}.
\newblock In \emph{Findings of the Association for Computational Linguistics: EMNLP 2024}, pages 16055--16074, Miami, Florida, USA. Association for Computational Linguistics.

\bibitem[{Bird et~al.(2009)Bird, Klein, and Loper}]{bird2009natural}
Steven Bird, Ewan Klein, and Edward Loper. 2009.
\newblock \emph{Natural language processing with {P}ython: {A}nalyzing text with the {N}atural {L}anguage {T}ool{K}it}.
\newblock O'Reilly Media, Inc.

\bibitem[{Birhane et~al.(2022)Birhane, Kalluri, Card, Agnew, Dotan, and Bao}]{birhane2022values}
Abeba Birhane, Pratyusha Kalluri, Dallas Card, William Agnew, Ravit Dotan, and Michelle Bao. 2022.
\newblock \href {https://doi.org/10.1145/3531146.3533083} {The values encoded in machine learning research}.
\newblock In \emph{Proceedings of the 2022 ACM Conference on Fairness, Accountability, and Transparency}, FAccT '22, page 173–184, New York, NY, USA. Association for Computing Machinery.

\bibitem[{Blodgett et~al.(2020)Blodgett, Barocas, Daum{\'e}~III, and Wallach}]{blodgett-etal-2020-language}
Su~Lin Blodgett, Solon Barocas, Hal Daum{\'e}~III, and Hanna Wallach. 2020.
\newblock \href {https://doi.org/10.18653/v1/2020.acl-main.485} {Language (technology) is power: A critical survey of {\textquotedblleft}bias{\textquotedblright} in {NLP}}.
\newblock In \emph{Proceedings of the 58th Annual Meeting of the Association for Computational Linguistics}, pages 5454--5476, Online. Association for Computational Linguistics.

\bibitem[{Chen et~al.(2025)Chen, Teplitskiy, and Jurgens}]{chen2025noisypathsourcecitation}
Hong Chen, Misha Teplitskiy, and David Jurgens. 2025.
\newblock \href {https://arxiv.org/abs/2502.20581} {The noisy path from source to citation: Measuring how scholars engage with past research}.
\newblock \emph{ArXiv preprint}, abs/2502.20581.

\bibitem[{Deardorff(2009)}]{deardorff_sage_2009}
Darla~K. Deardorff. 2009.
\newblock \href {https://doi.org/10.4135/9781071872987} {\emph{The {SAGE} {Handbook} of {Intercultural} {Competence}}}.
\newblock SAGE Publications, Inc, 2455 Teller Road,Thousand Oaks California 91320.

\bibitem[{Delgado et~al.(2023)Delgado, Yang, Madaio, and Yang}]{delgado2023participatory}
Fernando Delgado, Stephen Yang, Michael Madaio, and Qian Yang. 2023.
\newblock \href {https://doi.org/10.1145/3617694.3623261} {The participatory turn in {AI} design: {T}heoretical foundations and the current state of practice}.
\newblock In \emph{Proceedings of the 3rd ACM Conference on Equity and Access in Algorithms, Mechanisms, and Optimization}, EAAMO '23, New York, NY, USA. Association for Computing Machinery.

\bibitem[{Dementieva et~al.(2023)Dementieva, Babakov, and Panchenko}]{dementieva-etal-2023-detecting}
Daryna Dementieva, Nikolay Babakov, and Alexander Panchenko. 2023.
\newblock \href {https://aclanthology.org/2023.ranlp-1.31/} {Detecting text formality: A study of text classification approaches}.
\newblock In \emph{Proceedings of the 14th International Conference on Recent Advances in Natural Language Processing}, pages 274--284, Varna, Bulgaria. INCOMA Ltd., Shoumen, Bulgaria.

\bibitem[{Flesch(1948)}]{flesch1948new}
Rudolph Flesch. 1948.
\newblock A new readability yardstick.
\newblock \emph{Journal of Applied Psychology}, 32(3):221.

\bibitem[{Fok et~al.(2023)Fok, Kambhamettu, Soldaini, Bragg, Lo, Hearst, Head, and Weld}]{fok2023scim}
Raymond Fok, Hita Kambhamettu, Luca Soldaini, Jonathan Bragg, Kyle Lo, Marti Hearst, Andrew Head, and Daniel~S Weld. 2023.
\newblock \href {https://doi.org/10.1145/3581641.3584034} {Scim: Intelligent skimming support for scientific papers}.
\newblock In \emph{Proceedings of the 28th International Conference on Intelligent User Interfaces}, IUI '23, page 476–490, New York, NY, USA. Association for Computing Machinery.

\bibitem[{Grattafiori et~al.(2024)Grattafiori, Dubey, Jauhri, Pandey, Kadian, Al-Dahle, Letman, Mathur, Schelten, Vaughan, Yang, Fan, Goyal, Hartshorn, Yang, Mitra, Sravankumar, Korenev, Hinsvark, Rao, Zhang, Rodriguez, Gregerson, Spataru, Roziere, Biron, Tang, Chern, Caucheteux, Nayak, Bi, Marra, McConnell, Keller, Touret, Wu, Wong, Ferrer, Nikolaidis, Allonsius, Song, Pintz, Livshits, Wyatt, Esiobu, Choudhary, Mahajan, Garcia-Olano, Perino, Hupkes, Lakomkin, AlBadawy, Lobanova, Dinan, Smith, Radenovic, Guzmán, Zhang, Synnaeve, Lee, Anderson, Thattai, Nail, Mialon, Pang, Cucurell, Nguyen, Korevaar, Xu, Touvron, Zarov, Ibarra, Kloumann, Misra, Evtimov, Zhang, Copet, Lee, Geffert, Vranes, Park, Mahadeokar, Shah, van~der Linde, Billock, Hong, Lee, Fu, Chi, Huang, Liu, Wang, Yu, Bitton, Spisak, Park, Rocca, Johnstun, Saxe, Jia, Alwala, Prasad, Upasani, Plawiak, Li, Heafield, Stone, El-Arini, Iyer, Malik, Chiu, Bhalla, Lakhotia, Rantala-Yeary, van~der Maaten, Chen, Tan, Jenkins, Martin, Madaan, Malo, Blecher,
  Landzaat, de~Oliveira, Muzzi, Pasupuleti, Singh, Paluri, Kardas, Tsimpoukelli, Oldham, Rita, Pavlova, Kambadur, Lewis, Si, Singh, Hassan, Goyal, Torabi, Bashlykov, Bogoychev, Chatterji, Zhang, Duchenne, Çelebi, Alrassy, Zhang, Li, Vasic, Weng, Bhargava, Dubal, Krishnan, Koura, Xu, He, Dong, Srinivasan, Ganapathy, Calderer, Cabral, Stojnic, Raileanu, Maheswari, Girdhar, Patel, Sauvestre, Polidoro, Sumbaly, Taylor, Silva, Hou, Wang, Hosseini, Chennabasappa, Singh, Bell, Kim, Edunov, Nie, Narang, Raparthy, Shen, Wan, Bhosale, Zhang, Vandenhende, Batra, Whitman, Sootla, Collot, Gururangan, Borodinsky, Herman, Fowler, Sheasha, Georgiou, Scialom, Speckbacher, Mihaylov, Xiao, Karn, Goswami, Gupta, Ramanathan, Kerkez, Gonguet, Do, Vogeti, Albiero, Petrovic, Chu, Xiong, Fu, Meers, Martinet, Wang, Wang, Tan, Xia, Xie, Jia, Wang, Goldschlag, Gaur, Babaei, Wen, Song, Zhang, Li, Mao, Coudert, Yan, Chen, Papakipos, Singh, Srivastava, Jain, Kelsey, Shajnfeld, Gangidi, Victoria, Goldstand, Menon, Sharma, Boesenberg,
  Baevski, Feinstein, Kallet, Sangani, Teo, Yunus, Lupu, Alvarado, Caples, Gu, Ho, Poulton, Ryan, Ramchandani, Dong, Franco, Goyal, Saraf, Chowdhury, Gabriel, Bharambe, Eisenman, Yazdan, James, Maurer, Leonhardi, Huang, Loyd, Paola, Paranjape, Liu, Wu, Ni, Hancock, Wasti, Spence, Stojkovic, Gamido, Montalvo, Parker, Burton, Mejia, Liu, Wang, Kim, Zhou, Hu, Chu, Cai, Tindal, Feichtenhofer, Gao, Civin, Beaty, Kreymer, Li, Adkins, Xu, Testuggine, David, Parikh, Liskovich, Foss, Wang, Le, Holland, Dowling, Jamil, Montgomery, Presani, Hahn, Wood, Le, Brinkman, Arcaute, Dunbar, Smothers, Sun, Kreuk, Tian, Kokkinos, Ozgenel, Caggioni, Kanayet, Seide, Florez, Schwarz, Badeer, Swee, Halpern, Herman, Sizov, Guangyi, Zhang, Lakshminarayanan, Inan, Shojanazeri, Zou, Wang, Zha, Habeeb, Rudolph, Suk, Aspegren, Goldman, Zhan, Damlaj, Molybog, Tufanov, Leontiadis, Veliche, Gat, Weissman, Geboski, Kohli, Lam, Asher, Gaya, Marcus, Tang, Chan, Zhen, Reizenstein, Teboul, Zhong, Jin, Yang, Cummings, Carvill, Shepard, McPhie,
  Torres, Ginsburg, Wang, Wu, U, Saxena, Khandelwal, Zand, Matosich, Veeraraghavan, Michelena, Li, Jagadeesh, Huang, Chawla, Huang, Chen, Garg, A, Silva, Bell, Zhang, Guo, Yu, Moshkovich, Wehrstedt, Khabsa, Avalani, Bhatt, Mankus, Hasson, Lennie, Reso, Groshev, Naumov, Lathi, Keneally, Liu, Seltzer, Valko, Restrepo, Patel, Vyatskov, Samvelyan, Clark, Macey, Wang, Hermoso, Metanat, Rastegari, Bansal, Santhanam, Parks, White, Bawa, Singhal, Egebo, Usunier, Mehta, Laptev, Dong, Cheng, Chernoguz, Hart, Salpekar, Kalinli, Kent, Parekh, Saab, Balaji, Rittner, Bontrager, Roux, Dollar, Zvyagina, Ratanchandani, Yuvraj, Liang, Alao, Rodriguez, Ayub, Murthy, Nayani, Mitra, Parthasarathy, Li, Hogan, Battey, Wang, Howes, Rinott, Mehta, Siby, Bondu, Datta, Chugh, Hunt, Dhillon, Sidorov, Pan, Mahajan, Verma, Yamamoto, Ramaswamy, Lindsay, Lindsay, Feng, Lin, Zha, Patil, Shankar, Zhang, Zhang, Wang, Agarwal, Sajuyigbe, Chintala, Max, Chen, Kehoe, Satterfield, Govindaprasad, Gupta, Deng, Cho, Virk, Subramanian, Choudhury,
  Goldman, Remez, Glaser, Best, Koehler, Robinson, Li, Zhang, Matthews, Chou, Shaked, Vontimitta, Ajayi, Montanez, Mohan, Kumar, Mangla, Ionescu, Poenaru, Mihailescu, Ivanov, Li, Wang, Jiang, Bouaziz, Constable, Tang, Wu, Wang, Wu, Gao, Kleinman, Chen, Hu, Jia, Qi, Li, Zhang, Zhang, Adi, Nam, Yu, Wang, Zhao, Hao, Qian, Li, He, Rait, DeVito, Rosnbrick, Wen, Yang, Zhao, and Ma}]{grattafiori2024llama3herdmodels}
Aaron Grattafiori, Abhimanyu Dubey, Abhinav Jauhri, Abhinav Pandey, Abhishek Kadian, Ahmad Al-Dahle, Aiesha Letman, Akhil Mathur, Alan Schelten, Alex Vaughan, Amy Yang, Angela Fan, Anirudh Goyal, Anthony Hartshorn, Aobo Yang, Archi Mitra, Archie Sravankumar, Artem Korenev, Arthur Hinsvark, and 542 others. 2024.
\newblock \href {https://arxiv.org/abs/2407.21783} {The {L}lama 3 herd of models}.
\newblock \emph{Preprint}, arXiv:2407.21783.

\bibitem[{Guo et~al.(2024)Guo, Shang, Vazirgiannis, and Clavel}]{guo-etal-2024-curious}
Yanzhu Guo, Guokan Shang, Michalis Vazirgiannis, and Chlo{\'e} Clavel. 2024.
\newblock \href {https://doi.org/10.18653/v1/2024.findings-naacl.228} {The curious decline of linguistic diversity: Training language models on synthetic text}.
\newblock In \emph{Findings of the Association for Computational Linguistics: NAACL 2024}, pages 3589--3604, Mexico City, Mexico. Association for Computational Linguistics.

\bibitem[{Gururaja et~al.(2023)Gururaja, Bertsch, Na, Widder, and Strubell}]{gururaja-etal-2023-build}
Sireesh Gururaja, Amanda Bertsch, Clara Na, David Widder, and Emma Strubell. 2023.
\newblock \href {https://doi.org/10.18653/v1/2023.emnlp-main.822} {To build our future, we must know our past: Contextualizing paradigm shifts in natural language processing}.
\newblock In \emph{Proceedings of the 2023 Conference on Empirical Methods in Natural Language Processing}, pages 13310--13325, Singapore. Association for Computational Linguistics.

\bibitem[{Hovy and Yang(2021)}]{hovy-yang-2021-importance}
Dirk Hovy and Diyi Yang. 2021.
\newblock \href {https://doi.org/10.18653/v1/2021.naacl-main.49} {The importance of modeling social factors of language: Theory and practice}.
\newblock In \emph{Proceedings of the 2021 Conference of the North American Chapter of the Association for Computational Linguistics: Human Language Technologies}, pages 588--602, Online. Association for Computational Linguistics.

\bibitem[{Jiang et~al.(2025)Jiang, August, Soldaini, Lo, and Antoniak}]{jiang2025automaticdetectionresearchvalues}
Hang Jiang, Tal August, Luca Soldaini, Kyle Lo, and Maria Antoniak. 2025.
\newblock \href {https://arxiv.org/abs/2502.16390} {Automatic detection of research values from scientific abstracts across computer science subfields}.
\newblock \emph{ArXiv preprint}, abs/2502.16390.

\bibitem[{Jurgens et~al.(2018)Jurgens, Kumar, Hoover, McFarland, and Jurafsky}]{jurgens-etal-2018-measuring}
David Jurgens, Srijan Kumar, Raine Hoover, Dan McFarland, and Dan Jurafsky. 2018.
\newblock \href {https://doi.org/10.1162/tacl_a_00028} {Measuring the evolution of a scientific field through citation frames}.
\newblock \emph{Transactions of the Association for Computational Linguistics}, 6:391--406.

\bibitem[{Kroeber and Kluckhohn(1952)}]{culture_kk_book}
A.~L. Kroeber and Clyde Kluckhohn. 1952.
\newblock \href {https://iiif.lib.harvard.edu/manifests/view/drs:427692955$1i} {\emph{Culture: A Critical Review of Concepts and Definitions}}.
\newblock Peabody Museum Press, Cambridge, Massachusetts.

\bibitem[{Lepp and Smith(2025)}]{lepp2025you}
Haley Lepp and Daniel~Scott Smith. 2025.
\newblock ``you cannot sound like {GPT}''': {S}igns of language discrimination and resistance in computer science publishing.
\newblock \emph{arXiv preprint arXiv:2505.08127}.

\bibitem[{Leydesdorff et~al.(2019)Leydesdorff, Wagner, and Bornmann}]{leydesdorff2019interdisciplinarity}
Loet Leydesdorff, Caroline~S Wagner, and Lutz Bornmann. 2019.
\newblock Interdisciplinarity as diversity in citation patterns among journals: {R}ao-{S}tirling diversity, relative variety, and the {G}ini coefficient.
\newblock \emph{Journal of Informetrics}, 13(1):255--269.

\bibitem[{Liang et~al.(2024)Liang, Zhang, Wu, Lepp, Ji, Zhao, Cao, Liu, He, Huang, Yang, Potts, Manning, and Zou}]{liang2024mappingincreasingusellms}
Weixin Liang, Yaohui Zhang, Zhengxuan Wu, Haley Lepp, Wenlong Ji, Xuandong Zhao, Hancheng Cao, Sheng Liu, Siyu He, Zhi Huang, Diyi Yang, Christopher Potts, Christopher~D Manning, and James~Y. Zou. 2024.
\newblock \href {https://arxiv.org/abs/2404.01268} {Mapping the increasing use of {LLM}s in scientific papers}.
\newblock \emph{ArXiv preprint}, abs/2404.01268.

\bibitem[{Liao et~al.(2024)Liao, Antoniak, Cheong, Cheng, Lee, Lo, Chang, and Zhang}]{liao2024llmsresearchtoolslarge}
Zhehui Liao, Maria Antoniak, Inyoung Cheong, Evie Yu-Yen Cheng, Ai-Heng Lee, Kyle Lo, Joseph~Chee Chang, and Amy~X. Zhang. 2024.
\newblock \href {https://arxiv.org/abs/2411.05025} {{LLM}s as research tools: {A} large scale survey of researchers' usage and perceptions}.
\newblock \emph{ArXiv preprint}, abs/2411.05025.

\bibitem[{Linxen et~al.(2021)Linxen, Sturm, Br\"{u}hlmann, Cassau, Opwis, and Reinecke}]{linxen2021weird}
Sebastian Linxen, Christian Sturm, Florian Br\"{u}hlmann, Vincent Cassau, Klaus Opwis, and Katharina Reinecke. 2021.
\newblock \href {https://doi.org/10.1145/3411764.3445488} {How {WEIRD} is {CHI}?}
\newblock In \emph{Proceedings of the 2021 CHI Conference on Human Factors in Computing Systems}, CHI '21, New York, NY, USA. Association for Computing Machinery.

\bibitem[{Lu et~al.(2024)Lu, Sclar, Hallinan, Mireshghallah, Liu, Han, Ettinger, Jiang, Chandu, Dziri, and Choi}]{lu2025aihumanityssalieriquantifying}
Ximing Lu, Melanie Sclar, Skyler Hallinan, Niloofar Mireshghallah, Jiacheng Liu, Seungju Han, Allyson Ettinger, Liwei Jiang, Khyathi Chandu, Nouha Dziri, and Yejin Choi. 2024.
\newblock \href {https://arxiv.org/abs/2410.04265} {{AI} as humanity's {S}alieri: {Q}uantifying linguistic creativity of language models via systematic attribution of machine text against web text}.
\newblock \emph{ArXiv preprint}, abs/2410.04265.

\bibitem[{Lucy et~al.(2023)Lucy, Dodge, Bamman, and Keith}]{lucy-etal-2023-words}
Li~Lucy, Jesse Dodge, David Bamman, and Katherine Keith. 2023.
\newblock \href {https://doi.org/10.18653/v1/2023.findings-acl.433} {Words as gatekeepers: Measuring discipline-specific terms and meanings in scholarly publications}.
\newblock In \emph{Findings of the Association for Computational Linguistics: ACL 2023}, pages 6929--6947, Toronto, Canada. Association for Computational Linguistics.

\bibitem[{Majumder et~al.(2024)Majumder, Surana, Agarwal, Hazra, Sabharwal, and Clark}]{majumder2024datadrivendiscoverylargegenerative}
Bodhisattwa~Prasad Majumder, Harshit Surana, Dhruv Agarwal, Sanchaita Hazra, Ashish Sabharwal, and Peter Clark. 2024.
\newblock \href {https://arxiv.org/abs/2402.13610} {Data-driven discovery with large generative models}.
\newblock \emph{ArXiv preprint}, abs/2402.13610.

\bibitem[{Michael et~al.(2023)Michael, Holtzman, Parrish, Mueller, Wang, Chen, Madaan, Nangia, Pang, Phang, and Bowman}]{michael-etal-2023-nlp}
Julian Michael, Ari Holtzman, Alicia Parrish, Aaron Mueller, Alex Wang, Angelica Chen, Divyam Madaan, Nikita Nangia, Richard~Yuanzhe Pang, Jason Phang, and Samuel~R. Bowman. 2023.
\newblock \href {https://doi.org/10.18653/v1/2023.acl-long.903} {What do {NLP} researchers believe? {R}esults of the {NLP} community metasurvey}.
\newblock In \emph{Proceedings of the 61st Annual Meeting of the Association for Computational Linguistics (Volume 1: Long Papers)}, pages 16334--16368, Toronto, Canada. Association for Computational Linguistics.

\bibitem[{Nathani et~al.(2025)Nathani, Madaan, Roberts, Bashlykov, Menon, Moens, Budhiraja, Magka, Vorotilov, Chaurasia, Hupkes, Cabral, Shavrina, Foerster, Bachrach, Wang, and Raileanu}]{nathani2025mlgymnewframeworkbenchmark}
Deepak Nathani, Lovish Madaan, Nicholas Roberts, Nikolay Bashlykov, Ajay Menon, Vincent Moens, Amar Budhiraja, Despoina Magka, Vladislav Vorotilov, Gaurav Chaurasia, Dieuwke Hupkes, Ricardo~Silveira Cabral, Tatiana Shavrina, Jakob Foerster, Yoram Bachrach, William~Yang Wang, and Roberta Raileanu. 2025.
\newblock \href {https://arxiv.org/abs/2502.14499} {Mlgym: A new framework and benchmark for advancing {AI} research agents}.
\newblock \emph{ArXiv preprint}, abs/2502.14499.

\bibitem[{Ovalle et~al.(2023)Ovalle, Subramonian, Gautam, Gee, and Chang}]{10.1145/3600211.3604705}
Anaelia Ovalle, Arjun Subramonian, Vagrant Gautam, Gilbert Gee, and Kai-Wei Chang. 2023.
\newblock \href {https://doi.org/10.1145/3600211.3604705} {Factoring the matrix of domination: A critical review and reimagination of intersectionality in {AI} fairness}.
\newblock In \emph{Proceedings of the 2023 AAAI/ACM Conference on AI, Ethics, and Society}, AIES '23, page 496–511, New York, NY, USA. Association for Computing Machinery.

\bibitem[{Pawar et~al.(2024)Pawar, Park, Jin, Arora, Myung, Yadav, Haznitrama, Song, Oh, and Augenstein}]{pawar2024surveyculturalawarenesslanguage}
Siddhesh Pawar, Junyeong Park, Jiho Jin, Arnav Arora, Junho Myung, Srishti Yadav, Faiz~Ghifari Haznitrama, Inhwa Song, Alice Oh, and Isabelle Augenstein. 2024.
\newblock \href {https://arxiv.org/abs/2411.00860} {Survey of cultural awareness in language models: Text and beyond}.
\newblock \emph{ArXiv preprint}, abs/2411.00860.

\bibitem[{Qadri et~al.(2025)Qadri, Diaz, Wang, and Madaio}]{qadri2025casethickevaluationscultural}
Rida Qadri, Mark Diaz, Ding Wang, and Michael Madaio. 2025.
\newblock \href {https://arxiv.org/abs/2503.19075} {The case for ``thick evaluations'' of cultural representation in {AI}}.
\newblock \emph{ArXiv preprint}, abs/2503.19075.

\bibitem[{Rao et~al.(2024)Rao, Yerukola, Shah, Reinecke, and Sap}]{rao2024normad}
Abhinav Rao, Akhila Yerukola, Vishwa Shah, Katharina Reinecke, and Maarten Sap. 2024.
\newblock \href {https://arxiv.org/abs/2404.12464} {Normad: A benchmark for measuring the cultural adaptability of large language models}.
\newblock \emph{Preprint}, arXiv:2404.12464.

\bibitem[{Rao and Tetreault(2018)}]{rao-tetreault-2018-dear}
Sudha Rao and Joel Tetreault. 2018.
\newblock \href {https://doi.org/10.18653/v1/N18-1012} {Dear sir or madam, may {I} introduce the {GYAFC} dataset: Corpus, benchmarks and metrics for formality style transfer}.
\newblock In \emph{Proceedings of the 2018 Conference of the North {A}merican Chapter of the Association for Computational Linguistics: Human Language Technologies, Volume 1 (Long Papers)}, pages 129--140, New Orleans, Louisiana. Association for Computational Linguistics.

\bibitem[{Robinson et~al.(2024)Robinson, Alvarez, and Mekler}]{Robinson_2024}
Raquel~Breejon Robinson, Alberto Alvarez, and Elisa~D. Mekler. 2024.
\newblock \href {https://doi.org/10.1145/3613905.3644051} {How to write a {CHI} paper (asking for a friend)}.
\newblock In \emph{Extended Abstracts of the CHI Conference on Human Factors in Computing Systems}, CHI ’24. ACM.

\bibitem[{Sambasivan et~al.(2021)Sambasivan, Arnesen, Hutchinson, Doshi, and Prabhakaran}]{sambasivan2021reimagining}
Nithya Sambasivan, Erin Arnesen, Ben Hutchinson, Tulsee Doshi, and Vinodkumar Prabhakaran. 2021.
\newblock \href {https://doi.org/10.1145/3442188.3445896} {Re-imagining algorithmic fairness in {I}ndia and beyond}.
\newblock In \emph{Proceedings of the 2021 ACM Conference on Fairness, Accountability, and Transparency}, FAccT '21, page 315–328, New York, NY, USA. Association for Computing Machinery.

\bibitem[{Schmidgall et~al.(2025)Schmidgall, Su, Wang, Sun, Wu, Yu, Liu, Liu, and Barsoum}]{schmidgall2025agentlaboratoryusingllm}
Samuel Schmidgall, Yusheng Su, Ze~Wang, Ximeng Sun, Jialian Wu, Xiaodong Yu, Jiang Liu, Zicheng Liu, and Emad Barsoum. 2025.
\newblock \href {https://arxiv.org/abs/2501.04227} {Agent laboratory: {U}sing {LLM} agents as research assistants}.
\newblock \emph{ArXiv preprint}, abs/2501.04227.

\bibitem[{Shaib et~al.(2024)Shaib, Elazar, Li, and Wallace}]{shaib-etal-2024-detection}
Chantal Shaib, Yanai Elazar, Junyi~Jessy Li, and Byron~C Wallace. 2024.
\newblock \href {https://doi.org/10.18653/v1/2024.emnlp-main.368} {Detection and measurement of syntactic templates in generated text}.
\newblock In \emph{Proceedings of the 2024 Conference on Empirical Methods in Natural Language Processing}, pages 6416--6431, Miami, Florida, USA. Association for Computational Linguistics.

\bibitem[{Si et~al.(2024)Si, Yang, and Hashimoto}]{si2024llmsgeneratenovelresearch}
Chenglei Si, Diyi Yang, and Tatsunori Hashimoto. 2024.
\newblock \href {https://arxiv.org/abs/2409.04109} {Can {LLM}s generate novel research ideas? {A} large-scale human study with 100+ {NLP} researchers}.
\newblock \emph{ArXiv preprint}, abs/2409.04109.

\bibitem[{Sorensen et~al.(2024)Sorensen, Moore, Fisher, Gordon, Mireshghallah, Rytting, Ye, Jiang, Lu, Dziri, Althoff, and Choi}]{sorensen2024roadmap}
Taylor Sorensen, Jared Moore, Jillian Fisher, Mitchell Gordon, Niloofar Mireshghallah, Christopher~Michael Rytting, Andre Ye, Liwei Jiang, Ximing Lu, Nouha Dziri, Tim Althoff, and Yejin Choi. 2024.
\newblock Position: A roadmap to pluralistic alignment.
\newblock In \emph{Proceedings of the 41st International Conference on Machine Learning}, ICML'24. JMLR.org.

\bibitem[{Subramonian et~al.(2024)Subramonian, Gautam, Klakow, and Talat}]{subramonian-etal-2024-understanding}
Arjun Subramonian, Vagrant Gautam, Dietrich Klakow, and Zeerak Talat. 2024.
\newblock \href {https://doi.org/10.18653/v1/2024.emnlp-main.184} {Understanding {\textquotedblleft}democratization{\textquotedblright} in {NLP} and {ML} research}.
\newblock In \emph{Proceedings of the 2024 Conference on Empirical Methods in Natural Language Processing}, pages 3151--3166, Miami, Florida, USA. Association for Computational Linguistics.

\bibitem[{West and Portenoy(2016)}]{west-portenoy-2016-delineating}
Jevin West and Jason Portenoy. 2016.
\newblock \href {https://aclanthology.org/W16-1508} {Delineating fields using mathematical jargon}.
\newblock In \emph{Proceedings of the Joint Workshop on Bibliometric-enhanced Information Retrieval and Natural Language Processing for Digital Libraries ({BIRNDL})}, pages 63--71.

\bibitem[{Xu et~al.(2025)Xu, Jojic, Rao, Brockett, and Dolan}]{xu2024echoesaiquantifyinglack}
Weijia Xu, Nebojsa Jojic, Sudha Rao, Chris Brockett, and Bill Dolan. 2025.
\newblock \href {https://arxiv.org/abs/2501.00273} {Echoes in {AI}: {Q}uantifying lack of plot diversity in {LLM} outputs}.
\newblock \emph{ArXiv preprint}, abs/2501.00273.

\bibitem[{Zhang et~al.(2017)Zhang, Hamilton, Danescu-Niculescu-Mizil, Jurafsky, and Leskovec}]{zhang2017community}
Justine Zhang, William Hamilton, Cristian Danescu-Niculescu-Mizil, Dan Jurafsky, and Jure Leskovec. 2017.
\newblock \href {https://doi.org/10.1609/icwsm.v11i1.14904} {Community identity and user engagement in a multi-community landscape}.
\newblock \emph{Proceedings of the International AAAI Conference on Web and Social Media}, 11(1):377--386.

\bibitem[{Zheng et~al.(2023)Zheng, Chiang, Sheng, Zhuang, Wu, Zhuang, Lin, Li, Li, Xing, Zhang, Gonzalez, and Stoica}]{zheng2023judging}
Lianmin Zheng, Wei-Lin Chiang, Ying Sheng, Siyuan Zhuang, Zhanghao Wu, Yonghao Zhuang, Zi~Lin, Zhuohan Li, Dacheng Li, Eric~P. Xing, Hao Zhang, Joseph~E. Gonzalez, and Ion Stoica. 2023.
\newblock Judging {LLM}-as-a-judge with {MT}-bench and {C}hatbot {A}rena.
\newblock In \emph{Proceedings of the 37th International Conference on Neural Information Processing Systems}, NeurIPS '23, Red Hook, NY, USA. Curran Associates Inc.

\bibitem[{Zhou et~al.(2025)Zhou, Bamman, and Bleaman}]{zhou2025culturetriviasocioculturaltheory}
Naitian Zhou, David Bamman, and Isaac~L. Bleaman. 2025.
\newblock \href {https://arxiv.org/abs/2502.12057} {Culture is not trivia: Sociocultural theory for cultural {NLP}}.
\newblock \emph{ArXiv preprint}, abs/2502.12057.

\end{thebibliography}
